\documentclass[lettersize,journal]{IEEEtran}
\usepackage{amsmath,amsfonts}
\usepackage{algorithm}
\usepackage{array}
\usepackage{textcomp}
\usepackage{stfloats}
\usepackage{url}
\usepackage{verbatim}
\usepackage{graphicx}
\usepackage{cite}
\usepackage[switch]{lineno}
\graphicspath{{./images/}}
\usepackage[pagebackref,breaklinks,colorlinks,citecolor=darkerblue]{hyperref}  

\usepackage{booktabs}       
\usepackage{nicefrac}       
\usepackage{microtype}      
\usepackage{listings}
\usepackage{float}
\usepackage{multirow}
\usepackage{subfigure}
\usepackage{wrapfig}
\usepackage{xcolor}  
\usepackage{colortbl}

\usepackage{mathtools} 
\usepackage{bbm} 
\usepackage{amsmath} 
\usepackage{enumitem} 
\usepackage{caption} 
\usepackage[edges]{forest}
\usepackage{bm}
\usepackage{algpseudocode}

\usepackage{tabularx}
\usepackage{mathabx}
\usepackage{cancel}

\definecolor{darkerblue}{rgb}{0,0.05,0.6}
\definecolor{darkgreen}{RGB}{34,139,34}

\definecolor{hidden-draw-green}{RGB}{92,226,194}
\definecolor{hidden-draw}{RGB}{106,142,189} 
\definecolor{hidden-blue}{RGB}{194,232,247} 
\definecolor{hidden-green}{RGB}{202,246,229}

\begin{document}

\title{A Survey on Mixup Augmentations and Beyond}

\author{Xin Jin$^*$, Hongyu Zhu$^*$, Siyuan Li$^*$, Zedong Wang, Zicheng Liu, Juanxi Tian, Chang Yu,\\ Huafeng Qin, Stan Z. Li~\IEEEmembership{Fellow,~IEEE}
\thanks{Xin Jin, Hongyu Zhu, Siyuan Li, Zedong Wang, Zicheng Liu, Juanxi Tian, Chang Yu, and Stan Z. Li are with the School of Engineering, Westlake University, Hangzhou, Zhejiang Province, China.
E-mails: jinxin4@ctbu.edu.cn, zhuhongyu@ctbu.edu.cn, lisiyuan@westlake.edu.cn, wangzedong@westlake.edu.cn, liuzicheng@westlake.edu.cn, juanxitian1031@gmail.com, yuchang@westlake.edu.cn, stan.zq.li@westlake.edu.cn
}
\thanks{Xin Jin, Hongyu Zhu, and Huafeng Qin are with the School of Artificial Intelligence, Chongqing Technology and Business University, Chongqing, China.
}
\thanks{Xin Jin, Hongyu Zhu, and Siyuan Li contributed equally to this work.}
\thanks{Corresponding Author: Stan Z. Li }
}

\markboth{PrePrint Version}{}


\maketitle

\begin{abstract}

As Deep Neural Networks have achieved thrilling breakthroughs in the past decade, data augmentations have garnered increasing attention as regularization techniques when massive labeled data are unavailable. Among existing augmentations, Mixup and relevant data-mixing methods that convexly combine selected samples and the corresponding labels are widely adopted because they yield high performances by generating data-dependent virtual data while easily migrating to various domains.
This survey presents a comprehensive review of foundational mixup methods and their applications.
We first elaborate on the training pipeline with mixup augmentations as a unified framework containing modules. A reformulated framework could contain various mixup methods and give intuitive operational procedures.
Then, we systematically investigate the applications of mixup augmentations on vision downstream tasks, various data modalities, and some analysis \& theorems of mixup. Meanwhile, we conclude the current status and limitations of mixup research and point out further work for effective and efficient mixup augmentations. This survey can provide researchers with the current state of the art in mixup methods and provide some insights and guidance roles in the mixup arena.
An online project with this survey is available at \url{https://github.com/Westlake-AI/Awesome-Mixup}.

\end{abstract}


\section{Introduction}
\label{sec:introduction}
\IEEEPARstart{D}{eep} Neural Networks (DNNs), such as Convolutional Neural Networks (CNNs) and Transformers, since their powerful feature representation ability that has been successfully applied to a variety of tasks, \emph{e.g.} Image Classification, Object Detection, and Natural Language Processing (NLP), \emph{etc.} To accomplish progressively more challenging tasks, DNNs employ a large number of learnable parameters, and that means that without numerous training data, models could easily overfit and fail to generalize. However, training data in some scenarios was unavailable and expensive to collect. Causing DNNs to generalize beyond limited training data is one of the fundamental problems of deep learning.

To address the data-hungry problem, researchers have proposed Data Augmentation (DA) techniques. Compared to ``model-centric'' and regularization methods, DA is a ``data-centric'' regularization technique that prevents over-fitting by synthesizing virtual training data. DA could introduce useful invariant features by constructing different versions of the same sample. The increase in dataset size and inductive bias brought about by DA also achieves a regularization effect to relieve the over-fitting problem. Recently, data augmentation has been shown to improve the generalization of deep learning models and become a key factor in achieving state-of-the-art performance. Data augmentation can synthesize new data through contrast combination, mixup, and generation.

In this survey, we focus on a burgeoning field-\textbf{Mixup}. 
MixUp\cite{zhang2018mixup} generates augmented samples by interpolating two samples with their one-hot labels to one. Essentially, Mixup-based methods mix multiple samples to generate augmented data. In contrast, most existing augmentation techniques modify single samples without altering their unique labels. Unlike these methods, Mixup generates augmented samples from two or more examples, leading to multiple labels that better reflect real-world conditions. Additionally, Mixup demonstrates strong transferability across different datasets and domains. In comparison, alternative combination methods often require extensive time to identify suitable augmentation strategies. The generative method is challenging to apply to large datasets, as it requires an additional generator and discriminator, hindering transferability and limiting application scenarios. In contrast, Mixup does not rely on label-retaining operations but uses a learnable approach to create more effective augmented samples. Unlike traditional data augmentation methods that process single samples, Mixup generates virtual training data by combining multiple samples, producing a large volume of training data without the need for domain knowledge."

Currently, mixup has been successfully applied to a variety of tasks and training paradigms, including Supervised Learning (SL), Self-Supervised learning (SSL), Semi-Supervised Learning (Semi-SL), NLP, Graph, and Speech. In Fig. \ref{fig: timeline}, we summarise the timeline of some mainstream methods under those training paradigms and data modalities:

\begin{figure*}[t]
    \centering
    \includegraphics[scale=0.58]{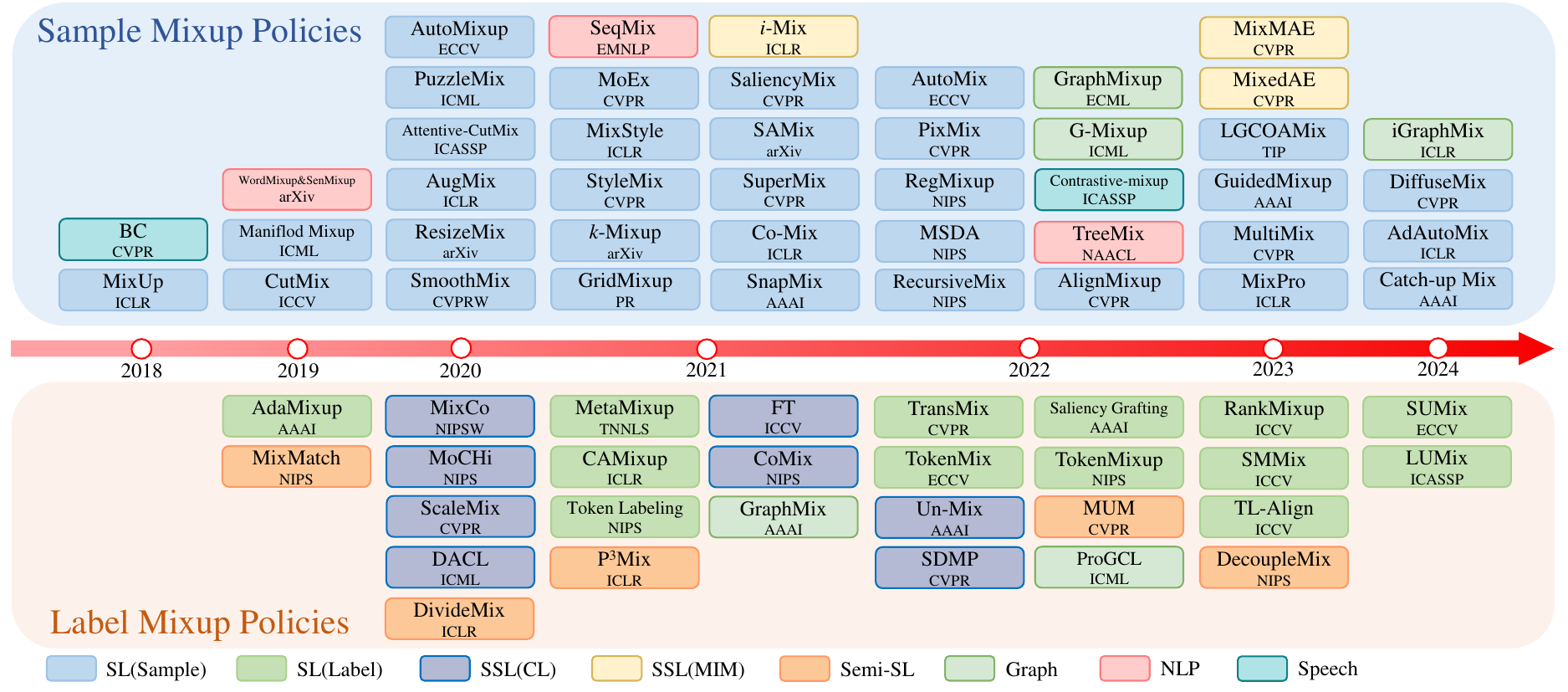}
    \caption{Research timeline in Mixup methods can be broadly categorized into \textbf{Sample Mixup Policies} and \textbf{Label Mixup Policies} from 2018 to 2024 according to the unified framework. We summarized some mainstream methods in 7 classes based on training paradigms and data modalities: SL based on Sample-level, SL based on Label-level, SSL based on CL, SSL based on MIM, Semi-SL, Graph, NLP, and Speech.}
    \label{fig: timeline}
\end{figure*}

\textbf{SL (Sample).} In 2018, MixUp\cite{zhang2018mixup} proposed a static linear interpolation way of mixing samples. In 2019, CutMix\cite{yun2019cutmix} and Manifold Mixup\cite{verma2019manifold} improved mixup into cutting-based and feature-based. There are ad-hoc methods. But from 2020 to 2023, numerous methods further improved mixup methods in static linear, cutting-based, and feature-based ways, also turning into an adaptive way. Until 2024, DiffuseMix\cite{islam2024diffusemix} combined a generative model and the mixup method.

\textbf{SL (Label).} In 2019, AdaMixup\cite{guo2018adamix} find that mixing ratio $\lambda$ affectd model performance, it's called \emph{``Mainfold Intrusion''}. Thus, from 2020 to 2024, many methods emerged to optimize these ratios based on CNNs or Vision Transformers (ViTs). Also, techniques such as CAMixup\cite{wen2020camix} in 2021 and RankMixup\cite{noh2023rankmixup} in 2023 were introduced to enhance model calibration.

\textbf{SSL (CL) \& SSL (MIM).} Contrastive Learning (CL) shows power ability in the Image Classification task. To improve model performance, researchers proposed lots of CL methods with mixup, these methods used mixup to obtain ``semi-positive'' samples to capture more features. CL + Mixup usually modifies its loss term for suitable SSL tasks. Masked Image Modeling (MIM) proposed reconstruction samples from mixed samples, they argued that mixed samples will share more features which could learn some high-dimension information. MixMAE\cite{liu2023mixmae} and MixedAE\cite{chen2023mixedae} had shown this point of view in 2023.

\textbf{Semi-SL.} can leverage both labeled and unlabeled information. In 2019, MixMatch\cite{berthelot2019mixmatch} used this way to improve the model performance and become more robust since mixed samples could be used as a clean image with a noise image. For PUL, P$^3$Mix\cite{li2021yourp3mix} obtained better accuracy by mixed samples from the decision boundary that is close to the boundary in 2021. DecoupledMix\cite{liu2022decoupledmix} proposed decoupled one of the samples predicted to obtain cleaner pseudo labels in 2023.

\textbf{Data Modality.} Not only design for \textbf{Image}.
For \textbf{NLP}, WordMixup \& SenMixup\cite{guo2019WordMixup&SenMixu} proposed two mixing ways for text, mixing with sentences or with embeddings in 2019. Following those two basic ways, many methods were proposed with specific modifications. \emph{e.g.} SeqMix\cite{zhang2020seqmix} proposed mixing embeddings based on their saliency in 2021, and TreeMix\cite{zhang2022treemix} proposed to decompose sentences into substructures by using constituent syntactic analysis and recombine them into new sentences by mixing.
For \textbf{Graph}, In 2021 and 2022, GraphMix\cite{verma2021graphmix} and ProGCL\cite{xia2021progcl} proposed graph classification with the mixup method, and they proposed some new loss terms combine mixup and graph for hard sample mining. And GraphMixup\cite{wu2021graphmixup}, G-Mixup\cite{han2022g-mix}, and iGraphMix\cite{jeong2023igraphmix} obtained mixed graph samples through saliency information to improve the model's classification ability and robustness in 2022 and 2024.
For \textbf{Speech}, BC\cite{cvpr2018bc} and Contrastive-mixup\cite{zhang2022contrastive-mixup} mix the speech data by linear interpolation directly.

Overall, compared to three published surveys \cite{cao2022mixup-survey1}, \cite{lewy2023mixup-survey2} and \cite{naveed2024mixup-survey3} on Mixup, our contributions include:
\begin{itemize}[leftmargin=1.5em]
    \item We provide a timely literature review and a comprehensive framework to conceptualize two different improvement strategies (Sample and Label) for the mixup method, using SL as an example. These two strategies could correspond to different training paradigms and data modalities.

    \item We carefully review and discuss the details of the techniques of various categories of mixup methods, such as Static Linear, Saliency-based, Attention-based, \emph{etc.}, to give researchers a better overview of the method involved, leading to further understanding and insights.

    \item We report a systematic survey of the application of mixup methods to downstream tasks, propose technical challenges and further demonstrate their broad applicability to other modalities and domains beyond vision tasks, \emph{e.g.}, audio, speech, graphics, biology, \emph{etc.}

    \item We further summarize the mixup method as a trainable paradigm, compared to other surveys that utilize it as a DA tool and methods. Additionally, we appeal to researchers to contribute a unified framework for mixup to address a variety of tasks rather than discrete task-specific modifications.
    
\end{itemize}

\section{Preliminary}
\label{sec:preliminary}
\begin{figure*}[t]
    \centering
    \includegraphics[scale=0.4]{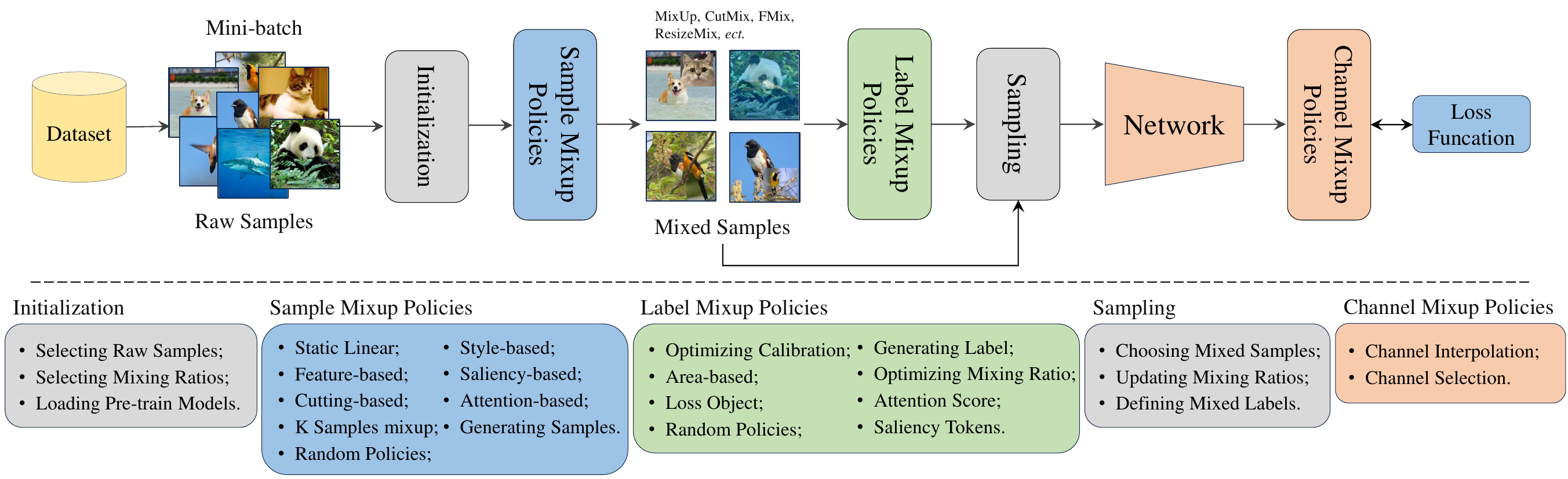}
    \caption{The unified framework of Mixup methods. The top part is the process of mixup methods. Sampling a Mini-batch of raw samples from the dataset. Then, the mixed samples are obtained through the \textbf{Initialization} and \textbf{Sample Mixup Policies} modules. After the \textbf{Label Mixup Policies} and \textbf{Sampling} modules, encoded by a network and through \textbf{Channel Mixup Policies}. Finally, the loss by the specific loss function. The down part displays detailed ways of each module in the mixup process.}
    \label{fig: pipeline}
\end{figure*}
\subsection{Notions}
Table \ref{Notion} and Table \ref{Abbreviations} list the notations and abbreviations used in this survey. We define a total sample set as $\mathbb{X} \in \mathbb{R}^{C \times W \times H}$, and corresponding label set as $\mathbb{Y} \in \mathbb{R}^{k}$. In Computer Vision (CV) tasks, $x \in \mathbb{X}^{C \times W \times H}$, in NLP tasks, $x \in \mathbb{X}^{C \times L}$, where $L$ denotes the original sentence, and in Graph Neural Network (GNN) tasks, $v$ noticed as a node and $\mathcal{V}$ represent the distribution of its neighborhood. In mixup methods, $\hat{x}$ and $\hat{y}$ denote mixed samples and labels, respectively. $\lambda$ denotes the mixing ratio that samples from the Beta or Uniform distribution. In addition, we use $\mathcal{M} \in \{0, 1\}$ to represent the mask obtained from some Ad-Hoc or Adaptive methods. Training models denotes $f_\theta(\cdot)$, where $\theta$ is learnable parameters. $f^\prime(\cdot)$ denotes pre-trained or teacher models, and $f^\star_\theta(\cdot)$ denotes optimized model fixed from $f_\theta(\cdot)$. In Self-Supervised Learning or Semi-Supervised Learning, $\tau$ denotes the temperature used for the Sharpen function or scaling of the pseudo-labels.
\begin{table}[ht]
    \centering
    \setlength{\tabcolsep}{0.2mm}
    \caption{Summary of the frequently used notations in mixup methods.}
    \resizebox{1.0\linewidth}{!}{
    \begin{tabular}{c c |c c} \hline
    \multicolumn{2}{c|}{Basic Notions} & \multicolumn{2}{c}{Functional Notions} \\ \hline
    $\mathbb{X}$ / $\mathbb{Y}$ / $\mathbb{N}$ & Sample / Label / Token set & $\mathcal{A}(\cdot)$ & Augmentation function \\
    $B$ / $C$ / $L$ & Batch-size / Channel / Length & $\mathcal{P}(\cdot)$ & Paste function \\
    $W$, $H$ / $w$, $h$ &  Image / Patch width, height & $T(\cdot)$ & Resizing function \\
    $x$ / $\hat{x}$ / $x_u$ & Raw/Mixed/Unlabeled sample & $\textrm{A}(\cdot)$ & Attention function \\
    $y$ / $\hat{y}$ & Raw / Mixed label & $s(\cdot)$ & Cosine similarity function\\
    $z$ / $\hat{z}$ & Raw / Mixed feature maps & $\textrm{norm}(\cdot)$ & Normalization function \\
    $c$ / $k$ & A class / Total classes & $\text{Bern}(\cdot)$ & Bernoulli matrix \\
    $\lambda$  & Mixing ratio  & $\text{Beta}(\alpha, \alpha)$ & Beta distribution \\
    $\lambda_{s}$ / $\lambda_{c}$ / $\lambda_\tau$ & Style / Content / Scale ratio & $U(\alpha, \beta)$ & Uniform distribution \\
    $v$ & A node & $\nabla(\cdot)$ & Gradient function \\
    $\bm{M}$ & Matrix & $\phi(\cdot)$ & Softmax function \\
    $\alpha$, $\beta$ & Parameter of distribution  & $\delta(\cdot)$ & Dirac function \\
    $\nu$ & A variable hyperparameter & $\mathcal{L}(\cdot)$ & Loss function \\
    $\mathcal{M} \in \{0, 1\}$ & Mask & $\text{En}(\cdot)$ & Encoder \\
    $\textrm{w}$ & Weight factor & $\text{De}(\cdot)$ & Decoder\\
    $\tau$ & Temperature & $\text{Dis}(\cdot)$ & Discriminator\\
    $Am$ & Amplitude & $\text{Cls}(\cdot)$ & Classifier \\
    $l$ & The $l$-th layer & $\odot$ & Element-wise multiplication \\
    $\mu$ / $\sigma$ & Mean / Standard deviation  & $\oplus$ & Dissimilarity operation \\
    $\theta$ & Model learnable parameter & $f(\cdot)$ & Traning Model \\
    $p$ / $P$ & Probability / Joint distribution & $f^\prime(\cdot)$ & Teacher model \\
    $\mathcal{V}$ & Neighborhood distribution & $f^\star(\cdot)$ & Optimized model\\ \hline

    \end{tabular}
    }
    \label{Notion}
\end{table}

\subsection{Mixup Framework Modules}
In this subsection, we will detail each module's functions in the mixup method pipeline, as shown in Fig. \ref{fig: pipeline}.

\begin{itemize}[leftmargin=1.5em]
    \item \textbf{Initialization.} Before mixup, some methods select raw samples within the mini-batch to filter those suitable for mixing, \emph{e.g.} Co-Mix\cite{kim2020co} selected suitable samples in the mini-batch to maximize the diversity of the mixed samples obtained. Besides filtering the samples, some saliency-based methods leveraged pre-trained models to locate and obtain feature maps for the samples. Finally, each method obtained the mixup ratio $\lambda$ from the Beta distribution.
    
    \item \textbf{Sample Mixup Policies.} In Supervised Learning, we divide the policies into 9 classes, and we detail these classes in Fig. \ref{mixup4cv}. \textbf{Static Linear} methods used $\lambda$ mixed two or more samples based on linear interpolation. \textbf{Fearture-based} methods used raw samples feature maps obtained by $f_\theta(\cdot)$, and mixed them in interpolation linear. \textbf{Cutting-based} methods are used in various ways, such as cutting, resizing, or stacking to mix samples, with the mixing ratio $\lambda$ from the mask area. \textbf{K Sample mixup} methods used more than 2 samples mixing. \textbf{Random Policies} methods combined lots of different augmentation methods and some hand-crafted mixup methods, and the policy is chosen by each method's weight factor. \textbf{Style-based} mixed samples from their style and content by an additional style extractor. \textbf{Saliency-based} methods used sample feature maps to locate their saliency information and obtained max feature mixed samples. \textbf{Attention-based} methods, similar to saliency-based methods, utilized attention scores rather than saliency maps. \textbf{Generating Samples} used some generative models such as GAN-based models\cite{goodfellow2020gans} and Diffusion-based models\cite{brooks2023instructpix2pix} to generating mixed samples.
    
    \item \textbf{Label Mixup Policies.} We divide into 8 classes in SL and also display them in Fig. \ref{mixup4cv} in detail. \textbf{Optimizing Calibration} methods used the ECE metric to rank the mixed samples and select them for improving classification performance and model calibration. \textbf{Area-based} methods used mask region redefine mixing ratio $\lambda$. \textbf{Loss Object} methods redefined a new mixup classification loss or proposed a new loss as a regularization method. \textbf{Random Policies} methods combined the other augmentation methods with mixup methods or proposed new training strategies for the mixup. \textbf{Optimizing Mixing Ratio} methods used learnable parameter as $\lambda$, obtained reliable mixing ratio by different mixed samples. \textbf{Generating Label} methods generated mixed labels by mixed samples rather than using one-hot labels. \textbf{Attention Score} methods used raw samples' attention maps to obtain the ratio, or used the mixed samples' attention maps to compute the mixing ratio through the score from each sample. \textbf{Saliency Token} methods used each raw sample's saliency maps and divided them into tokens, computed the mixing ratio by the tokens used.
    
    \item \textbf{Sampling.} Some methods focus solely on sample policies to improve the model's performance and ability. They employed other strategies to fix the ratio $\lambda$ or labels, some methods computed all pixels on the mask and fixed $\lambda$, while others set a weight factor for mixed samples.
    
    \item \textbf{Channel Mixup Policies.} Different from samples or labels, channels with lots of high-level features. Manifold Mixup\cite{verma2019manifold} obtained mixed samples by linear interpolation, and Catch up-Mix\cite{kang2024catchupmix} obtained mixed samples by selecting some feature maps for further improving filter capacity.

\end{itemize}
\begin{figure*}[ht]
    \centering
    \includegraphics[width=0.8\linewidth]{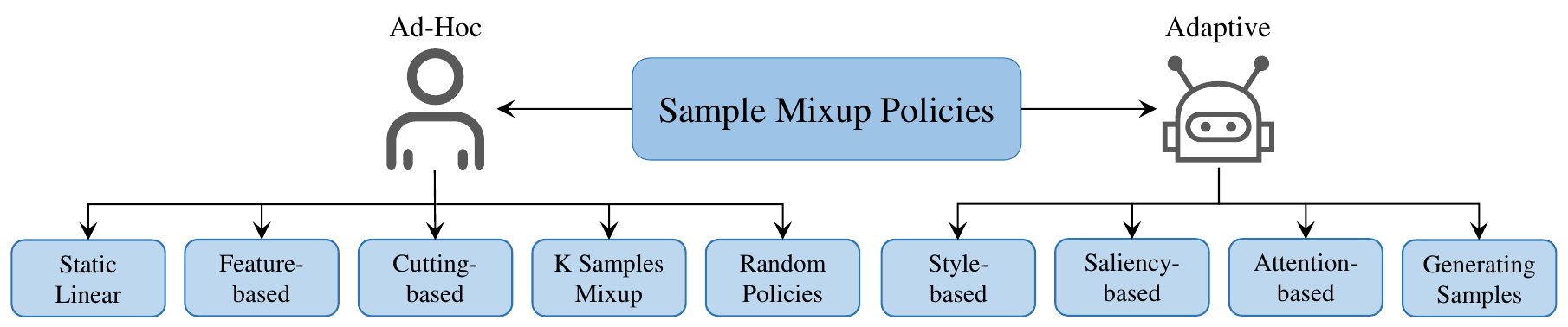}
    \caption{Illustration of sample mixup policies in SL, we divided them into two branches: \textbf{Ad-Hoc} and \textbf{Adaptive}, and divided them into nine detailed types.}
    \label{fig: sample_mixup_policies}
\end{figure*}
\subsection{Main Steps of Mixup Method}
As shown at the top of Fig. \ref{fig: pipeline}, the mixup methods followed these steps: \textbf{(\romannumeral1).} Loading mini-batch raw samples from the training dataset; \textbf{(\romannumeral2).} For some downstream tasks, which include selecting raw samples and retaining reliable samples, some saliency-based or attention-based methods obtained the feature regions or tokens by loading a pre-trained model. Then, define the mixing ratio $\lambda$, which samples from the Beta distribution or the Uniform distribution. \textbf{(\romannumeral3).} After initialization, raw samples were mixed with other samples by sample mixup policies. We illustrate those policies in subsection 3.1. \textbf{(\romannumeral4).} When the mixed samples $\hat{x}$, there were two choices: One was sampling, some methods updated the mixing ratio by mask $\mathcal{M}$ total pixels, some selected mixed samples for retaining more diversity or challenging samples, and some methods redefined the mixing ratio. Another was label mixup policies, and we illustrate those policies in subsection 3.2 and further mining labels $\hat{y}$. \textbf{(\romannumeral5).} The last step was channel mixup policies, the mixed samples $\hat{x}$ encoded by networks and mapping into a high latent space, some methods interpolated each other or selected feature maps for high-dimensional features $\hat{z}$. Then, continue encoding the feature vector for the different tasks, optimizing the network according to different loss functions.

\section{Mixup Methods for CV tasks}
\label{sec:mixup4cv}
\begin{figure*}[t]
    \centering
    \includegraphics[width=0.8\linewidth]{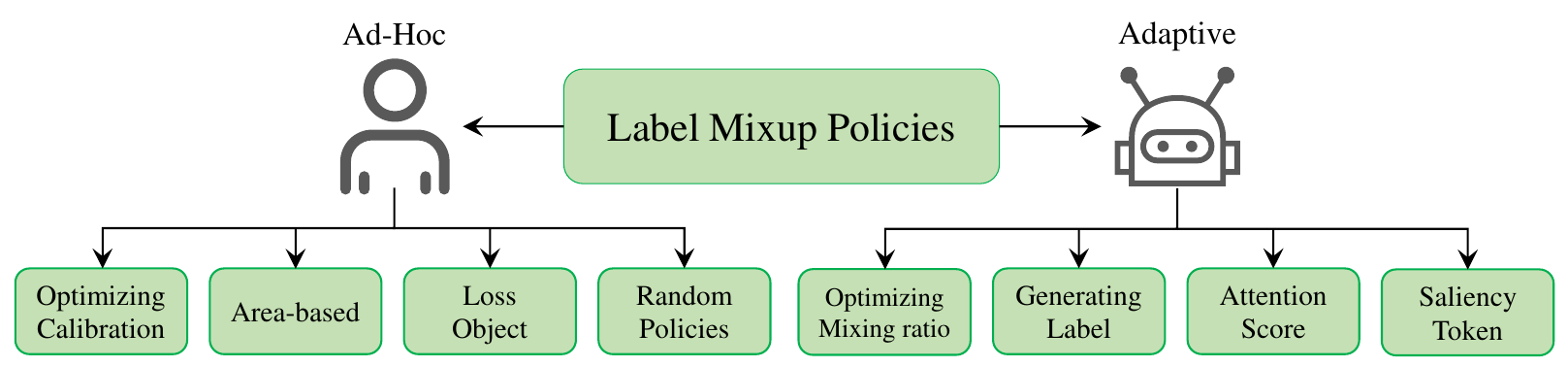}
    \caption{Illustration of label mixup policies in SL, we divided them into two branches: \textbf{Ad-Hoc} and \textbf{Adaptive}, and divided them into eight detailed types.}
    \label{fig: label_mixup_policies}
\end{figure*}

This section focuses on mixup methods used for CV tasks. We reviewed widely of these methods and divided them into four categories: \textbf{(1)} Supervised Learning, \textbf{(2)} Self-Supervised Learning, \textbf{(3)} Semi-Supervised Learning, and \textbf{(4)} some mainstream downstream tasks in the CV: Regression, Long-tail, Segmentation, and Object Detection. 
Fig. \ref{fig: sample_mixup_policies} and Fig. \ref{fig: label_mixup_policies} summarise some mixup methods in SL tasks.

\begin{figure*}[ht]
    \centering
    \includegraphics[width=1.0\linewidth]{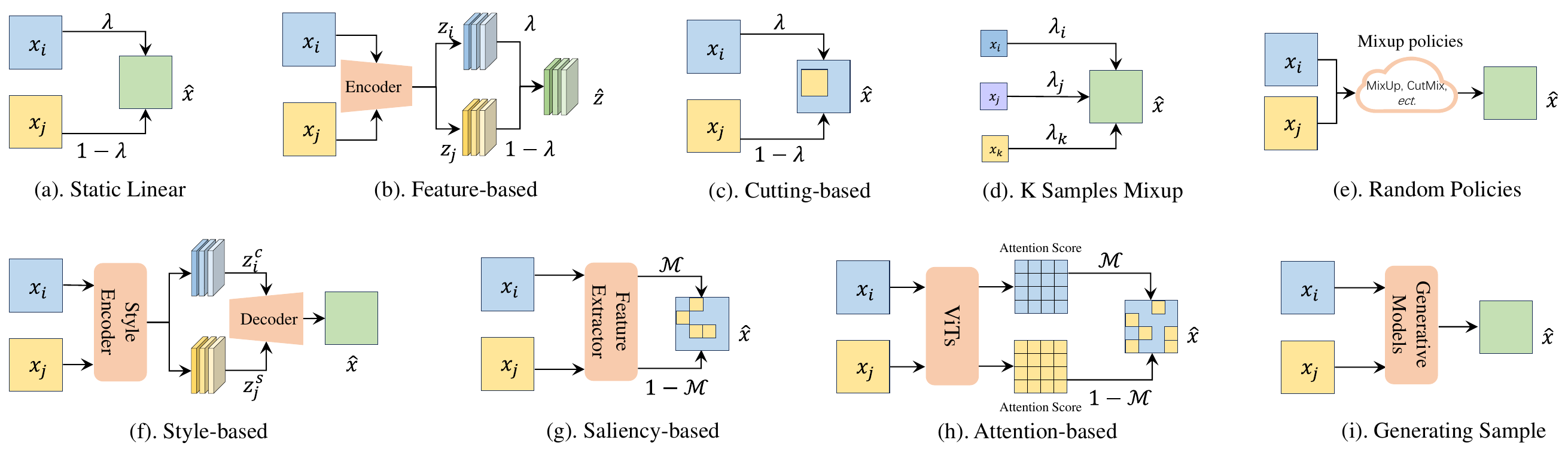}
    \caption{Illustration of simplifying the flow of sample mixup policies in supervised learning (SL). Note that $x_i$ and $x_j$ denote different samples, $z$ denotes feature maps, $\hat{x}$ denotes mixed sample, $\lambda$ was the mixing ratio, and $\mathcal{M}$ denotes mask. (a) \textbf{Static Linear} interpolates samples directly; (b) \textbf{Feature-based} interpolates sample's feature maps; (c) \textbf{Cutting-based} uses cut or resize way mixing samples; (d) \textbf{K Samples Mixup} mixing more than two samples; (e) \textbf{Random Policies} randomly choose mixup policy; (f) \textbf{Style-based} uses a style-transfer encoder to extract content and style and decode the mixed samples; (g) \textbf{Saliency-based} and (h) \textbf{Attention-based} apply a per-train CNN or ViT, mixing samples according to the saliency map or attention score; (i) \textbf{Generating Sample} uses Generative Models to obtain mixed samples.}
    \label{fig: sample_method}
\end{figure*}

\subsection{Sample Mixup Policies in SL}
\label{sec:3.1}
\subsubsection{\textbf{Static Linear}}
\label{sec:3.1.1}
Static Linear methods use ratios to interpolate globally linearly to get mixed samples. Where MixUp\cite{zhang2018mixup} is the seminal work of mixup-based methods, as shown in Eq. \ref{eq:1}, the  efficient, and simple method can bring immediate performance to the model. At the same time, BC\cite{cvpr2018bc} also proposed a similar process to MixUp. Unlike MixUp, which obtained the mixing ratio $\lambda$ from the Beta distribution, BC chose to receive $\lambda$ from the Uniform distribution.
\begin{equation}
    \begin{aligned}
    \label{eq:1}
        \hat{x} = \lambda * x_i + (1-\lambda) * x_j, \ \hat{y} = \lambda * y_i + (1-\lambda) * y_j,
    \end{aligned}
\end{equation}

AdaMixup\cite{guo2018adamix} and Local Mixup\cite{baena2022localmixup} further considered the problem of \emph{``Manifold Intrusion"}. AdaMixup used a learnable ratio to get reliable mixed samples and Local Mixup mining mixed sample pairs ($x_i, x_j$) by weighting the training losses of the mixed samples with $w$. The weight of each mixed sample, depending on the distance between the corresponding points of ($x_i, x_j$), is used to avoid interpolation between samples that are too far away from each other in the input samples.

\subsubsection{\textbf{Feature-based}}
\label{sec:3.1.2}
Feature-based methods transfer mixup methods from pixel-level to latent-level. Manifold Mixup\cite{verma2019manifold} used the ratio $\lambda$ linearly mixing the sample features from the input samples encoded by models:
\begin{equation}
    \label{eq:3}
    \begin{aligned}
        \hat{z} & = \lambda * f_\theta(x_i) + (1-\lambda) * f_\theta(x_j),
    \end{aligned}
\end{equation}
where the $f_\theta(\cdot)$ denotes the model encoder, and the $\hat{z}$ denoted the mixed features. However, PatchUp\cite{faramarzi2020patchup} chose to use the generated mask to mask features in two ways: Hard PatchUp and Soft PatchUp. The hard PatchUp indicates that the mask is binary, and the Soft PatchUp way indicates that the mask is $\lambda$-valued similarly:
\begin{equation}
    \label{eq:4}
    \begin{aligned}
        \left\{\begin{matrix} 
          \mathcal{M}_h = 1,  & \text{Hard\ PatchUp}, \\  
          \mathcal{M}_s = \lambda, & \text{Soft\ PatchUp},
        \end{matrix}\right. 
    \end{aligned}
\end{equation}
\begin{equation}
    \label{eq:5}
    \begin{aligned}
        \hat{z} & = \mathcal{M}_{h/s} \odot f_\theta(x_i) + (1 - \mathcal{M}_{h/s}) \odot f_\theta(x_j),
    \end{aligned}
\end{equation}
where the $\mathcal{M}_{h/s}$ denotes the mask which obtained from Eq. \ref{eq:4}, and the $\odot$ is element-wise multiplication.

MoEx\cite{li2021moex} operated on the training image in feature space, exchanging the moments of the learned features between images and interpolating the target labels. Finally, the mixing constants $\lambda$ are used to combine the Mixup Cross-Entropy (MCE) loss function by Eq. \ref{eq:6}. This process is designed to model the extraction of additional training signals from the moments. This is evidenced by studies\cite{Huang2017ArbitraryST, Li2019PositionalN} demonstrating that moments extracted from instance and positional normalization can approximately capture the style and shape information of an image.
\begin{equation}
    \label{eq:6}
    \begin{aligned}
        \mathcal{L}_{MCE} & = \lambda * \mathcal{L}(z^{(j)}_i, y_i) + (1-\lambda) * \mathcal{L}(z^{(j)}_i, y_j),
    \end{aligned}
\end{equation}
where the $z^{(j)}_i$ denotes the feature representation of $x_i$, which has been injected with the moments of $x_j$. In another way, Catch-up Mix\cite{kang2024catchupmix} found that some CNNs train to produce some powerful filters, and the model will tend to choose their features. Dropping some of the slower-learning filters will limit the performance of the model. Catch-up Mix proposed a filtering module mixing the features learned by poor filters to obtain mixed features, which further improves their capabilities.

\subsubsection{\textbf{Cutting-based}}
\label{sec:3.1.3}
Cutting-based methods used some masks to mix samples. Unlike Cutout drops a patch in samples, CutMix\cite{yun2019cutmix} randomness generated rectangular binary masks $\mathcal{M}$ and mix samples according to Eq. \ref{eq:7}:
\begin{equation}
    \label{eq:7}
    \begin{aligned}
        \hat{x} & = \mathcal{M} \odot x_i + (1 - \mathcal{M}) \odot x_j,
    \end{aligned}
\end{equation}
where the $\mathcal{M} \in \frac{r_w r_h}{WH}$, and $r_w = W \sqrt{1-\lambda}$, $r_h = H \sqrt{1-\lambda}$. GridMix\cite{Baek2021gridmix} and MixedExamples\cite{summers2019mixedexamples} used grid masks similar to GridMask rather than a single patch to obtain mixed samples. SmoothMix\cite{lee2020smoothmix} proposed a smoothing mask for mixing samples with a smooth boundary. Similarly,  StarMix\cite{jin2024starlknet} employed a smooth mask for mixing samples for the vein identification task. SuperGridMix\cite{hammoudi2022superpixelgridcut} used Superpixel segmentation methods to get the mask. MSDA\cite{park2022msda} used MixUp and CutMix to obtain mixed samples. ResizeMix\cite{qin2020resizemix} posits that regardless of the chosen cutting or grinding way, there is a possibility that the source sample's feature may be lost due to randomness. The resize way, however, is designed to retain the complete information of the source sample while also maximizing the feature information of the mixed samples to the greatest extent possible according to Eq. \ref{eq:8}:
\begin{equation}
    \label{eq:8}
    \begin{aligned}
        \hat{x} = \mathcal{P}(T(x_i, \lambda_\tau), x_j),\
        \hat{y} = \lambda * y_i + (1-\lambda) * y_j,
    \end{aligned}
\end{equation}
where $T(\cdot)$ denotes the resizing function and the scale rate $\lambda_\tau$ is sampled from the uniform distribution $\lambda_\tau \sim U(\alpha, \beta)$, where $\alpha$ and $\beta$ denote the lower and upper bound of the range, respectively. $\mathcal{P}(\cdot)$ denotes paste the score sample $x_i$ to the target sample $x_j$. The $\lambda$ is defined by the size ratio of the patch and target sample.

Unlike mixed in pixel-level. FMix\cite{harris2020fmix} argued that effective feature is often continuous in an image, so the samples are transformed from the RGB channel to the Fourier space using the Fourier transform, sampling the high\&low-frequency information to generate binary masks. Pani MixUp\cite{sun2024panimixup} interpolated mixing by GNNs to build relationships between sample patches. Specifically, patch sets are constructed first, then k-nearest sets are further built using the similarity of patches, and similar patches are linearly interpolated to get the mixed samples.

\begin{figure}[ht]
    \centering
    \includegraphics[scale=0.15]{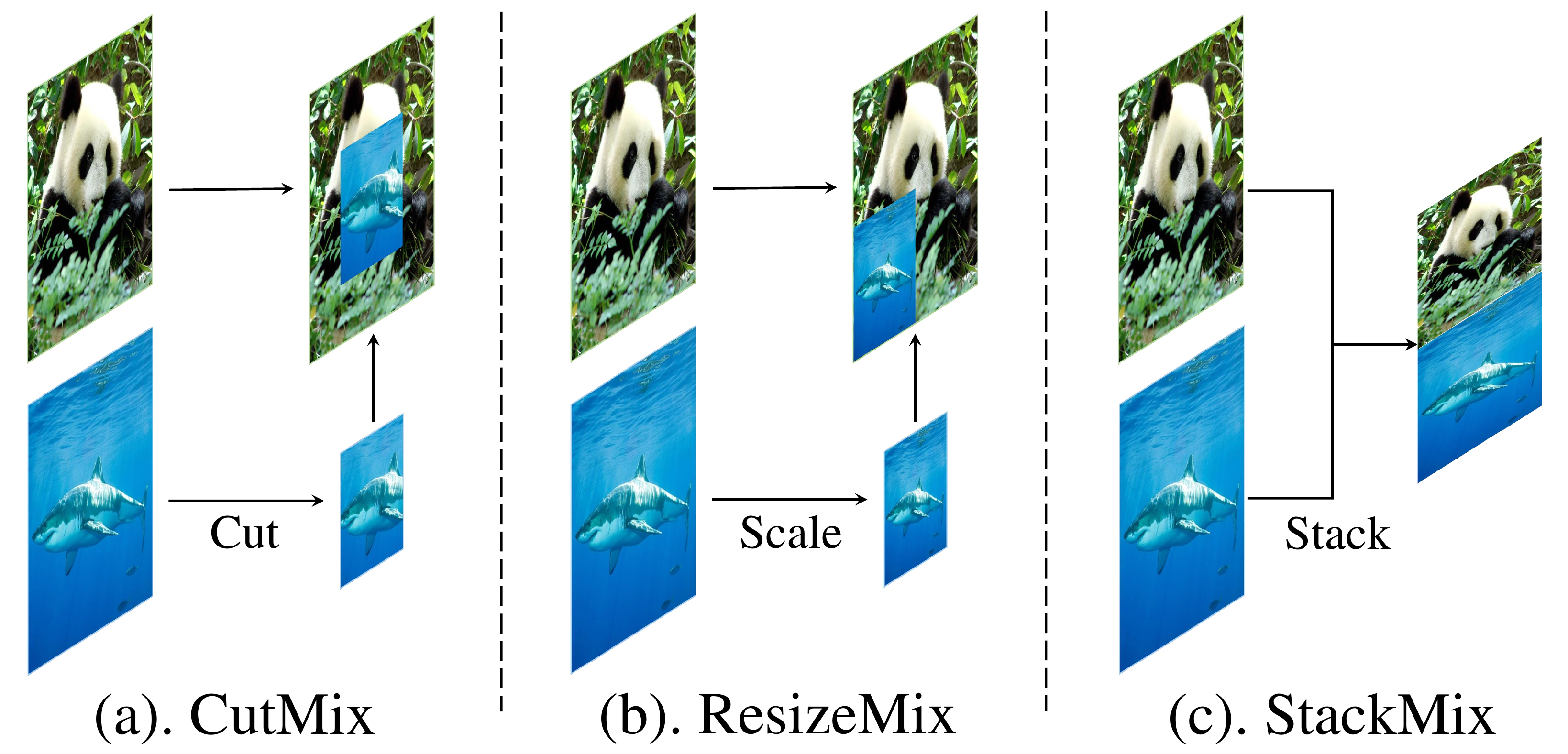}
    \caption{The process of obtaining the mixed samples based on (a). CutMix, (b). ResizeMix \& (c). StakcMix method.}
    \label{fig: cut_resize_satck}
\end{figure}

Moreover, instead of using some mask-based ways, some works, \emph{e.g.} StackMix\cite{chen2022stackmix}, RICAP\cite{takahashi2019ricap}, YOCO\cite{han2022yoco} stacking samples directly. Some cutting-based mixup methods are shown in Fig. \ref{fig: cut_resize_satck}.

\subsubsection{\textbf{K Samples Mixup}}
\label{sec:3.1.4}
Lots of mixup methods focus on two samples mixing, lacking analysis and design of the performance impact of mixing more than two samples on model performance. $k$-mixup\cite{greenewald2021kmixup} created more mixed samples by interpolating the samples under the Wasserstein metric while perturbing $k$-batches in terms of the other K samples. DCutMix\cite{jeong2021dcutmix} obtained mixed samples by applying CutMix multiple times. Experimental analyses of multi-mixed samples showed that multi-sample mixing of Landspace guided the model towards the widest (flattest) and deepest local minima. MixMo\cite{rame2021mixmo} proposed a multi-input, multi-output model (MIMO model) that fuses MixUp with an integrated MIMO model, intending to find the best mixing module to handle multiple sub-networks.

Some work such as RICAP, Cut-thumbnail\cite{xie2021Cut-thumbnail}, and Mosaic\cite{redmon2016yolo}  obtained mixed samples to improve the performance of models by stacking or cutting multi-samples.

\subsubsection{\textbf{Random Policies}}
\label{sec:3.1.5}
RandomMix\cite{liu2024randomix} combined MixUp, CutMix, ResizeMix, and FMix into a policy set, randomly choosing one of them and their weight to get mixed samples. Furthermore, AugRMixAT\cite{liu2022augrmixat}, add additional policies, \emph{e.g.} Augmentation methods and Adversarial Attack aim to get a more robust model.

\subsubsection{\textbf{Style-based}}
\label{sec:3.1.6}
\textbf{Static ways.} AugMix\cite{hendrycks2019augmix} aimed to improve model robustness and uncertainty estimation. Using randomness and multiple augmentation methods to mix multiple samples to obtain a mixed sample, the model is trained using the Jensen-Shannon loss of consistency according to Eq. \ref{eq:2}:
\begin{equation}
    \label{eq:2}
    \begin{aligned}
        \hat{x} & = \lambda*(\textrm{w}_i*\mathcal{A}(x_i) + \textrm{w}_j*\mathcal{A}(x_j)) + (1-\lambda)*x_z,
    \end{aligned}
\end{equation}
where $\textrm{w}_i$, and $\textrm{w}_j$ are randomly generated weights from the Dirichlet distribution. PixMix\cite{hendrycks2022pixmix} and IPMix\cite{huang2024ipmix} further improved AugMix by mixing additional fractals datasets with the samples, mixing with a $\mathcal{A}(x)$, and repeating this several times to get the final mixed samples. This is used to tackle some Out-of-Distribution (OOD) scenarios and improve the robustness of the model. DJMix\cite{hataya2020djmix} also extended AugMix by using a mixture of the input image and the discretized image, which is intended to improve the robustness of CNNs.

\textbf{Dynamic ways.} StyleMix\cite{cvpr2021stylemix} proposed the concept of differentiating and mixing the content and style features of input image pairs to create more abundant and robust samples. 
Specifically, StyleMix employed the adaptive instance normalization layer (AdaIN)\cite{Huang2017ArbitraryST} to exchange styles for images $x_i$ and $x_j$ (Eq. \ref{eq:9}). To obtain a new mixed image $\hat{x}$, StyleMix linearly interpolates feature maps with a content ratio $\lambda_c$ and a style ratio $\lambda_s$ by Eq. \ref{eq:10}. 
\begin{equation}
    \label{eq:9}
    \begin{aligned}
       \hat{z}_{i,j}  = \sigma(z_j)(\frac{z_i-\mu(z_i)}{\sigma(z_i)})+\mu(z_j),
    \end{aligned}
\end{equation}
\begin{equation}
    \label{eq:10}
    \begin{aligned}
       \hat{x}  & = \text{De}( \nu(z_i)+( 1 - \lambda_c - \lambda_s + t))z_j \\
       & +(\lambda_c - \nu ) \hat{z}_{i,j} + (\lambda_s - \nu )\hat{z}_{j,i},
    \end{aligned}
\end{equation}
where $\hat{z}_{i,j}$ has the content component of $x_i$ and the style of $x_j$, $\text{De}(\cdot)$ is the pre-trained style decoder and $\nu \in [\max(0,\lambda_c + \lambda_s - 1),\min(\lambda_c, \lambda_s)]$ is a variable. MixStyle\cite{zhou2021mixstyle} was inspired by AdaIN, which effectively combines style transfer and mixup across source domains. In contrast to StyleMix\cite{cvpr2021stylemix}, MixStyle does not include a decoder for image generation attached, but rather. Instead, it perturbs the style information of source domain training instances to implicitly synthesize novel domains. Specifically, MixStyle computes the mixed
feature statistics $\hat{z}_{i,j}^{\sigma}$ and $\hat{z}_{i,j}^{\mu}$ by Eq. \ref{eq:11} and utilizes them to the style-normalized $x$ by Eq. \ref{eq:12}:
\begin{equation}
    \label{eq:11}
    \begin{aligned}
        \hat{z}_{i,j}^{\sigma} & = \lambda* \sigma(x_i) + (1-\lambda) * \sigma(x_j), \\
        \hat{z}_{i,j}^{\mu} & = \lambda* \mu(x_i) + (1-\lambda) * \mu(x_j),
    \end{aligned}
\end{equation}
\begin{equation}
    \label{eq:12}
    \begin{aligned}
       \textrm{MixStyle}(x) = \hat{z}_{i,j}^{\sigma}\frac{x-\mu(x)}{\sigma(x)} + \hat{z}_{i,j}^{\mu},
    \end{aligned}
\end{equation}
where $x=[x_i,x_j]$ is sampled from domains $i$ and domains $j$.

AlignMixup\cite{2021alignmix} obtained the mixed feature maps by using assignment matrix $M_R$, which Sinkhorn-Knopp obtains \cite{knight2008sinkhorn} on similarity matrix $e^{-M/\tau}$. $z_j$ is aligned to $z_i$ according to $M_R$, giving rise to $z^\prime_j$=$M_R(z_j)$, then interpolate between $z_i$, $z^\prime_j$. MultiMix\cite{venkataramanan2022multimix} obtained augmented feature maps by additional attention and $\lambda$ element-wise multiplication of the matrix $M_a$, and obtained mixed feature maps by Eq. \ref{eq:13}:
\begin{equation}
    \label{eq:13}
    \begin{aligned}
       \hat{z} & = \textrm{norm}(M_a(\textrm{A}_i, \lambda)) \odot z_i  + \textrm{norm}(M_a(\textrm{A}_j, 1-\lambda) \odot z_j,
    \end{aligned}
\end{equation}
where $\textrm{norm}(\cdot)$ denotes normalization, and $\textrm{A}$ is obtained from $z$ after GAP (Global Average Pooling) or $z$ after CAM (Class Activation Maps)\cite{selvaraju2017grad} calculation.

\subsubsection{\textbf{Saliency-based}}
\label{sec:3.1.7}
Saliency-based methods to maximize features in the mixed samples use additional feature extractors, or backward gradient obtained saliency maps\cite{simonyan2014saliency} to locate feature locations to guide mask $\mathcal{M}$ generation. 

\textbf{Additional Feature Extractor.} SaliencyMix\cite{uddin2020saliencymix} used a feature extractor to select the peak salient region from the saliency map in the source image and mix the source patch with the target image. Attentive-CutMix\cite{icassp2020attentive} use a pre-trained classification model $f^\prime _\theta(\cdot)$ as a teacher model aims to generate a mask of source sample feature patches, first obtain a heatmap (generally a 7$\times$7 grid map) of the source image, then select the top-$N$ patches from this 7$\times$7 grid as attentive region patches to cut from the source image. Similarly, FocusMix\cite{kim2020focusmix} used CAM to generate masks for mixed samples. AttributeMix\cite{li2020attributemix} build a $K$-Dictionary to get a different feature on each sample and mix for the fine-grained classification task. GraSalMix\cite{hong2023gradsalmix} was found to locate feature regions in the samples more accurately based on network gradient than CAM and used cutting-based mixing for the $z$ obtained for each network stage. RecursiveMix\cite{yang2022recursivemix} advised to keep historical samples for mixing. Use regions of interest (RoI)\cite{he2017maskrcnn} and ResizeMix for mixed samples to be remixed with the raw samples.

PuzzleMix\cite{kim2020puzzle} first fed the samples into the model to get the gradient, then backward the gradient to get the saliency map of the samples. It divides the samples into multiple patches based on the saliency map and generates masks with a ratio $\lambda$ corresponding to the number of patches. To prevent the problem of feature loss due to two sample features at the same location, Kim. \emph{et al.} proposed an optimizing transport algorithm to maximize the retention of the feature of the samples. Similarly, they proposed Co-Mix\cite{kim2020co}. Instead of being limited to mixing between pairs of samples ($x_i, x_j$), the mixup process is guided by finding multiple samples in a mini-batch for mixing and turning the search for the optimal mixup method into an optimization problem through multiple mixed samples. To improve the diversity of the mixed samples, a regularization loss is introduced to penalize the generation of overly similar images, making the mixed samples more diverse. GuidedMixup\cite{kang2023guidedmixup} proposed a greedy pairing algorithm to find the best sets with a large distance between salient regions among the mini-batch images, and the token level was transferred to pixel-level to maximize saliency. The mixing ratio is adjusted by pixel for more detail and smoother mixing. SuperMix\cite{wu2023supermix} used a per-trained model as a teacher model sampling and generating a set of masks from Dirichlet distribution and used Newton method optimized masks. LGCOAMix\cite{dornaika2023lgcoamix} used the Superpixel method to obtain mixed samples and re-calculate $\lambda$ by the feature maps. LGCOAMix also used a Superpixel pooling and self-attention module to get a local classification loss and Superpixel contrastive loss.

\textbf{End-to-End way.} Using saliency or gradient information often incurs additional time overhead, AutoMix\cite{liu2022automix} used an end-to-end method to get a mixed sample that is optimal in terms of both time overhead and network performance. AutoMix proposed a lightweight generator, Mixblock, to generate masks according to the Eq. \ref{eq:14} automatically:
\begin{equation}
\label{eq:14}
    \begin{aligned}
        \hat{x} = \mathcal{M}_\theta(z^l_{i,\lambda},z^l_{j,1-\lambda})\odot x_i + (1 - \mathcal{M}_\theta(z^l_{i,\lambda},z^l_{j,1-\lambda}))\odot x_j,
    \end{aligned}
\end{equation}
where $z^l_{i, \lambda}$ denotes $\lambda$ embedded feature maps at $l$-th layer. $\mathcal{M}_\theta$ denotes the Mixblock module with learnable parameters $\theta$. SAMix\cite{li2021samix} further improves Mixblock and extends AutoMix in supervised learning tasks to self-supervised tasks. TransformMix\cite{cheung2024transformmix} used a teacher model obtained samples CAMs and mixing by a new mixing module. AdAutoMix\cite{qin2023adautomix} used adversarial training to obtain more challenging samples and improve the classification model's ability. It introduced a new Mixblock that enables mixing $N$ samples instead of two.

\subsubsection{\textbf{Attention-based}}
\label{sec:3.1.8}
Inspired by Vision Transformer (ViT)\cite{iclr2021vit}, attention score maps also can be metrics such as CAM or gradient. TransMix\cite{chen2022transmix} re-calculated the $\lambda$ by the area attention score of each sample. TokenMixup\cite{choi2022tokenmixup} and TokenMix\cite{liu2022tokenmix} selected the top-$N$ high-scoring tokens from the sample's attention map to use as masks, and $N= \lambda \mathbb{N}$. TokenMixup used a ScoreNet to select samples mixing at the input step.

ScoreMix\cite{stegmuller2023scorenet} and SMMix\cite{chen2023smmix} use the maximum and minimum attention regions of the sample attention map to mixing samples. These methods cut the target sample maximum region and paste it into the source sample minimum region to retain more feature information.

\subsubsection{\textbf{Generating Samples}}
\label{sec:3.1.9}
AAE\cite{liu2018aae} obtained mixed samples by using the GANs model, first generating mixed feature maps similar to Manifold Mixup, and then using a Decoder to generate mixed samples, $\hat{z} = \lambda * \text{En}(x_i) + (1-\lambda) * \text{En}(x_j)$, $\hat{x} = \text{De}(\hat{z})$. And like GANs, AAE used a Discriminator to get adversarial loss and total loss according to Eq. \ref{eq:15}:
\begin{equation}
    \label{eq:15}
    \begin{aligned}
    \mathcal{L}_{total} & = \mathcal{L}_{rec}(\vert\vert x - \hat{x} \vert\vert_{2}) \\ & + \mathcal{L}_{adv} (\log \text{Dis}(\hat{z}) + (1-(\log \text{Dis}(\hat{z})))
    ),
    \end{aligned}
\end{equation}
where $\mathcal{L}_{rec}$ denotes the reconstruction loss, and $\mathcal{L}_{adv}$ denotes the adversarial loss.
Similarly, AMR\cite{beckham2019amr}, ACAI\cite{berthelot2018acai}, AutoMixup\cite{zhu2020automixup}, and VarMixup\cite{mangla2020varmixup} all used generative models such as AE, VAE\cite{kingma2013vae}, and GAN to interpolate feature maps in the latent space, then pass through a decoder to get the generated mixed samples. Inspired by the Text-to-Image model, DiffuseMix\cite{islam2024diffusemix} used a Stable Diffusion-based multi-language model named InstructPix2Pix\cite{brooks2023instructpix2pix}, which obtained the initial augment samples using the defined prompt and then obtained the final mixed samples using YOCO and PixMix methods.  

\subsection{Label Mixup Policies in SL}
\subsubsection{\textbf{Optimizing Calibration}}
\label{sec:3.2.1}
Calibration is a metric that measures how well a model's confidence aligns with its accuracy. A model should neither be overconfident nor underconfident, as this can lead to incorrect decisions. The Expected Calibration Error (ECE) is a key metric for evaluating model calibration. Manifold Mixup found that MixUp significantly improves network confidence, but it also noted that overconfidence is not desirable.
CAMixup\cite{wen2020camix} found that a simple application of MixUp with Cross-Entropy (CE) increases the ECE of the model, which is contrary to current general intuition. Wen \emph{et al.} used Label Smoothing for experiments to verify that it is a labeling issue. CAMixup adjusted the mixing ratio based on the difference between the average accuracy and confidence of the class classification $\lambda$. 

Normally, the model is expected to be more confident in predicting simpler classes. For harder classes, the model is encouraged to be less confident.
RankMixup\cite{noh2023rankmixup} proposed a Mixup-based Ranking Loss (MRL):
\begin{equation}
    \label{eq:16}
    \begin{aligned}
        \mathcal{L}_{MRL} = \max(0, \max \hat{p}_i^k - \max p_i^k + m),
    \end{aligned}
\end{equation}
where $\hat{p}_i^k$ is a predicted probability of the class $k$ for the mixed sample $\hat{x}_i$, margin $m$ determines acceptable discrepancies between the confidences of raw and augmented samples. It encourages the confidence level of the mixed samples to be lower than that of the raw samples to maintain the ranking relationship. The expectation is that higher confidence values favor larger mixed samples with $\lambda$ so that the confidence values and the order of the mixing ratio are proportional to each other. SmoothMixup\cite{jeong2021smoothmixup} using the inherent robustness of smoothing classifiers to identify semantically non-categorical samples and to improve the performance of classifiers.

\subsubsection{\textbf{Area-based}}
\label{sec:3.2.2}
RICAP\cite{takahashi2019ricap} uses the area of each sample in the mixed sample to calculate the $\lambda$. Some hand-crafted methods like CutMix, Smooth, GridMix, and StackMix, \emph{ect.} use the mask area to calculate the $\lambda$. RecursiveMix similarly uses the history region to calculate the $\lambda_i = \sum \mathcal{M}_i$ for obtaining the mixed labels.

\subsubsection{\textbf{Loss Object}}
\label{sec:3.2.3}
Before describing the relevant methods, it is important to understand the loss function in the mixup task:
\begin{equation}
    \label{eq:17}
    \begin{aligned}
        \mathcal{L}_{MCE} = \lambda \mathcal{L}_{CE}(f_\theta(\hat{x}), y_i) + (1-\lambda) \mathcal{L}_{CE}(f_\theta(\hat{x}), y_j). 
    \end{aligned}
\end{equation}
In the mixup task, since there are 2 or more samples and labels, Softmax will weaken the information of one of the classes. When one of the $\lambda$ is too high, it will produce high entropy, but the confidence level may be low, seriously affecting the prediction of a single class. Liu \emph{et. al} proposed DecoupledMix\cite{liu2022decoupledmix}, an effective loss function suitable for mixup methods, with a decoupled regularisation, which can adaptively use ``hard-sample" to mine features. Specifically, DecoupledMix decouples the two classes in mixup, removes its probability $p$ in the denominator when the final prediction is softmax, and uses two-hot to do the regularization according to Eq. \ref{eq:18}:
\begin{equation}
    \label{eq:18}
    \begin{aligned}
    \phi\left(z_{(i, j)}\right)^{a, b} & = \frac{\exp (z_{(i, j)}^{a} )}{\bcancel{\exp (z_{(i, j)}^{b} )}+\sum_{c \neq b} \exp (z_{(i, j)}^{c} )}, \\
    \mathcal{L}_{DM} & = -y_{[i, j]}^{T} \log (\phi (z_{(i, j)} ) ) y_{[i, j]},
    \end{aligned}
\end{equation}
where $\phi(\cdot)$ is the proposed decoupled Softmax, $(a,b)$ denotes the classes of the mixed samples, and $y_{[i, j]}$ is two-hot label encoding.

MixupE\cite{zou2023mixupe} analyzed mainstream mixup methods and found that mixup implicitly regularizes infinitely many directional derivatives of all orders, contrary to the popular belief that this can be replaced by second-order regularization. Therefore, MixupE proposed an improved method by enhancing the effect of mixup's implicit regularization of directional derivatives.

\subsubsection{\textbf{Random Policies}}
\label{sec:3.2.4}
To address the problem of slow convergence of mixup and difficulty in selecting $\alpha$, Yu \emph{et al.} proposed the mixup without hesitation (mWh)\cite{yu2021mwh} method, which accelerates mixup by periodically switching off the mixup way. mWh analyzed and demonstrated through experiments that mixup is effective in the early epochs, while it may be detrimental to the learning method in the later epochs. Therefore, MixUp is gradually replaced by basic data augmentation to shift from exploration to exploitation. RegMixup\cite{pinto2022regmixup} similarly analyzed the effect of $\alpha$ on mixup when training models and found that different $\alpha$ values have an effect on the entropy resulting from the model, and also found that the MCE loss brings complementary effects when it is used in combination with the one-hot CE loss.

\subsubsection{\textbf{Optimizing Mixing Ratio}}
\label{sec:3.2.5}
AdaMixup\cite{guo2018adamix} argued that simple interpolation to obtain mixed samples is the same as the raw samples of other classes, causing \emph{``Manifold Intrusion"}. Experiments found that the model performance is most affected at $\lambda$=0.5. Thus, AdaMixup proposed a method to learn $\lambda$. Two neural networks are introduced to generate the mixing ratio $\lambda$ and to judge whether the mixed samples cause \emph{``Manifold Intrusion"}. MetaMixup\cite{mai2021metamixup} used Meta-Learning to optimize the mixup. The aim is a bi-fold optimization: the model $f_\theta(\cdot)$ and the mixing ratio $\lambda$. In this case, the Train-State trained the model with mixed samples using the learned $\lambda$. The obtained gradient optimizes the parameters $\theta$ of the model and also passes them to Meta-State, which uses the validation set to obtain the gradient to modify the mixing ratio $\lambda$.

LUMix\cite{sun2022lumix} LUMix differs from previous methods to address label noise in the mixup process. It proposes two approaches: detecting significant targets in the input from the prediction and adding random samples from a uniform distribution to the label distribution to simulate the label uncertainty, the final ratio is $\lambda$ = $\lambda_{0}*(1 - \nu_s- \nu_r) + \lambda_{s}*\nu_s + \lambda_{r}*\nu_r$. $\lambda_{0}$ is set manually, $\lambda_{s}$ is calculated from the predictive distribution according to Eq. \ref{eq:19}, and $\lambda_{r}$ is sampled from $\text{Beta}(\alpha, \alpha)$.
\begin{equation}
    \label{eq:19}
    \begin{aligned}
    \hat{p}_{i} & =\frac{e^{p_i}}{\sum_{c=1}^{k} e^{p_c}}, \ \lambda_{s} =\frac{\hat{p}_i}{\hat{p}_i + \hat{p}_j},
    \end{aligned}
\end{equation}
SUMix\cite{qin2024sumix} is similar to LUMix in that it is assumed that it can be modeled in prediction to get a proper $\lambda$. In addition, SUMix applied the idea of metric learning to regularize the loss function by capturing the uncertainty in the samples.

\subsubsection{\textbf{Generating Label}}
\label{sec:3.2.6}
Simply interpolating the samples linearly may lead to the problem of \emph{``Manifold Intrusion"}, GenLabel\cite{sohn2022genlabel} proposed to relabel the mixed samples. Firstly, learning the class-conditional data distribution using generative models for each class $c$, denoted as $p_c(x)$. Then, based on the likelihood $p_c(\hat{x})$ of the mixed samples extracted from each class $c$, the mixed samples are re-generated with the label $y^{gen}$ after obtaining the similarity according to Eq. \ref{eq:20}:
\begin{equation}
    \label{eq:20}
    \begin{aligned}
    y^{gen} & = \sum_{c=1}^{k} \frac{p_{c}(\hat{x})}{\sum_{c^{\prime}=1}^{k} p_{c^{\prime}}(\hat{x}) }e_c,
    \end{aligned}
\end{equation}
where $p_c(\hat{x})$ denotes density estimated by generative model for input feature $x \in \mathbb{X}$, conditioned on class $c \in [k]$, $k$ is the total classes.

\subsubsection{\textbf{Attention Score}}
\label{sec:3.2.7}
When using ViTs for classification tasks, a class token is often added, and the final prediction is based on this class token after the models have been correlated with each other. However, the feature used this way is limited since only the last class token is used after many feature tokens are generated in the raw image. Therefore, the supervision effect is significantly reduced. Token Labeling\cite{jiang2021tokenlabeling} labeled all the tokens in the ViTs for the classification task. Compared to the original loss function, the loss function of Token Labeling adds the classification loss of tokens.
\begin{equation}
    \label{eq:21}
    \begin{aligned}
    \mathcal{L}_{total} & = \mathcal{L}_{CE}(p^{cls}, y) + \textrm{w} * \frac{1}{\mathbb{N}} \sum^{\mathbb{N}}_{i=1}\mathcal{L}_{CE}(\textrm{token}_i, y_i),
    \end{aligned}
\end{equation}
where the $p^{cls}$ denotes the predication of the class token, and $\textrm{token}_i$ denotes the predication of the $i$-index token. TokenMix\cite{liu2022tokenmix} and TokenMixup\cite{choi2022tokenmixup} used the raw samples' attention score for calculating the $\lambda$ of classes. Mixpro\cite{zhao2023mixpro} argued that the attention scores obtained by the model in the early stages are inaccurate and give incorrect information, so they proposed a progressive factor. Mixpro combines with the region and attention scores to calculate the $lambda$. Differ from MixPro, TL-Align\cite{xiao2023tl-align} used the mixed samples in the ViTs when each layer is through $\bm{M}_Q * \bm{M}_K^T$ to calculate the attention maps, and the attention maps are combined with the labels. TL-Align's labels are resized into a matrix first by tokens to get the mixed labels matrix, and then multiplied with attention maps and residuals to get the aligned labels, and then each layer in the ViTs makes the labels aligned to get the final mixed labels.

\subsubsection{\textbf{Saliency Token}}
\label{sec:3.2.8}
SnapMix\cite{huang2020snapmix} used CAM to get feature maps of the raw samples and then mixed them with randomly generated masks, the ratio $\lambda$ calculated based on the feature information in the mask region, and the whole region ratio, which results in the two samples sum of ratio is not 1 in SnapMix. Saliency Grafting\cite{park2021saliencygraft} extracted the saliency maps in the samples by pre-trained CNNs, and the obtained saliency maps are subjected to a softmax function and threshold decision to get a preliminary binary mask, the threshold is obtained from the mean value of the saliency maps. Then the final mask $\mathcal{M}$ is constructed by taking the Hadamard product of $S^{\prime\prime}_i$ and a region-wise i.i.d. random Bernoulli matrix of same dimensions $P\sim \text{Bern}(p_B)$: $\mathcal{M} = P*S^{\prime\prime}_i$.

\subsection{Self-Supervised Learning}
\label{sec:ssl}

\begin{figure}[ht]
    \centering
    \includegraphics[scale=0.16]{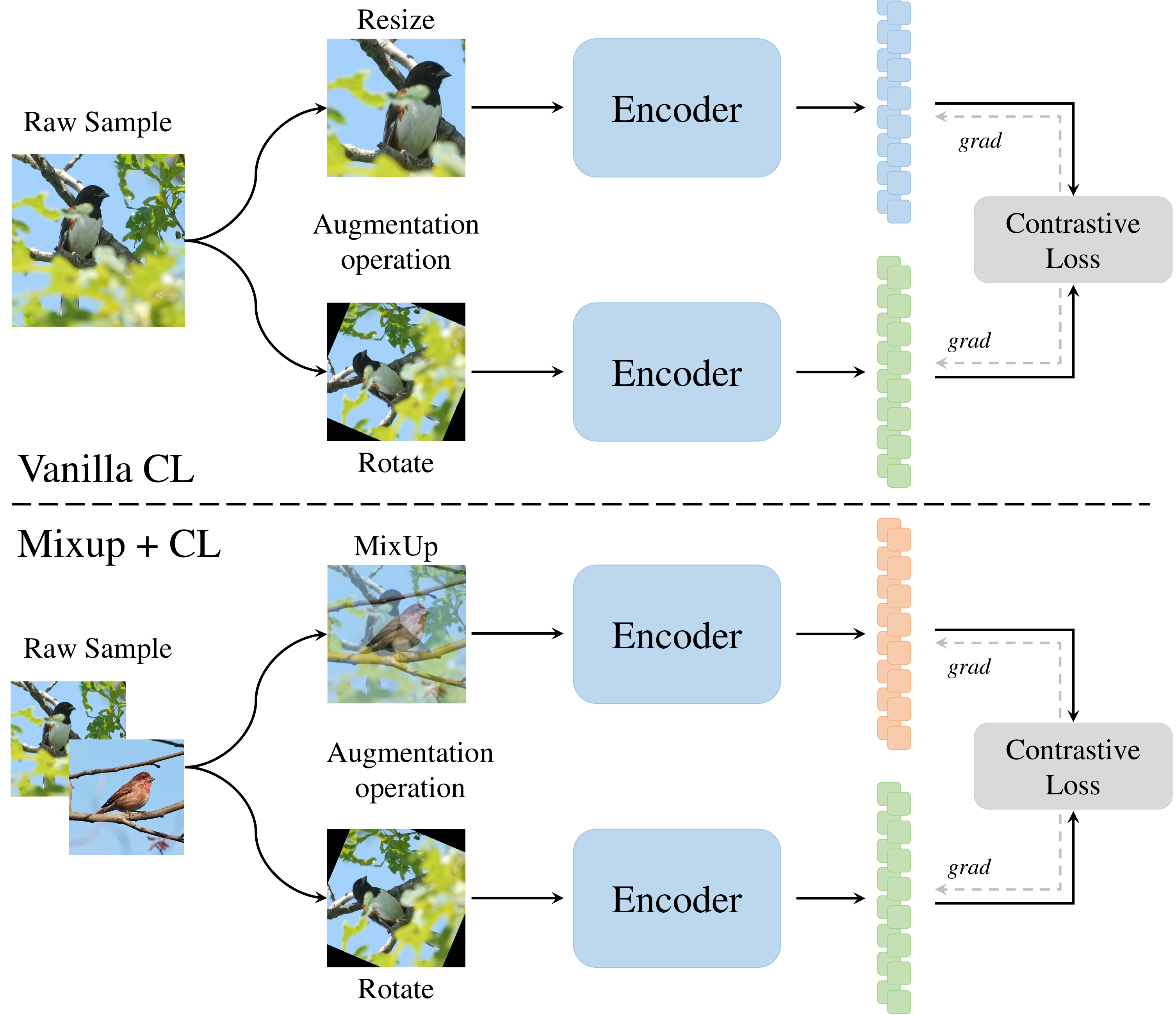}
    \caption{The different framework between the Vanilla CL method and the CL method with mixup. Vanilla CL methods used two different $\mathcal{A}(\cdot)$ for the same raw samples as ``positive'' samples, Mixup + CL used mixup and augmentation, aiming to obtain ``semi-positive'' samples. }
    \label{fig: cl_method}
\end{figure}
\subsubsection{\textbf{Contrastive Learning}}
Contrastive learning (CL) has emerged as a prominent training paradigm in SSL, where the objective is to map similar ``positive" samples into a representation space proximate to an ``anchor" while mapping disparate ``negative" samples into regions more distant. Mixup methods have been extensively employed in CL to generate diverse or challenging samples. Fig. \ref{fig: cl_method} shows the pipeline of the vanilla CL method and CL with mixup.

\textbf{Loss Object Improvement.} In terms of improvement, MixCo\cite{kim2020mixco} argued that mixed samples with features of semi-positive samples contribute to learning better representations, and using CL with the positive and negative samples and targets of mixed samples can alleviate the instance discrimination problem by allowing the model to learn the implicit relationship between the positives and negatives, as shown in Eq. \ref{eq:22}:
\begin{equation}
    \label{eq:22}
    \begin{aligned}
    \mathcal{L}_{MixCo} & = - \sum_{i=1}^{N}(\lambda \cdot \log(\frac{\exp(\frac{\hat{z}_i \cdot z_i}{\tau})}{\textstyle \sum_{n=1}^{N}\exp(\frac{\hat{z}_i \cdot z_n}{\tau})})) \\
    & + ((1-\lambda) \cdot \log(\frac{\exp(\frac{\hat{z}_i \cdot z_k}{\tau})}{\textstyle \sum_{n=1}^{N}\exp(\frac{\hat{z}_i \cdot z_n}{\tau})})),
    \end{aligned}
\end{equation}
where the $\tau$ denotes the softening temperature for similarity logits, $N$ is the number of mixed samples. Different from MixCo, \emph{i}-Mix\cite{lee2020i-mix} proposed to insert virtual labels in batch size and mixed samples and to insert corresponding virtual labels in the input-level and label-level, respectively. Transforming unsupervised loss functions to supervised losses (\emph{e.g.}, CE losses) according to Eq. \ref{eq:23} and Eq. \ref{eq:24}:
\begin{equation}
    \label{eq:23}
    \begin{aligned}
    \mathcal{L}_{SimCLR} & = - \log \frac{\exp(s(\hat{z}_i, z_{(N+i) \ \textrm{mod} \ 2N})/\tau)}{\textstyle \sum_{k=1, k\neq i}^{2N}\exp(s(\hat{z}_i , z_k)/\tau)},
    \end{aligned}
\end{equation}
\begin{equation}
    \label{eq:24}
    \begin{aligned}
    \mathcal{L}_{N-pair} & = - \sum_{n=1}^{N} \textrm{w}_{i,n} \log \frac{\exp(s(\hat{z}_i, z_n)/\tau)}{\textstyle \sum_{k=1}^{N}\exp(s(\hat{z}_i , z_k)/\tau)},
    \end{aligned}
\end{equation}
where $N$ is the batch size. Similarly to the \emph{i}-Mix concept, MCL\cite{wickstrom2022mcl} put forth a CL framework based on the tenets of label smoothing. This framework employed a novel contrastive loss function intending to enhance the classification efficacy of clinical medicine time series data through the use mixup for the generation of new samples.

SAMix\cite{li2021samix} proposed improved infoNCE, $\mathcal{L}_{+}$ and $\mathcal{L}_{-}$, for mixup-based SSL. $\mathcal{L}_{+}$ is called the local term since features of another class are added to the original infoNCE loss; $\mathcal{L}_{-}$ is called the global term, and is the mixed infoNCE by Eq. \ref{eq:25}:
\begin{equation}
    \label{eq:25}
    \begin{aligned}
    \mathcal{L}_{+} & = - \lambda \log \hat{p}_{i} - (1 - \lambda) \log \hat{p}_{j}, \\
    \hat{p}_{i} & = \frac{\exp(\hat{z} \ z_i/\tau)}{\exp(\hat{z} \ z_i/\tau) + \exp(\hat{z} \ z_j/\tau)}.
    \end{aligned}
\end{equation}

Since SimSiam\cite{chen2021simsiam} only relies on pairs from the same sample's random augmentation with large same information between pairs, it may reduce discriminative power for images with large intra-class variations. Guo \emph{et al.} proposed MixSiam\cite{guo2021mixsiam}. It maximizes the significance of two augmented samples by inputting to the encoder and obtaining maximum significance representation via the Aggregation operation. Another branch used mixed samples as inputs and expected model predictions to be close to the discriminant representation. Thus, Guo \emph{et al.} argued that the model can access more variable samples in a sample and predict their most significant feature representations.


\textbf{Hard Sample Mining.} In CL, training with hard samples can further improve model performance. MoCHi\cite{kalantidis2020mochi} enhances model performance by encoding raw samples to obtain representations, then computing and ranking the feature similarity with the positive samples, and taking $N$ features with low similarity to be Manifold Mixup. This process yields more challenging negative samples. Similarly, Co-Tuning\cite{zhang2021co-tuneing} proposed a contrastive regularization way for generating sample pairs to mine hard sample pairs. BSIM\cite{chu2020bsim} improved classification by measuring the joint similarity between two raw samples and mixed samples (\emph{i.e.}, semi-positive pairs.) Chu \emph{et al.} argued that learning joint similarity helps to improve performance when features are more uniformly distributed in the latent space. FT\cite{zhu2021ft} proposed a Positive Extrapolation \& Negative Interpolation to improve MoCo\cite{2020mocov2}. Positive Extrapolation does Manifold Mixup, turning the raw positive pairs into a transformation to increase the hardness of the \emph{Memory Bank}; Negative Interpolation uses Mixup to generate mixed samples, enhancing the diversity of negative samples.

In Graph CL, the most challenging negatives may be classified as false negatives if they are selected exclusively based on the similarities between the anchor and themselves. ProGCL\cite{xia2021progcl} said that similarity cannot be used to assess negative sample difficulty and proposed ProGCL-weight to avoid classifying positive samples as challenging negatives. ProGCL-mix is then proposed for more true negatives. Different from ProGCL, limited to graph tasks, DACL\cite{verma2021dacl} proposed a domain-agnostic CL approach using noise samples to generate challenging sample pairs via mixed samples. The CL formulation uses a random sample $x_r$ to satisfy the condition $s(z_i, \hat{z}) > s(z_i, z_r)$, where $s(\cdot)$ is a cosine similarity function between two vectors.

\textbf{Application with Mixup and CL.} CoMix\cite{sahoo2021comix} and MixSSL\cite{zhang2022mixssl} used similar ideas for DA by mixing samples from different domains in the Contrastive Self-Supervised Learning framework for video and medical images, and \emph{m}-Mix\cite{zhang2021m-mix} proposed to mix multiple samples and dynamically assign different $\lambda$ to mine hard negative samples. Un-Mix\cite{shen2022un-mix} considered that DA only changes the distance between samples, but the labels remain the same during training. However, mixup methods will adjust both the sample and the label space, and the degree of change is controllable, which can further capture more detailed features from unlabeled samples and models. 

Based on some crop and cut ways. PCEA\cite{liu2021pcea} proposed using 4 samples from several different augmentation operations to be pasted and then cropped into 4 individual samples, 2 randomly selected for CL training. Similarly, Li \emph{et al.} proposed the Center-wise Local Image Mixture (CLIM)\cite{li2020clim} method, which is situated within a CL framework. Following the identification of cluster centers, CutMix is employed to mix different images in the same class and different resolution images of the same image, which are considered positive sample pairs. CropMix\cite{han2022cropmix} crops an image on multiple occasions using distinct crop scales, thereby ensuring the capture of multi-scale information and resolving the issue that a single random crop tends to result in the loss of information. Subsequently, new input distributions are formed by blending multiple crop views, thereby augmenting the input while reducing label mismatches. CL main encoder is not only based on CNNs but also can be based on ViTs. PatchMix\cite{shen2023patchmix} argued that extant comparison frameworks ignore inter-instance similarities. \emph{e.g.}, disparate views of an image may be utilized as potential positive samples. Following the nature of ViTs, PatchMix randomly combines multiple images from a modest batch at the patch level to generate a sequence of mixed image patches for ViTs, thereby simulating the rich inter-instance similarities observed among natural images. Also, to improve the performance of ViTs, SDMP\cite{ren2022sdmp} regards these as additional positive pairs in SSL by modeling the semantic relationships between the mixed samples that share the same source samples. 

\begin{figure}[ht]
    \centering
    \includegraphics[scale=0.16]{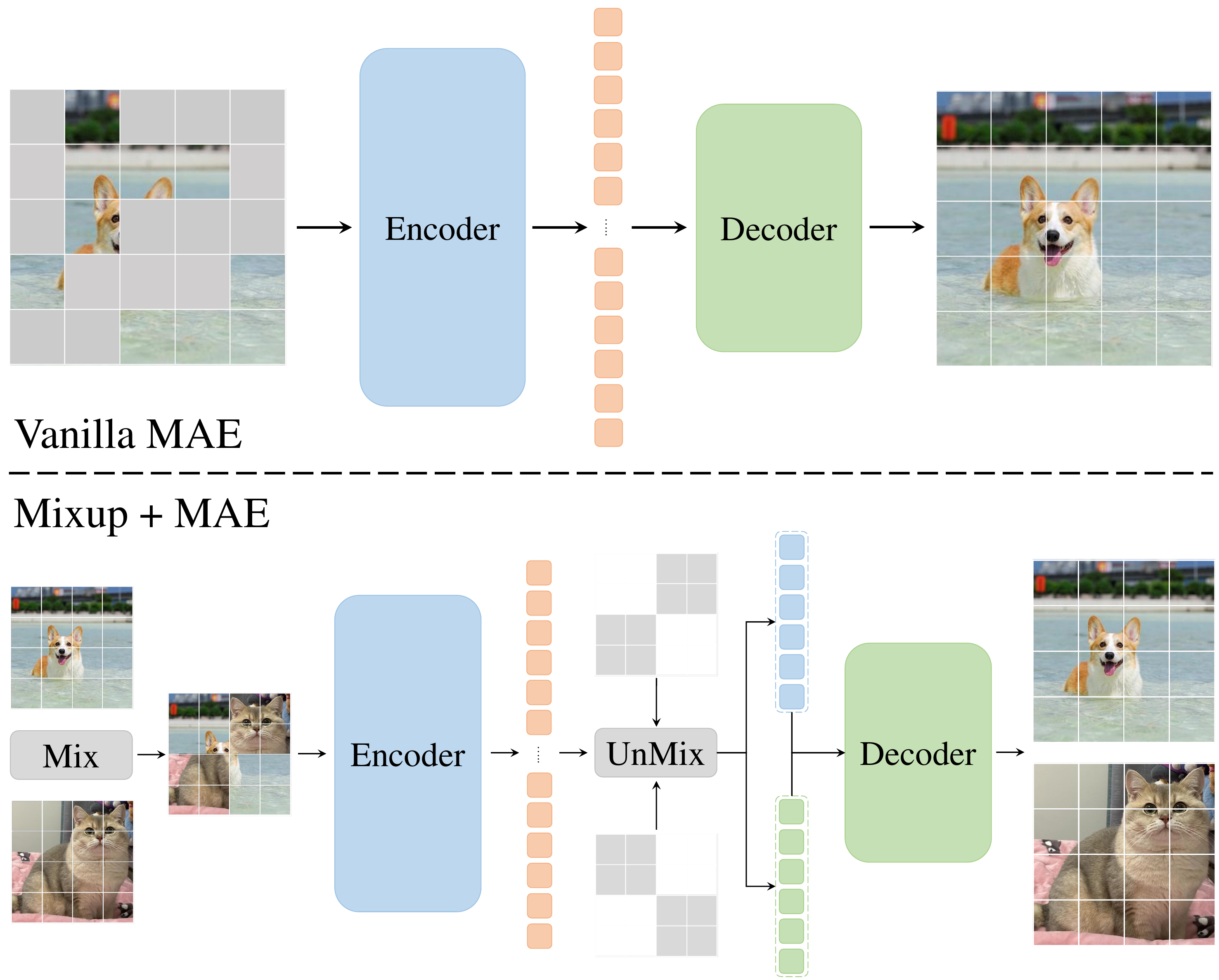}
    \caption{The framework comparison between the Vanilla MAE method and the MAE method with mixup. Vanilla MAE used an encoder and decoder to reconstruct the masked images into raw images. Mixup + MAE used mixed samples and decoded them into raw samples with an UnMix technical.}
    \label{fig: mae_method}
\end{figure}
\subsubsection{\textbf{Masked Image Modeling}}
Masked Autoencoder (MAE)\cite{he2022mae} uses the Masked Image Modeling (MIM) method, mapping the tokens into the semantic space using an encoder, while the pixels in the original space are reconstructed using the decoder. MAE methods\cite{li2022a2amim} demonstrate the powerful feature extraction and reconstruction power of the encoder and decoder. $i$-MAE\cite{zhang2022i-mae} aims to investigate two main questions: (a). Are the latent space representations in masked autoencoder linearly separable? (b). What is the degree of semantics encoded by MAE in the latent feature space? $i$-MAE demonstrated that it is possible to reconstruct the masked mixed samples with two independent linear layers. The weaknesses are that the semantic similarity of like with like is difficult to reconstruct and that intra-class separation requires knowledge of high-level visual concepts. Meanwhile, experiments have found MAE to be powerful in the latent feature space.

MixMAE\cite{liu2023mixmae} combined the advantages of SimMIM\cite{xie2022simmim} and MAE and proposed to encode and decode patch-level mixed samples, using limited information in encoding instead of dropping all and masking non-self-patches in decoding to prevent tangling between the two features during decoding. The proposed patch merging module is used to merge the feature information. Similarly, MixedAE\cite{chen2023mixedae} increases the number from two to four and proposes a Multi-head Homo Attention module, also using the Homo Contrastive method for encoding and decoding. The Homo Attention module identifies the same patch by enforcing attention to the highest-quality key patches using the top-$N$. Homo Contrastive aims to enhance the feature similarity of the same patches by supervised comparison. Fig. \ref{fig: mae_method}  shows the pipeline of the vanilla MAE method and MAE with mixup.

\begin{figure}[ht]
    \centering
    \includegraphics[scale=0.16]{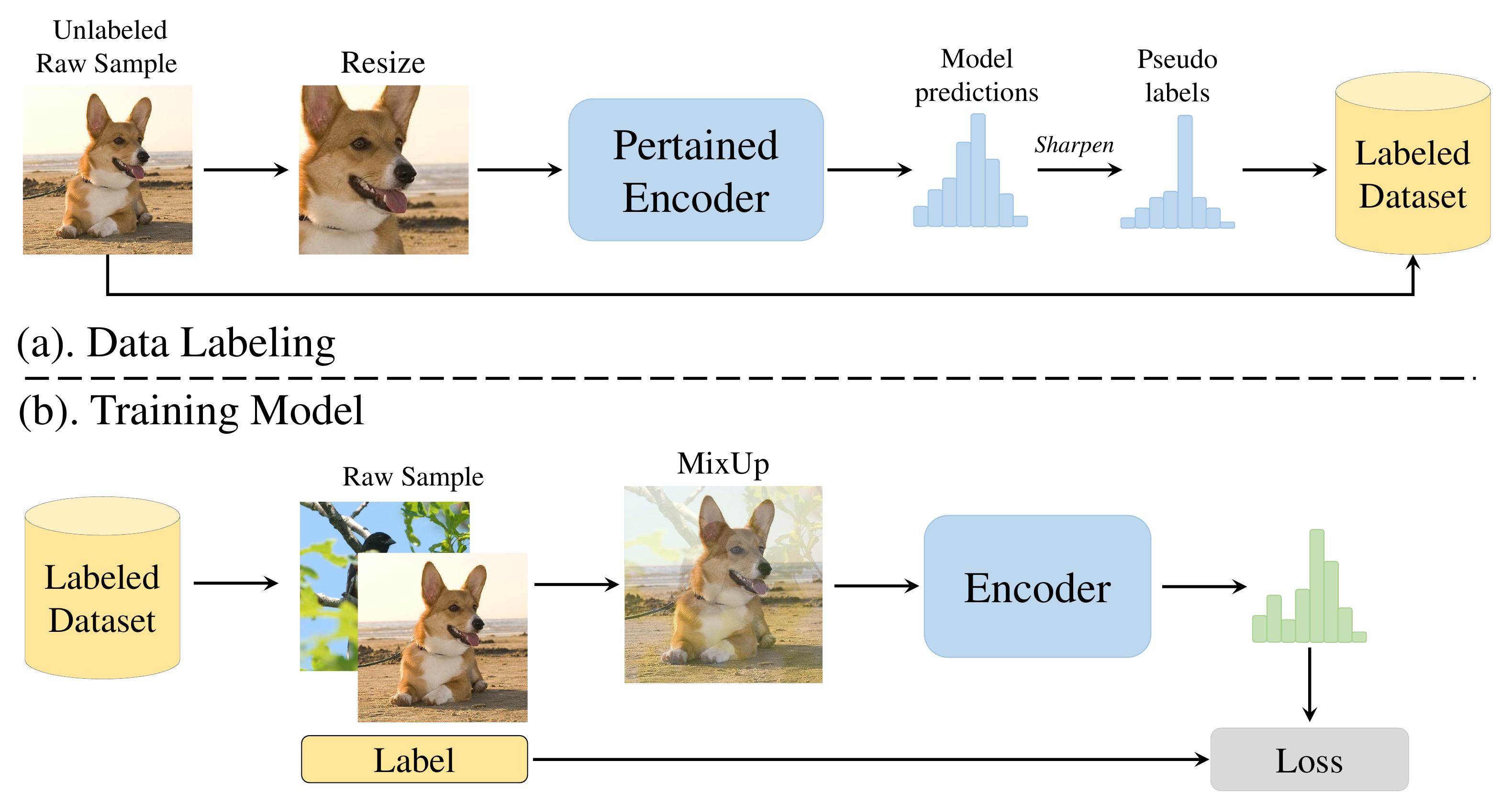}
    \caption{Mixup methods used in SemiSL task. The top part is the pipeline of data labeling, and the bottom part is the model training using mixed samples.}
    \label{fig: semisl_method}
\end{figure}
\subsection{Semi-Supervised Learning}
\label{sec:semisl}
In deep learning tasks, finding huge amounts of unlabeled samples $x_u$ is easy. It is very expensive to make labels. Therefore, researchers combine a large number of unlabeled samples $x_u$ into a limited number of labeled samples $x$ and train them together, which is expected to improve the model performance, thus giving way to Semi-Supervised Learning (Semi-SL). Semi-supervised learning avoids the waste of resources and, at the same time, solves the problems of poor model generalization ability in Supervised Learning and model inaccuracy in Unsupervised Learning. Fig. \ref{fig: semisl_method} shows the processes of mixup methods on Semi-SL tasks.

\textbf{Semi-SL.} MixMatch\cite{berthelot2019mixmatch} is capable of guessing low-entropy labels for data-augmented unlabeled examples and reduces the entropy of the labeling distribution $p$ using an additional sharpening function by Eq. \ref{eq:26}. These ``guess" labels can be utilized to compute the unlabeled loss $\mathcal{L}_u$, which serves as a component of the combined with labeled loss $\mathcal{L}$ to form loss $\mathcal{L}$ for Semi-SL by Eq. \ref{eq:27}. MixMatch employs mixup to mix labeled and unlabeled data, a novel approach separately.
\begin{equation}
    \label{eq:26}
    \begin{aligned}
    \text{Sharpen}(p,\tau)_i=p^{\frac{1}{\tau}}_i/
    \sum_{i=1}^{L}p^{\frac{1}{\tau}}_i,
    \end{aligned}
\end{equation}
\begin{equation}
    \label{eq:27}
    \begin{aligned}
    \mathcal{L}=\mathcal{L}+\textrm{w}_u\mathcal{L}_u,
    \end{aligned}
\end{equation}
where temperature $\tau$ is a hyperparameter, and a reduction in $\tau$ encourages the model to produce lower-entropy predictions. ReMixMatch\cite{kurakin2020remixmatch}  introduces two novel techniques, namely \textit{''Distribution Alignment''} and \textit{''Augmentation Anchoring''}, which are superimposed on top of the MixMatch algorithm. The objective of distribution alignment is to ensure that the marginal distribution of predictions on unlabeled data closely aligns with the marginal distribution of ground truth labels. Augmentation anchoring entails introducing multiple strongly augmented input versions into the model, intending to encourage each output to be close to the prediction for a weakly augmented version of the same input. ICT\cite{verma2022ict} calculated the consistency loss by comparing the prediction of mixed samples of unlabeled samples and the prediction of raw samples. 

Different from the previous method only considers predicted and pseudo-labels. CowMask\cite{french2020cowmask} used the teacher model to obtain pseudo-labels for raw samples and used CowMix to obtain predictions through the student model after obtaining mixed samples for consistency loss $\vert \hat{p}_s - p_t \vert{^2}$. The difference is that CowMask obtains the final pseudo-labels by performing a mix of pseudo-labels and the mean of the mask (as $\lambda$). $\epsilon$mu\cite{pisztora2022emu} mixed labeled and unlabeled samples to obtain augmented samples, and instead of relating the labels of mixup directly to $\lambda$, the labels of mixup are instead adjusted using a hyperparameter $\lambda_{eps}$. The DCPA\cite{chen2023dcpa} obtained a semi-supervised loss by predicting pseudo-labels using the teacher model, a supervised loss using mixed samples, and a contrasting loss using two different decoders that capture the variability of the features through a Sharpen operation, which is used to strengthen the consistency constraints, to bridge gaps in the outputs of the different decoders and to guide the learning direction of the model. MUM\cite{kim2022mum} has developed a new approach to semi-supervised object detection that addresses the limitations of traditional data augmentation methods and aims to preserve the location information of the bounding box. MixPL\cite{chen2023mixpl} proposed to compose the pseudo-labeled data by MixUp and Mosaic ways to soften the negative effect of missed detections and to balance the learning of the model at different scales.

\textbf{Label Noise Learning.} Unlike most existing Label Noise Learning (LNL) methods, DivideMix\cite{li2020dividemix} drops the labels of the samples that are most likely to be noisy and uses the noisy samples as unlabeled data to regularize the model, avoiding overfitting and improving the generalization performance. In addition, two networks are trained at the same time by co-dividing. This keeps the two networks diverging so that different types of errors can be filtered and confirmation bias in self-training can be avoided. Manifold DivideMix\cite{fooladgar2023mixematch} further improves DivideMix by selecting OOD samples and choosing noise samples based on the distribution of classification loss to find clean/noise samples to build a labeled/unlabeled dataset.

\textbf{Positive and Unlabeled Learning.} Positive and Unlabeled Learning (PUL) is a branch of Semi-SL that trains binary classifiers with only positive class and unlabeled data. Similarly, some mixup-based methods can be used to improve the performance of the network in PUL tasks. MixPUL\cite{wei2020mixpul} combined supervised and unsupervised consistency training to create augmented samples. To promote supervised consistency, MixPUL further rewards unsupervised consistency between unlabeled samples by alleviating the supervised problem by mining reliable samples from the unlabeled set by identifying reliable negative sample subsets. In PUL tasks, the decision boundary often biases toward the positive class, resulting in some likely positive samples being mislabeled as negatives. This issue causes the decision boundary deviation phenomenon in PUL. P$^3$Mix\cite{li2021yourp3mix} performs more accurate supervision by mixing unlabeled samples and positive samples from the decision boundary that are close to the boundary.

\subsection{CV Downstream Tasks}
\label{sec:cvdownstream}
\subsubsection{\textbf{Regression}}
Regression tasks differ from classification tasks in that they require a relatively accurate prediction of a result rather than a probability of a class. This implies that directly mixing two random samples does not ensure that the mixed samples are effective for the model and may be detrimental in some cases. MixRL\cite{hwang2021mixrl} proposed using the validation set to learn, for each sample, ``how many nearest neighbors should be mixed to obtain the best model performance". Similar to AutoAugment\cite{cubuk2018autoaugment}, Reinforcement Learning (RL) is used to find the lowest model loss on the validation set to determine this. Specifically, an RNN model is used to determine the optimal number of KNNs for the samples, and then an MLP is used to select all the samples.

C-Mixup\cite{yao2022c-mix} adjusts the sampling probability based on the similarity of the labels and then selects pairs of samples with similar labels for mixing. Simple implementation of vanilla MixUp causes noisy samples and labels. For a sample that has already been selected, C-Mixup uses a symmetric Gaussian kernel to calculate the sampling probability of another sample, where the closer sample is more likely to be sampled. Also, using low-dimensional label similarity for the calculation will reduce the overhead. Similarly to C-Mixup, ADA\cite{schneider2024ada} uses a prior distribution for select samples and labels, UMAP Mixup\cite{el2024umapmixup} uses the Uniform Manifold Approximation and Projection (UMAP) method as a regularizer to encourage and ensure that the mixup generates mixed samples that are on the flow of the feature and label data. ExtraMix\cite{kwon2022extramix} proposed a mixup technique that can be extrapolated, expanding the latent space and label distribution. Compared to existing mixup-based methods, ExtraMix minimizes label imbalance. In addition, cVAE\cite{kang2018cvae} is used to optimize pseudo labels in the mixing to deal with the fact that in materials science, new materials with excellent properties are usually located at the tail end of the label distribution. Warped Mix\cite{bouniot2023warpedmixup} proposed a framework that takes similarity into effect when interpolating without dropping diversity; Warped Mix argued that similarity should influence the interpolation ratio $\lambda$, not the selection. High similarity should lead to a strong $\lambda$, and low similarity should lead to essentially no change. The SupReMix\cite{wu2023supermix} aims to make better use of the inherent ``sequential" relationship between the inputs to facilitate the creation of ``harder" contrast pairs. The objective is to promote continuity as well as local linearity.

\subsubsection{\textbf{Long-tail Distribution}}
In nature, data tends to fall into long-tailed distributions. In deep learning, a superior model should be able to handle long-tailed distributions well. The augmented samples obtained by the mixup methods, since the existence of two different features of information, seem to be well suited to dealing with the long-tailed distribution problem. Remix\cite{chou2020remix} facilitates the labels of the minority class by providing higher weights for the minority class. Allows features and labels to differ in $\lambda$ when constructing mixed samples to provide a better trade-off between majority and minority classes. UniMix\cite{xu2021unimix} uses an improved mixed $\lambda$ and a sampler that facilitates minority classes. The fixed $\lambda$ is re-corrected by the UniMix factor for each class with prior knowledge sampling, more suitable for long-tail tasks based on the mixup method. 

OBMix\cite{zhang2022obmix} uses two samplers $S_1$ and $S_2$ sampling independently to sample all the data classes in the dataset, generating a mini-batch with uniformly distributed classes, and then mixing to achieve the balance between the majority and minority classes. DBN-Mix\cite{baik2024dbn-mix} combines two samples initially generated by the Uniform sampler and re-balance sampler modules to improve the learning of the representation of the minority class. Class-by-class Temperature scaling is also used to reduce the bias of the classifier towards the majority class. Since the samples generated by DBN-Mix are located near the boundaries of the minority class region, where the data points are sparsely distributed, they are used to capture the distribution of the minority class better.

\subsubsection{\textbf{Segmentation}}
Image segmentation is a very crucial research and application in CV, where the pixels in an image are divided into different parts and labeled with different labels by some specific models. Mixup methods preserve samples with diverse features, making them suitable for improving model performance in segmentation tasks. ClassMix\cite{olsson2021classmix} used a pre-trained model as a teacher to obtain the masks of per label and mixed different features obtained from the samples. To address some sites with a high level of symmetry, such as airports, ChessMix\cite{pereira2021chessmix} proposed to mix the patches on samples in a chessboard-like grid, achieved by mixing two different patches not directly located to the left, right, above, or below each other, which is done to avoid the problem of spatial outages. SA-MixNet\cite{feng2024samix-net} by pasting road areas from one image to another. Compared to other mixup methods, SA-MixNet retains the structural completeness of the roads. In addition, a discriminator-based regularization method is designed to increase road connectivity while maintaining the road structure. In Auto-Driving tasks, LaserMix\cite{kong2023lasermix} and UniMix\cite{zhao2024unimix} used the lidar dataset to mix samples.

Segmentation methods are also widely used in medical images. CycleMix\cite{zhang2022cyclemix} proposed using PuzzleMix to mix samples and further randomly occlude patches. Differing from some cutting-based methods, InsMix\cite{lin2022insmix} applies morphological constraints to maintain a clinical nuclear prior. In addition to foreground augmentation, background perturbation is proposed to utilize the pixel redundancy of the background. Smooth-GAN is proposed to harmonize the contextual information between the original nuclei and the template nuclei. DCPA\cite{chen2023dcpa} and MiDSS\cite{ma2024midss} used a pre-trained model as a teacher model to obtain the pseudo labels (some masks) to mix samples. The student model in the DCPA uses two different decoders to capture the variability between features after a Sharpen operation to obtain a contrasting loss, strengthen the consistency constraints, cover the gaps between the outputs of the different decoders, and guide the learning direction of the model. ModelMix\cite{zhang2024modelmix} proposed to construct virtual models using convex combinations of convolutional parameters from different encoders. The set of models is then regularized to minimize the risk of cross-task adjacency in an unsupervised and scribble-supervised way.

\subsubsection{\textbf{Object Detection}} 
MUM\cite{kim2022mum} (Mix\&UnMix) is proposed to be used to improve the performance of a semi-supervised teacher-student framework by mixing the image blocks and reconstructing them in the feature space. The samples are passed to the teacher model to generate pseudo labels. Mixed samples after feature extraction by the student model will be mixed feature maps that are restored based on their masks, the restored feature maps generate predicted labels, and compare pseudo labels to get the loss function. MixPL\cite{chen2023mixpl} uses the teacher model to generate pseudo-labels for both mixup and mosaic to get two different mixes of samples for training. Differing from obtained mixed samples, MS-DERT\cite{zhao2024msdert} proposed a new object detection architecture that mixed one-to-one and one-to-many supervision to improve the performance of models.

\begin{figure*}[t]
    \centering
    \includegraphics[width=0.8\linewidth]{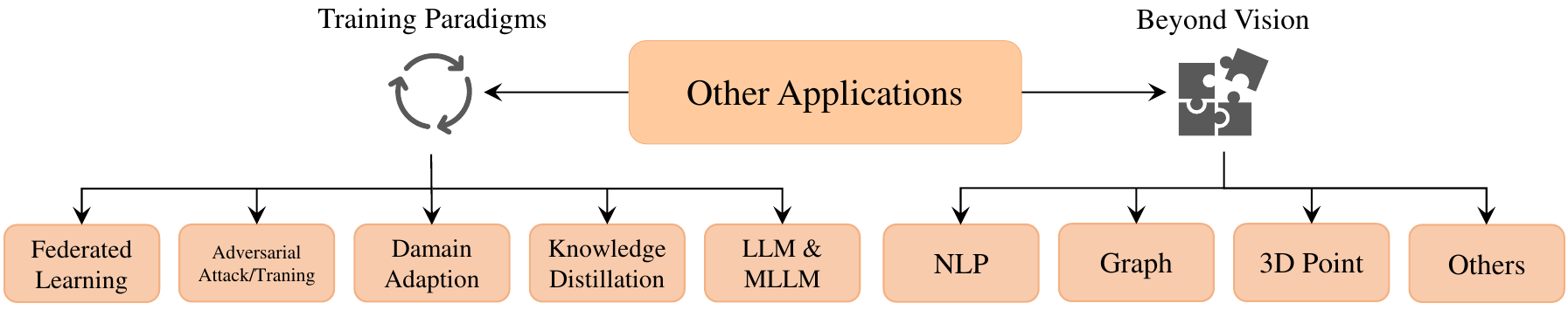}
    \caption{Illustration of other applications, we divided them into two branches: \textbf{Training Paradigms} and \textbf{Beyond Vision}, and divided them into nine types detailed.}
    \label{fig: other_app}
\end{figure*}
\section{Mixup for other applications}
\label{sec:mixup4other}
In this section, we discuss mixup-based methods applied to other tasks. As shown in Fig. \ref{fig: other_app}, we divided them into 2 main subsections: Traning Paradigms and Beyond Vision.

\subsection{Training Paradigms}
\subsubsection{\textbf{Federated Learning}}
Federated Learning (FL) represents a training data decentralized machine learning solution initially proposed by Google in 2016. It is designed to address the issue of data silos by conducting training on distributed data stored in a multitude of endpoints, to develop high-quality centralized machine learning models. To address the problem of homo-distributed data where user-generated data is distributed between devices and tags, Shin \emph{at el.} propose XOR Mixup\cite{shin2020xormixup}, which collects encoded data from other devices that are decoded using only each device's data. The decoding provides synthetic but realistic samples that induce a homo-distributed dataset for model training. The main idea of the XOR Mixup key idea is to utilize the dissimilarity operation property: ($x_i$ $\oplus$ $x_j$) $\oplus$ $x_j$ = $x_i$, $x_i$ and $x_j$ from two individual devices. 

FedMix\cite{yoon2021fedmix} proposed a simple framework, Mean Augmented Federated Learning (MAFL), in which clients send and receive locally averaged data according to the privacy requirements of the target application. To alleviate the performance degradation suffered due to the increased dissimilarity of local data between clients. Based on Federated Learning Distillation (FLD) and MixUp, Mix2FLD\cite{oh2020mix2fld} is proposed. Specifically, each device in Mix2FLD uploaded its local model outputs as in FD and downloaded the model parameters as in FL, thus coping with the uplink-downlink channel asymmetry. Between the uplink link and downlink, the server runs knowledge distillation to transfer the teacher's knowledge to the untrained student model (\emph{i.e.} the global model). However, this output-to-model conversion requires additional training samples to be collected from the device, incurring significant communication overhead while violating local data privacy. StatMix\cite{lewy2022statmix} computed image statistics for individual nodes, \emph{i.e.}, the mean and standard deviation of each color channel, as content and style; distributes the computed statistics to all nodes via a central server; and performs style delivery using these statistics in individual nodes.

\subsubsection{\textbf{Adversarial Attack \& Adversarial Training}}
Adversarial Attack \& Training\cite{szegedy2013adv} can markedly enhance model robustness since it encourages the model to explore some unseen regions and OOD, mixup methods, moreover, bolster model performance and forestall model overfitting in the task. To improve model robustness, M-TLAT\cite{laugros2020m-tlat} uses the MixUp in addition to a randomly generated dummy label, combines the mixed samples and their labels through a classifier to get the gradient perturbed noise $\delta$, and then mixes the noise and mixed samples to get the final augmented samples and labels according to the Eq. \ref{eq:28}:
\begin{equation}
    \label{eq:28}
    \begin{aligned}
    x_{adv} = \hat{x} - \epsilon * sign (\nabla_{\hat{x}} \mathcal{L}(\hat{x}, \hat{y}, \theta)),
    \end{aligned}
\end{equation}
where the $\epsilon * sign(\cdot)$ denotes the FGSM\cite{iclr2015fgsm} or PGD\cite{madry2017pgd} white box attack, and $\theta$ denotes the parameters of the models. The purpose of the M-TLAT is to increase Corruption robustness, while TLAT aims to increase Adversarial robustness and solve the generalized robustness problem.

MI\cite{pang2019MI} converts the Training Stage into the Inference Stage by linearly mixing the source samples with the target samples that have been noise-added by the adversarial attack, and during the Inference Stage, it is found to be optimal for MI compared to the direct use of the adversarial attack. Differ from MI, AVMixup\cite{lee2020avmixup} adds the perturbed noise in the input-level, and also analyses the effectiveness of soft-labeling on Adversarial Feature Overfitting (AFO). IAT\cite{lamb2019iat} combines the clean samples loss and adversarial attack samples loss for training the model and getting a better balance of robustness and accuracy. Similarly, Mixup-SSAT\cite{jiao2023mixup-ssat} explores the robustness of vehicle trajectory prediction and proposes an adversarial attack training method for trajectory prediction. The model's robustness and decision-making ability to address extreme cases are improved by adding human-perturbed historical trajectories.

AMP\cite{liu2021amp}, AOM\cite{bunk2021aom}, and AMDA\cite{si2021amda} proposed for the text classification tasks. Liu \emph{et al.} argued that the linear restrictions imposed by Vanilla MixUp under the input space tend to be underfitting, especially when the training samples are too less. Therefore, they proposed the Adversarial Mixing Policy (AMP), which adds an adversarial perturbation to the mixing ratio to relax the linear restrictions. AOM combines mixup methods and PGD adversarial optimization, intending to obtain more robust classification models at the expense of a small accuracy. Si \emph{et al.} argued that in NLP tasks, the sample search space created by simple adversarial training is poorly captured. AMDA is proposed to do the mixing in the latent space of the pre-trained model and use the generated samples to add to the training stage of the model, which can cover the space of the samples more maximally and be closer to the distribution of the raw samples.

\subsubsection{\textbf{Domain Adaption}}
Models trained on a specific domain typically demonstrate poorer performance when transferred to another domain. To address this issue, the domain adaptation strategy transfers model knowledge from the label-rich source domain to another label-scarce target domain. In the context of Unsupervised Domain Adaptation (UDA)\cite{mao2019VMT,wu2020dmrl,yan2020iimt,xu2020dm-ada,sahoo2021comix,zhang2023samixup,zhang2022mixssl} and Partial Domain Adaptation (PDA)\cite{sahoo2023slm}, there has been a notable push to enhance the effectiveness of models through the utilization of mixup ways for inter- or intra-domain data mixing.

Virtual Mixup Training (VMT)\cite{mao2019VMT}, as a regularization method, is capable of incorporating the locally-Lipschitz constraint into the region between the training data rather than being constrained to the area around the training points by applying penalties to the difference between the prediction $p_i$ and the virtual label $y_i^v$.
Dual Mixup Regularization Learning (DMRL)\cite{wu2020dmrl} to better ensure the discriminability of the latent space and extract domain-invariant features therein, the domain mixup regularization $\mathcal{L}^r_{adv}(\text{En}(x), \text{Dis}_d(x))$ and class mixup regularization $(\mathcal{L}^r_s(\text{En}(x), \text{Cls}(x))$ and $\mathcal{L}^r_t(\text{En}(x), \text{Cls}(x)))$ are incorporated into the traditional domain adversarial neural network according to Eq. \ref{eq:29}.
\begin{equation}
    \label{eq:29}
    \begin{aligned}
        \mathop{\min}_{\text{En},\text{Cls}} \mathop{\max}_{\text{Dis}_d}
        \mathcal{L}(\text{En}(x), \text{Cls}(x)) + \textrm{w}_d\mathcal{L}_{adv}(\text{En}(x), \text{Dis}_d(x)), 
    \end{aligned}
\end{equation}
where $\text{En}(\cdot)$, $\text{Dis}_d(\cdot)$, and $\text{Cls}(\cdot)$ represent the feature extractor, domain discriminator, and classifier, respectively. $\mathcal{L}^r_{adv}(\text{En}(x), \text{Dis}_d(x))$ denotes the loss function that distinguishes whether the sampled feature is from the source or target domain, $\mathcal{L}^r_s(\text{En}(x), \text{Cls}(x))$ is employed for the assessment of the loss of consistency in mixed source samples and their corresponding labels, and $\mathcal{L}^r_t(\text{En}(x), \text{Cls}(x)))$.

Domain Adaptation with Domain Mixup (DM-ADA)\cite{xu2020dm-ada}, which also ensures domain-invariance of the latent space, improves on VAE-GAN\cite{Larsen2015VAE_GAN} for adversarial networks by constraining domain-invariance not only on the source and target domains, but also on the intermediate representations between the two domains. In contrast to DMRL\cite{wu2020dmrl}, DM-ADA implements inter-domain data sample and latent space mixing and employs soft labels in the domain discriminator to inform the evaluation.
IIMT\cite{yan2020iimt} employs both intra-domain and inter-domain mixing within an adversarial learning framework to enforce training constraints. SLM\cite{sahoo2023slm} eliminates negative migration by removing outlier source samples and learns discriminative invariant features by labeling and mixing samples to improve the PDA algorithm.
CoMix\cite{sahoo2021comix} proposed a contrastive learning framework for learning unsupervised video domain-adaptive discriminative invariant feature representations. This is achieved by using background mixing, which allows for additional positives per anchor, thus adapting contrastive learning to leverage action semantics shared across both domains.

In the field of medical images, Mixup Self-Supervised Learning (MixSSL) \cite{zhang2022mixssl} framework for contrast-agnostic visual representation learning is employed to enhance the robustness of medical image classification models by mixing medical training image samples and natural image samples. SAMixup\cite{zhang2023samixup} employs Domain-Distance-modulated Spectral Sensitivity (DoDiSS) to extract sensitive features across domains in the image phase. Furthermore, it utilizes the DoDiSS map $z_s$ and adversarially learned parameter $\lambda_\theta$ as a weighting factor for cross-domain phase mixing operation:
\begin{equation}
    \label{eq:30}
    \begin{aligned}
        \hat{Am} = \lambda_\theta z_s Am^t + (1-\lambda_\theta) (1 - z_s) Am^s 
    \end{aligned}
\end{equation}

\subsubsection{\textbf{Knowledge Distillation}}
Knowledge Distillation (KD) uses a pre-trained teacher model to train a student model. This knowledge transfer method represents a promising paradigm for the efficient training of deep neural networks, offering the potential to obtain robust models with less data in less time. To transfer robustness effectively, MixACM\cite{muhammad2021mixacm} passes the mixed samples through a robust teacher and a student to obtain intermediate features, which are then passed through a mapping function to obtain activated channel maps. Finally, the student model minimizes the inter-network loss to complete the knowledge transfer. 

MixSKD\cite{yang2022mixskd} fused MixUp and Self-Knowledge Distillation into a unified framework that uses the probability distribution generated after the coding of the mixed images as a pseudo teacher distribution $f^\prime(\hat{x}_{i,j})$, guiding the original image pairs to produce consistent mixup predictions $\hat{p} = \lambda * f^\prime(x_i) + (1-\lambda) * f^\prime(x_j)$ in the network. This approach realizes the feature and probability space level of mutual distillation. Moreover, MixKD\cite{choi2023mixkd} discusses the impact of mixup augmentation on knowledge distillation and finds that ``smoothness" is the connecting link between the two. Through an analysis of the experiments, the article suggests using a partial mixup (PMU) strategy in KD along with using standard deviation as the temperature $\tau$ for scaling.

\subsubsection{\textbf{LLM \& MLLM}}
Large Language Model (LLM) and Multi-modal Large Language Model (MLLM) has show its power ability based on numerous training data and huge parameters, has been successfully utilized in lots of tasks. 

\textbf{Basic multi-modal:} VLMixer~\cite{wang2022vlmixer} integrates Cross-modal CutMix (CMC) and contrastive learning to transform uni-modal text into multi-modal text  \& images, improving instance-level alignment between different modalities. CMC transfers natural sentences to multi-modality by randomly replacing vision-based words with semantically similar image patches, increasing data diversity while preserving semantic integrity. Oh \emph{at el.} found that despite CLIP's~\cite{radford2021clip} goal to align image and text embeddings, it maintains separate subspaces with large gaps, leading to poor uniformity and alignment even after fine-tuning, which limits embedding transferability and robustness. $m^2$-Mix ~\cite{oh2024m2-mix} addresses this by mixing image and text embeddings, generating hard-negative samples on the hyper-ball, and fine-tuning the model on hard-negative, raw, and positive samples using contrastive loss.
$\mathcal{P}$owMix~\cite{georgiou2023powmix} is an improvement of MultiMix, consists of five components: different number of mixing samples, mixing ratio reweighting, anisotropic/dynamic mixing, and cross-modal label mixing.

\textbf{LLMs/MLLMs:} MixGen~\cite{hao2023mixgen} generates new image-text pairs by augmenting images with MixUp and linking texts to maintain semantic relationships. VQAMix~\cite{gong2022vqamix} proposes two approaches—ignoring mismatch mixed labels and conditional dynamic learning to adapt LLM to VQA data modality. This enables applying the Mixup method to VQA tasks. In LLM training, multi-domain corpora are commonly used. These corpora are often mixed based on human a priori knowledge. To identify an optimal mixing ratio, RegMix~\cite{liu2025regmix} trains small-scale proxy models based on small-scale corpora with different mixing ratios. Subsequently, a linear regression model is employed to predict the most suitable mixing ratio.
Beyond dataset-level mixing, researchers Zhou \emph{et. al.}\cite{zhou2025m3} proposed linearly interpolating the weight parameters of multiple LLMs, which were fine-tuned for different tasks. They posited the mixed LLM could inherit the capabilities of the original fine-tuned LLMs.

\subsection{Beyond Vision}
\subsubsection{\textbf{Natural Language Processing}}
NLP is a significant subfield of artificial intelligence that enables computers to understand, interpret, and generate human language. Mixup methods have been extensively utilized in CV and explored to some extent in NLP, achieving enhancements of text data through the mix at the sample, token, and hidden space levels.

\begin{figure}[h]
    \centering
    \includegraphics[scale=0.4]{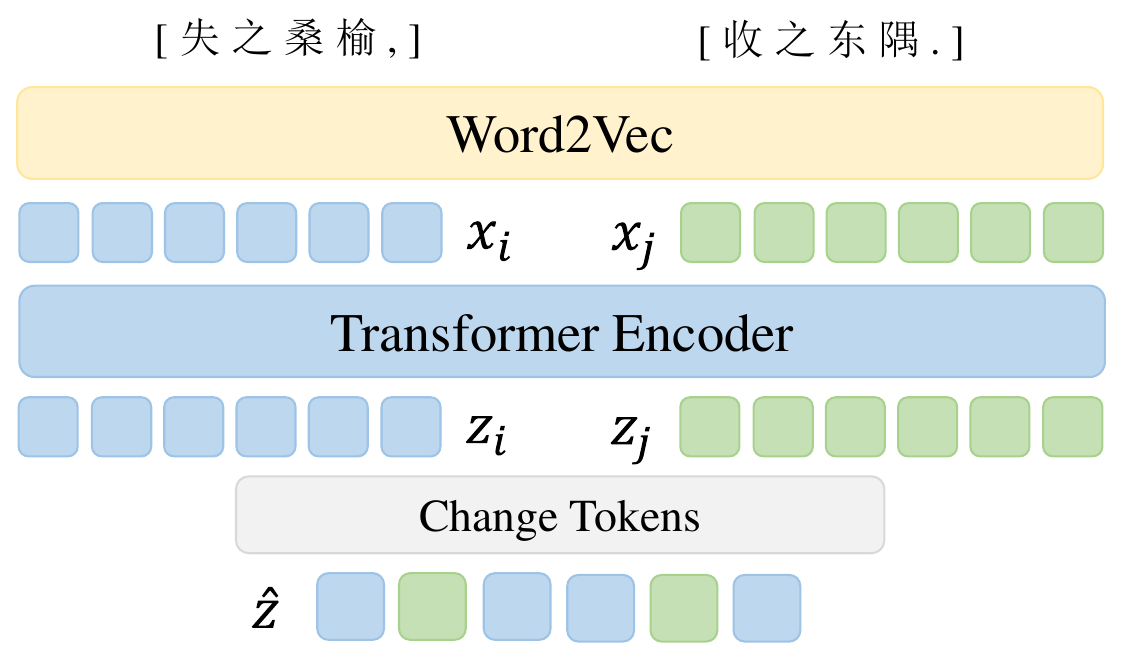}
    \caption{Mixup methods used in text dataset for the text classification task. Two sentence embedding by Word2Vec\cite{mikolov2013word2vec}, $x_i$, $x_j$ encoded and mixed their tokens.}
    \label{fig: NLP_method}
\end{figure}

\textbf{Text Classification.}
WordMixup \& SenMixup\cite{guo2019WordMixup&SenMixu} proposed two ways, WordMixup obtains mixed samples by linearly interpolating the mixup at the input level, while SenMixup is mixed by the difference in the latent space as shown in Fig. \ref{fig: NLP_method}. Since some sentences have different sizes, WordMixup fills both sentences with 0 to the same size and then interpolates each dimension of each word in the sentence. SeqMix\cite{zhang2020seqmix} searches for pairs of matching sequences at each iteration and mixes them in feature space and label space; a discriminator is used to determine whether the resulting sequences are plausible or not. This discriminator calculates the perplexity scores of all the candidate sequences created and selects the low perplexity sequences as plausible sequences. Similarly, Mixup-Transformer\cite{sun2020mixup-transformer} takes the two sentences through the Transformer to obtain their representation and then linearly interpolates them to obtain mixed samples for classification. Seq-level Mix\cite{guo2020seq-level-mix} obtained soft mixed samples by randomly combining parts of two sub-parts of a sentence. This prevents the model from memorizing sub-parts and motivates the model to rely on the composition of sub-parts to predict the output. 

MixText\cite{chen2020mixtext}, as a semi-supervised method, generated a substantial number of augmented samples by mixing labeled and unlabeled text in the latent space. Furthermore, it minimizes the classification loss of labeled, unlabeled, and augmented samples, as well as the consistency loss between pseudo and ground truth labels. Similarly, EMix\cite{jindal2020EMix} used interpolations of word embedding and latent layer representations to construct virtual examples in the sentence classification task, thereby mitigating model overfitting. It is worth noting that EMix utilizes the standard deviation $\sigma$ of the text to define ratio $\lambda_t$, thus performing a mixing calculation that takes into account the differences in text energies:
\begin{equation}
    \label{eq:31}
    \begin{aligned}
     &\hat{z} =\frac{\lambda_t z_i + (1- \lambda_t) z_j}{\sqrt{\lambda_t^2 + ( 1 - \lambda_t)^2}},
    \end{aligned}
\end{equation}
where $\lambda_t =\frac{1}{1+\frac{\sigma_i}{\sigma_j} \cdot \frac{1-\lambda}{\lambda}}$. Unlike the above methods, TreeMix\cite{zhang2022treemix} enhances the diversity of the generated samples by employing the constituency parsing tree to decompose sentences into sub-structures and recombine them into novel sentences through mixup. Nevertheless, the task of devising valid labels for the enhanced samples represents a significant challenge, and TreeMix employs the proportion of words in the new sentences as the mixup ratio for label convex combination:
\begin{equation}
    \label{eq:32}
    \begin{aligned}
     \hat{y}=\frac{L_i-|t^K_i|}{L_i-|t^K_i|+|t^L_j|}y_i+
     \frac{|t^L_j|}{L_i-|t^K_i|+|t^L_j|}y_j,
    \end{aligned}
\end{equation}
where $L$ is the length of sentence $x$ and $|t|$ is length of sub-sentences, $K$ is length of others. $L_i-|t^K_i|$ words from $x_i$ are kept and $|t^L_j|$ words from $x_j$ are inserted in new sentence.

Since the text consists of discrete tokens of variable length, there is an issue of applying MixUp to NLP tasks. Yoon \emph{et al.} proposed SSMix\cite{yoon2021ssmix}, which is optimized for NLP tasks by applying it on the input level instead of mixing it in the latent space as in the previous approach. SSMix preserves the locality of the two raw texts by span-based mixing while synthesizing a sentence, preserving the tokens more relevant for prediction that depend on saliency information. $\max(\textrm{mix}(\lambda * x_i, x_j), 1)$ finds tokens of the same length for replacement. Different from WordMix, Nonlinear Mixup\cite{guo2020nonlinearmixup} used a unique mixing $\lambda$ for each dimension of each word in a given sentence, changing the samples and labels to nonlinear interpolation, and the label mixing is based on input by adaptive learning. Specifically, for sentences shown by $x \in \mathbb{R}^{\mathbb{N} \times C}$, the Nonlinear Mixup's mixing strategy is a matrix $\lambda \in \mathbb{R}^{\mathbb{N} \times C}$, where each element of $\lambda$ is sampled independently from the Beta distribution.

\textbf{Neural Machine Translation.}
Neural machine translation (NMT) has demonstrated considerable success in enhancing the quality of machine translation. In contrast to the traditional diverse generation of translations, which is based on a modeling perspective, MixDiversity\cite{li2021MixDiversity} is a data-driven approach that generates different translations for source sentences by mixing different sentence pairs sampled from the training set with the source sentences during the decoding process, to improve translation diversity. AdvAug\cite{cheng2020advaug} proposed an adversarial augmentation method in NMT tasks. This method involves replacing words in a sentence to form adversarial samples and then selecting samples from the domain for convex combination over the aligned word embeddings. AdvAug enhances the diversity of the adversarial samples and improves the robustness of the NMT model by mixing observed sentences of different origins. Multilingual Mix\cite{cheng2022Multilingualmix} fuses two multilingual training examples by introducing a multilingual crossover encoder-decoder to generate crossover examples that inherit the combinations of traits of different language pairs, thereby better utilizing cross-linguistic signals. In cross-domain translation tasks, there is often an issue that the translated text tends to add noise. X-Mixup\cite{yang2022Xmixup} proposed imposes \emph{Scheduled Sampling}\cite{bengio2015scheduled} and \emph{Mixup Ratio} to handle the distribution shift problem and data noise problem, respectively.

\textbf{Others.}
STEMM\cite{fang2022stemm} proposed that the speech samples are subjected to a w2t and a CNN to obtain their semantic representations, and the text samples are divided into tokens and then subjected to an embedding layer to obtain the representations; since the degree of speech representations tends to be larger than that of text representations, the authors use a token-level impose on them. LADA\cite{chen2020ladamixup} proposed a local additivity-based data augmentation method for the Semi-SL Named Entity Recognition task, similar to MixText\cite{chen2020mixtext}, by interpolating among tokens within one sentence or different sentences in the hidden space. This method creates an infinite amount of labeled data, thereby improving both entity and context learning. Pre-trained language models, due to their excessive parameterization, may encounter significant miscalibration between in-distribution and OOD data during fine-tuning. To address this issue, CLFT\cite{kong2020clft} generated pseudo-on-manifold samples by interpolating within the data manifold. This approach aims to impose a smoothness regularization, enhancing the calibration of in-distribution data. Concurrently, CLFT encourages the model to produce uniform distributions for pseudo-off-manifold samples, mitigating the over-confidence issue associated with OOD data. HypMix\cite{sawhney2021hypmix} posits that interpolation in Euclidean space can introduce distortion and noise, thereby proposing a novel approach to mixup data in the latent space of Riemannian hyperbolic space. This method offers a more effective capture of the complex geometry inherent to input and hidden state hierarchies.

\begin{figure}[ht]
    \centering
    \includegraphics[scale=0.25]{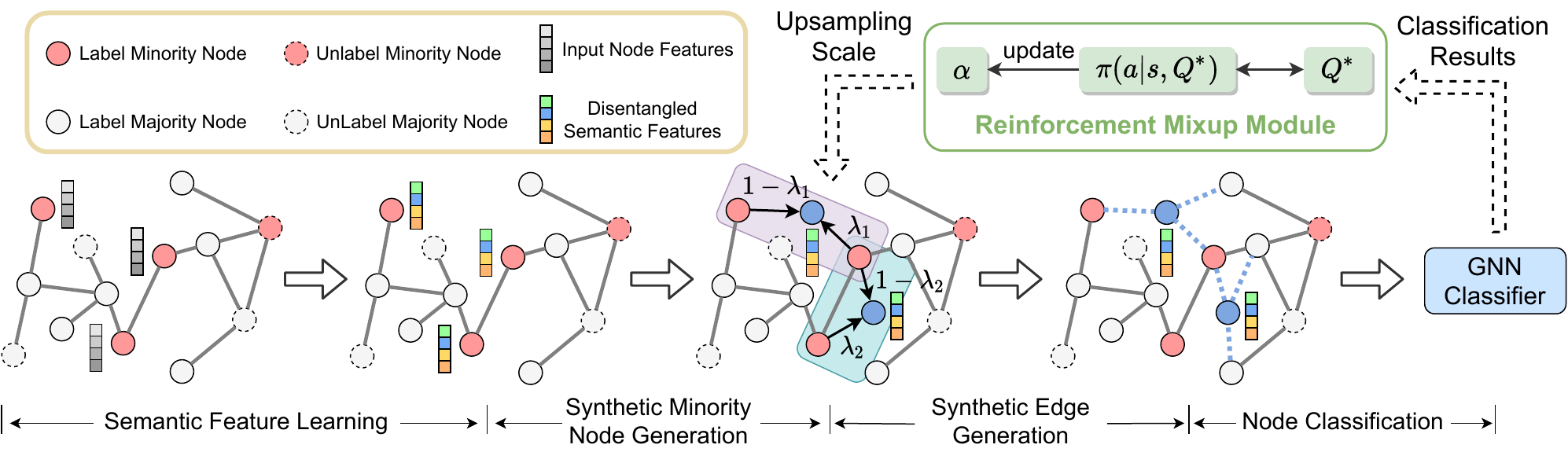}
    \caption{Mixup methods used for the graph classification task. The figure is reproduced from \cite{wu2021graphmixup}.}
    \label{fig: GNN_method}
    \vspace{-0.5em}
\end{figure}

\subsubsection{\textbf{Graph Neural Network}}
Graph Convolutional Network (GCN) has received attention and research because of its unique computational power and has now become a major branch in the field of deep learning. Some traditional deep learning models have achieved good results on Euclidean spatial data (text, image, video, \emph{etc.}), but there are some limitations in processing non-Euclidean spatial data (\emph{e.g.}: social networks, information networks, \emph{etc.}). To solve this problem, researchers have introduced Graphs in the abstract sense of graph theory to represent non-Euclidean structured data. GNNs are used to process the data from a Graph to explore its features and patterns deeply. Xue \emph{et al.} proposed three augmentation methods\cite{xue2021NodeAug-I} for GNNs: NodeAug-I, NodeAug-N, and NodeAug-S. NodeAug-I obtained new virtual nodes $v$ by randomly sampling the feature information and labels of two nodes and mixing. The performance is limited because NodeAug-I ignores the edges of virtual nodes $v^v$. Therefore, NodeAug-N and NodeAug-S are further proposed. NodeAug-N obtains virtual edges by selecting $v_i$ with probability $\lambda$ and $v_j$ with probability $1 - \lambda$; since the virtual edges are directed, they do not affect the inference to the existing nodes. NodeAug-S obtains new virtual edges by aggregating the directed edges of all neighbors of $v_i$ by $\lambda$ scaling and all neighbors of $v_j$ by aggregating the directed edges of $v_j$ by $1 - \lambda$ scaling. 

Different from Xue \emph{et al.}, MixGNN\cite{wang2021mixgnn} used randomly paired nodes to mix their receptive field subgraphs. MixGNN proposed a double-branch mixup Graph Convolutional Network (GCN) to interpolate irregular graph topologies. At each layer, GCN is performed in two branches according to the topology of the paired nodes, and then aggregated representations are interpolated from the two branches before the next layer. NodeAug-I and MixGNN mixed at the input-level, without factoring in that some minority classes in the graph samples tend to be sparse, which is not conducive to direct mixing to get the samples. PMRGNN\cite{ma2022PMRGNN} is a PageRank-based method to solve the problem of difficult scale neighborhoods in node classification tasks. The model's performance is boosted by designing a PageRank-based random augmentation strategy, combining two encoders to complement the cross-representation between the features, and designing a regularization term for the graph to find more features from the neighboring nodes. Graph Transplant\cite{park2022Graph-Transplant} proposed mixup at the graph level, selecting the node and edge with the maximum saliency information (obtained from the gradient) of the two samples and then mixing them to create a new edge, which allows the mixing of irregular graphs at the data level.

GraphMix\cite{verma2021graphmix} is a regularization method for semi-supervised object classification based on GNN. It employs a Manifold Mixup as a data augmentation tool applied to the hidden layer, whereby parameters are shared directly between the Fully-Connected Network (FCN) and the GNN. This facilitates the transfer of discriminative node representations from the FCN to the GNN.
GraphSMOTE\cite{zhao2022GraphSMOTE} proposed to turn a sparse graph into a high dimension and dense representation by a GNN encoder, randomly selecting the minority sample $x_i$, assuming embedding as node $z_i$, and then finding the nearest neighboring node $z_n$ of the same label sample, and then mixing the difference between them to obtain the new node $\hat{v}$, and a decoder used to predict the new edges $u$. In GraphSMOTE, the Synthetic Minority Over-sampling Technique (SMOTE) strategy solves this problem by creating a minority pseudo-example of the class to balance the training data. 
Similarly, in addressing the issue of class imbalance during node classification, GraphMixup\cite{wu2021graphmixup} proposed the implementation of a semantic-level feature mixup, complemented by the introduction of a reinforcement mixup mechanism, to adaptively decide how many samples will be generated by the mixup of these few classes.
iGraphMix\cite{jeong2023igraphmix} proposed the irregularity and alignment issues for graph node classification. It aggregates the sampled neighboring nodes instead of only interpolating node features. Moreover, it analyzes the theoretical and demonstrates a better generalization gap.

\begin{figure}[h]
    \centering
    \includegraphics[scale=0.28]{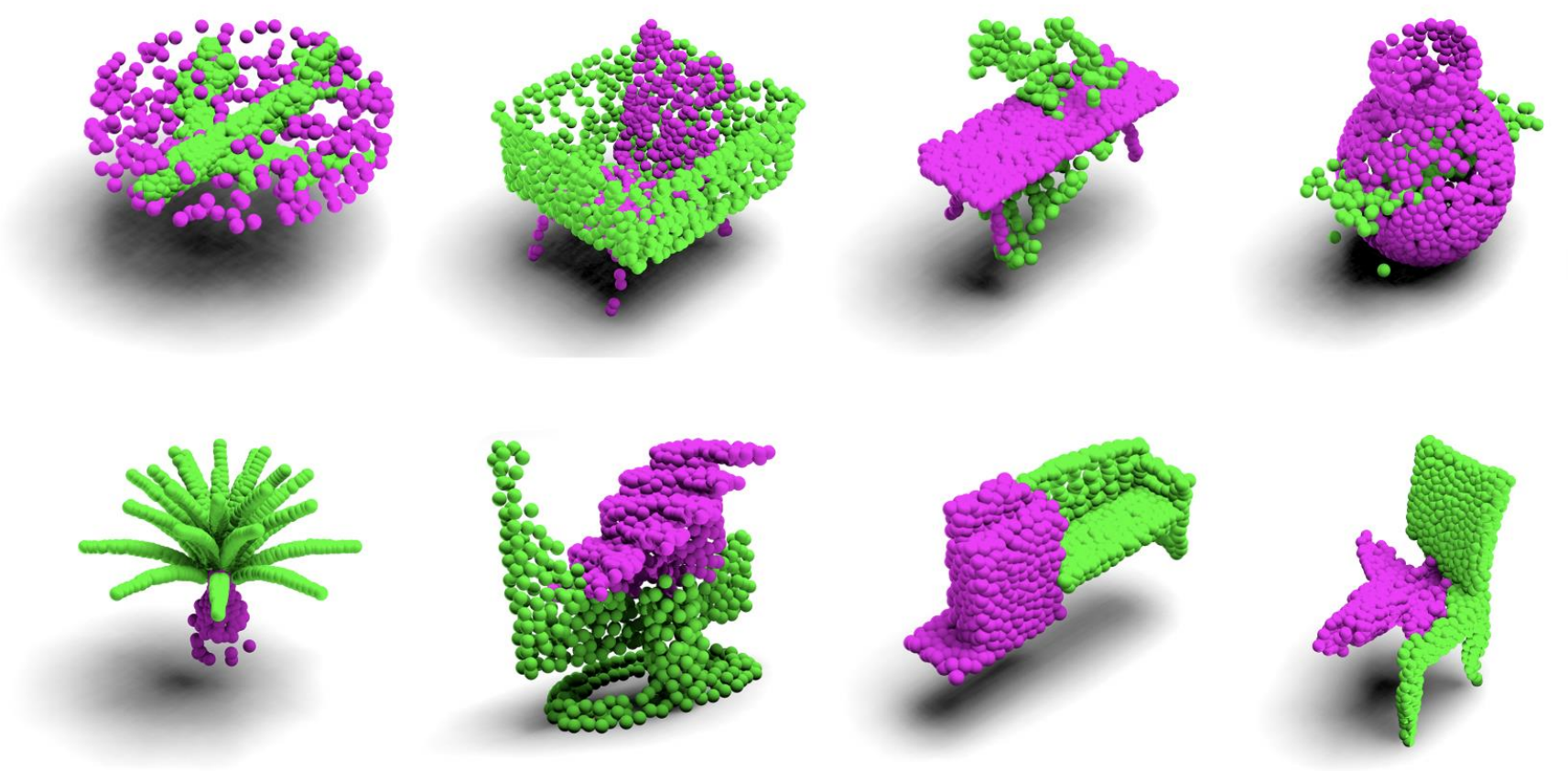}
    \caption{Visualization mixed samples based on 3D point clouds dataset, The figure is reproduced from \cite{zhang2022pointcutmix}.}
    \label{fig: 3d_point}
\end{figure}
\subsubsection{\textbf{3D Point Cloud}}
3D point cloud data is different from RGB images, which consist of several 3D coordinate sets of data. It is widely used in Auto-Driving and 3D reconstruction tasks. MixUp cannot be used directly in the point cloud task since point clouds do not have a one-to-one correspondence between the points of two different objects, so PointMixup\cite{chen2020pointmixup} reviewed the relationship between the data and the method and defined the augmentation of the point cloud data as a linear interpolation of the shortest paths using Earth Mover's Distance\cite{rubner2000emd} to calculate the smallest total shift required to match the corresponding points. The mixed samples are then created by optimally assigning a path function between the two point clouds. 

For specific network architectures, \emph{e.g.} PointNet++\cite{qi2017pointnet++}, RS-CNN\cite{liu2019rs-cnn}, \emph{etc.}, which are more concerned with local features, PointMixup can easily fall into the trap of being locally ambiguous and non-natural, and PointCutMix\cite{zhang2022pointcutmix} allows for natural mixing by creating a new sub-set, and then cutting and pasting two samples. RSMix\cite{lee2021rs-mix} redefines Rigid Subset (RS) by extending the concept of mask region from 2D to 3D, using kNN to maintain the shape of the point cloud and find the similar neighbors; RS is then extracted from each sample to mix the samples; it is possible to mix the two samples maintaining the original 3D shape, retaining part of the shape of the raw sample. Similar to RSMix and PointCutMix, PA-AUG\cite{choi2021pa-aug} divides the whole data into 4 or 8 blocks and randomly applies five augmentation methods \emph{e.g.} point Dropout, CutMix, CutMix \& MixUp, sparse sampling, and random noise creation. Differently from these Hand-crafted methods, Point MixSwap\cite{umam2022point-mixswap} proposed a learnable Attention module that decomposes a point cloud into several disjoint point subsets, called divisions, where each division has corresponding divisions in another point cloud. The augmented point cloud is synthesized by swapping these matches, resulting in a highly diverse output.

\subsubsection{\textbf{Others}}
Since the mixup method is a data-centric tool, it can serve any model-centric task. 

\textbf{Signal Data Type:} Constrastive-mixup\cite{zhang2022contrastive-mixup} uses mixup in speech recognition tasks, where two speech samples are mixed in the input-level and then classified, and unlike most mixup losses $\mathcal{L}_{MCE}$, Constrastive-mixup uses infoNCE loss for training. LLM\cite{chang2023LLM} proposed a learnable loss function to obtain augmented samples by mixing samples with random noise. Octave Mix\cite{hasegawa2021octave-mix} proposed to cross-process the low-frequency waveforms and high-frequency waveforms using frequency decomposition, extract the high and low-frequency information of $x_i$, $x_j$, respectively, cross-mix them, and then mix them by linear interpolation.

\textbf{Metric Learning:} To fix the bias that the sampling method in most CL imposes on the model and the extra overhead of additional hard negative samples, EE\cite{ko2020ee} proposed an extended method for metrics of learning loss in the embedding space, which creates synthetic points containing augmented information by combining feature points, and mines hard-negative pairs to obtain the most informative feature representations. Metrix\cite{venkataramanan2021metrix} proposed a generalized formulation containing the existing metric learning loss function, modified to adapt it to MixUp, and introduces \emph{metric-utilization} that demonstrates regions' improvement by exploring beyond the training classes spatial regions by mixing the samples during training. 

\textbf{AI for Biological:} DNABERT-S\cite{zhou2024dnabert-s} can efficiently cluster and separate different species in the embedding space. This improvement results from the proposed MI-Mix loss and Course Contrast Learning (C2LR) strategies. CL enables the model to distinguish between similar and dissimilar DNA sequences, and course learning progressively presents more challenging training samples, facilitating better learning and generalization.

\textbf{Low-level Task:} Yoo \emph{et al.} found that in cases where spatial relationships are important in Super-resolution tasks, previous methods using dropping or processing pixels or features severely hindered image restoration, and proposed CutBlur\cite{yoo2020cutblur} to cut Low-Resolution (LR) patches and paste them into the corresponding High-Resolution (HR) region. CutBlur`s key intuition is to enable the model to learn not only the \emph{``how"} but also the \emph{``where"} of the super-resolution.  SV-Miuxp\cite{tan2023sv-mix} has designed a learnable selection module that selects the most informative volume from two videos and mixes these volumes to obtain a new training video for Video Action Recognition.

SiMix\cite{patel2022simix} mixed two positive samples from the same mini-batch. The mixed samples form a virtual batch, which is then used for training. In Unsupervised Continuous Learning (UCL) tasks, which often suffer from catastrophic forgetting, LUMP\cite{madaan2021lump} improves the performance of the model by selecting a used sample from the buffer, mixing it with the sample at the current moment, and then performing unsupervised learning with the current augmented sample. ContextMix\cite{kim2024contextmix} combines mixup methods and instrumentation with static resized replacing to improve performance.
\section{Analysis and Theorems}
\label{sec:analy&ther}
The mixup method has made significant contributions to the development of CV and NLP, \emph{etc.}
The deeper research and exploration of the method are still being explored and discovered by researchers. In this section, we summarise some analysis and theorems about the mixup, focusing on three topics: (1) The hyperparameters and strategies in some mixup methods, such as the lambda selection, mask optimization, and the Alpha selection in the beta distribution. (2) The effects of mixup methods on the model Regularization, like robustness, and generalization. (3) The effects of mixup methods to improve the exploration of model calibration.

\subsection{Vicinal Risk Minimization}
\label{sec:vrm}
Supervised Learning aims to find a mapping function $f(\cdot)$ in a high-dimensional space that can model the relationship between an input random $x$ and an output random $y$ to a joint distribution $P(x,y)$ that is followed between $x$ and $y$. To keep the function $f$ constantly approximated, a loss function $\mathcal{L}$ is proposed to measure the difference between the model prediction $f(x)$ and the ground truth $y$. An optimization algorithm is also needed to minimize the average value of the loss function $\mathcal{L}$ over the joint distribution $P$ to obtain the optimal function $f^\star(\cdot)$:
\begin{equation}
\label{eq:33}
    \begin{aligned}
    f^\star = \textrm{argmin} \int \mathcal{L}(f(x), y)\textrm{d}P(x,y).
    \end{aligned}
\end{equation}

VRM\cite{chapelle2000vrm} aims to fix the issues in Empirical Risk Minimization (ERM) that models tend to learn by the memory of the training data rather than generalizing it and have difficulty with OOD adversarial samples, turned Eq. \ref{eq:34} into Eq. \ref{eq:35}:
\begin{equation}
\label{eq:34}
    \begin{aligned}
    f^\star= \textrm{argmin} \int \mathcal{L}(f(x), y) \textrm{d}P_{\delta}(x,y)=\frac{1}{n} \sum_{i=1}^{n} \mathcal{L}(f(x_i, y_i)),
    \end{aligned}
\end{equation}
\begin{equation}
\label{eq:35}
    \begin{aligned}
    f^\star= \textrm{argmin} \int \mathcal{L}(f(\hat{x}), \hat{y}) \textrm{d}P_{\mathcal{V}}(\hat{x},\hat{y})=\frac{1}{n} \sum_{i=1}^{n} \mathcal{L}(f(\hat{x}_i, \hat{y}_i)),
    \end{aligned}
\end{equation}
where the $P_{\delta}(x,y)= \frac{1}{n} \sum_{i=1}^{n} \delta(x = x_{i}, y = y_{i})$, and the $\delta(\cdot)$ denotes Dirac function. And for VRM, $P_{\mathcal{V}}(\hat{x},\hat{y}) = \frac{1}{n} \sum_{i=1}^{n} \mathcal{V}(\hat{x}, \hat{y} | x_i, y_i)$, $\mathcal{V}$ denotes the neighborhood distribution, used to measure the probability of the virtual sample pair ($\hat{x}, \hat{y}$) being found in the real sample pair ($x_i, y_i$).

\subsection{Hyperparameters \& Strategy}
In mixup methods, many hyperparameters can significantly impact the model's performance, \emph{e.g.} mixing ratio $\lambda$, mask $\mathcal{M}$, and $\alpha$ in Beta distribution. 

\textbf{Mixing Ratio \& Beta Distribution.} Guo \emph{et al.} found that different mixing ratios impact the model performance, resulting in the problem of \emph{``Manifold Intrusion"}, and experiments in the MINIST\cite{lecun1998minist} dataset found that when $\lambda$=0.5, the model performance is worse than other mixing ratios. AdaMixup\cite{guo2018adamix} proposed to use two individual models to create a mixing ratio $\lambda$ and determine the mixed samples that cause the \emph{``Manifold Intrusion"} problem. It is argued that the problem of \emph{``Manifold Intrusion"} can be reduced. The mixing ratio $\lambda$ was sampled from a Beta($\alpha, \alpha$) distribution, where the different hyperparameters $\alpha$ represent different curves. RegMixup\cite{pinto2022regmixup} explored the relationship between the $\alpha$ setting and model performance and proposed to combine the CE loss and MCE loss to further improve the performance, with the MCE loss as a regularizer. 

\textbf{Mask Policy \& Training Strategy.} Park \emph{et al.} proposed a unified theoretical analysis of mixed-sample data augmentation (MSDA), such as MixUp and CutMix. The theoretical results show that the regularization of the input gradient and Hessians is demonstrated regardless of which mixing strategy is chosen. MSDA\cite{park2022msda} combined MixUp and CutMix's strengths and designed HMix and GMix (global mixing and local mixing). To address the problem of slow convergence of mixup and difficulty in selecting $\alpha$, Yu \emph{et al.} proposed the mixup Without hesitation (mWh) strategy\cite{yu2021mwh}. It accelerates mixup by periodically turning off the mix operation.
mWh demonstrated through experimental analyses that mixup is effective in early epochs instead of being detrimental in later epochs. Therefore, basic data augmentation methods were gradually used to replace MixUp, and the model training gradually shifted from exploration to utilization. United MixUp\cite{archambault2019UntiedMixUp} argued that the mixup method is similar to adversarial training. Adversarial training samples a random noise from space and adds it to the source sample training model, allowing the model to learn features about the neighborhood of the source sample distribution. Similarly, MixUp is to sample a known ``noise" and add it to the source sample for training. To perturb an instance $x_i$, DAT chooses a random instance $x_j$, selects a random $\lambda$ from a prescribed distribution, and perturbs $x_i$ to $x_j$ by a factor of (1 - $\lambda$) of the distance from $x_i$ to $x_j$.

\subsection{Robustness \& Generalization}
Mixup helps models learn more robust features, thus preventing overfitting. Since mixup mixes multiple samples, it forces the model to learn features capable of recognizing various classes simultaneously. Creating diverse samples that cover the possible distribution improves the model's generalization ability and makes it perform better on unseen data.

Liang \emph{et al.}\cite{liang2018understanding} argued that there are two views to understanding mixup training. One of the views represented that mixup used linear interpolation of samples from different classes to create mixed samples. \textbf{Different linear interpolations created different samples, which gives the model more opportunities to sample features and avoid overfitting. This shows that the mixup is a DA method.} Another view represented that mixup allowed the model to learn multiple samples and avoid confusion between multiple samples so that \textbf{two different classes could be easily separated. This suggests that the mixup is a regularization.} 
For research on how mixup improves robustness and generalization, Zhang \emph{et al.} provided theoretical analyses. For robustness, \cite{zhang2021howdose} shows that the minimized mixup loss is a regularized version of standard empirical loss, leading to an upper bound on the two-step Taylor expansion to improve adversarial robustness. For generalization, mixup is a special data-adaptive regularization, controlling Rademacher Complexity\cite{bartlett2001rademacher} classes to reduce overfitting. Similarly, \cite{carratino2022onmixupre} shows that mixup can be evaluated as a standard empirical risk minimization estimator and that training with mixup is similar to learning on modified data filled with structured noise.

Liu \emph{et al.} reported a phenomenon in mixup training: on many general datasets, the performance of \textbf{mixup-trained models trained with numerous epochs begins to decay, creating a `U' curve}. This performance is further aggravated when the dataset samples are reduced. 
To understand mixup's performance, \cite{zhang2022when&how} theorized that mixup training adds labeling noise to mixed samples. Mixup improves generalization by fitting clean features early but overfits noise later. With labeled noise, training is driven by clean features early, but noise dominates later, moving model parameters away from the correct solution.
Teney \emph{et al.} found equality between selective mixup\cite{teney2023selective-mix} and resampling. The limits of the former were determined, the effectiveness of the latter was confirmed, and a better combination of their respective benefits was found, \emph{i.e.}, selective mixup is a variant of resampling, except that selective mixup can be performed across domains, and their ultimate goal is to smooth the distribution.

\subsection{Model Calibration}
For some high-risk applications, the confidence of the Machine Learning model in its predictions is critical. Model calibration is to keep the predicted probability of the model results consistent with the true empirical probability. Where ECE\cite{kuleshov2015ece} is one of the metrics. Manifold Mixup found that mixup can effectively improve the model calibration \cite{guo2017calibration}:
\begin{equation}
\label{eq:36}
    \begin{aligned}
        \mathrm{ECE}=\sum_{g=1}^{G} \frac{|B_g|}{n} |\operatorname{acc}(B_g)-\operatorname{conf}(B_g)|,
    \end{aligned}
\end{equation}
where $n$ is the number of samples, the $B_g$ is the set of indices of samples whose prediction confidence falls into the interval $I_g$ = ( $\frac{g-1}{G}$,$\frac{g}{G}$], and $G$ denotes interval bins. The $\operatorname{acc}(B_g)$ and $\operatorname{conf}(B_g)$ as the Eq. \ref{eq:37}:
\begin{equation}
\label{eq:37}
    \begin{aligned}
        \operatorname{acc}(B_g) = \frac{1}{|B_g|} \sum_{i \in B_g} (y_i, p_i), \
        \operatorname{conf}(B_g) = \frac{1}{|B_g|} \sum_{i \in B_g} p_i^c,
    \end{aligned}
\end{equation}
where the $y_i$ and $p_i$ denote the true class and predicted class, and $p_i^c$ is the confidence for sample $x_i$.

Thulasidasan \emph{et al.} argued that the problem of overconfidence or underconfidence is in the model since the labels are all one-hot. Through experiments with small- and large-scale image samples and experiments with NLP samples, \cite{thulasidasan2019onmixup} found that using mix-up results in the form of label smoothing, which provides additional benefits beyond high accuracy, leading to better calibrated models and improved overconfidence. This approach offers advantages over other label-smoothing methods. Zhang \emph{et al.} demonstrated theoretically that mixup improved calibration in high dimensions by investigating natural statistical models and that the calibration benefits of mixup increase as the model scale increases. The theory is supported by experiments on mainstream architectures and datasets. Mixup improved calibration, particularly when the number of parameters exceeds the number of samples. Furthermore, \cite{zhang2022when&how} investigated how the mix-up improves the calibration in Semi-SL. While mixing unlabeled samples made the model poorly calibrated, adding mixup training can alleviate this problem. Experimentally, it was shown that pseudo-labeling impairs calibration. However, combining mixup with pseudo-labeling can improve calibration.

\section{Discussion}
\label{sec:discussion}
This section bases our discussion on the following two points for mixup methods: challenge problems \& future works.

\subsection{Challenge Problems}
\label{sec: challenge problems}
Mixup mainly focuses on image classification tasks based on SL. Then it expands to other training paradigms (\emph{e.g.}, SSL, Semi-SL, FL, KD, MM, \emph{etc.}), and not only image classification but also on text, speech, graph, 3d point clouds classification tasks. However, mixup has shown excellent robustness and generalization in all of these tasks, as well as improving model performance. However, there are still some problems with the current mixup method that deserve to be explored and resolved by researchers in specific tasks and scenarios.
\begin{itemize}[leftmargin=1.5em]
    \item \textbf{Mixed Sample Generation \& Selection.} The reliable mixed samples are the focus of mainstream methods in image classification tasks. Still, it is more difficult to analyze or evaluate in downstream tasks (\emph{e.g.}, detection, segmentation, and multimodal VQA, \emph{etc.}), and these methods can only expand the training samples by using basic MixUp + CutMix. For the regression task, it's important to select ``true" samples for training the model. \textbf{So, how to extend mixup methods for generating or selecting reliable mixed samples to downstream tasks is the focus of further improving the performance of the model.}

    \item \textbf{Mixed Labels Improvement.} The mixing ratio determined the relation between samples and labels and the calculating loss in the mixup method. Its importance has been found and evidenced in DecoupledMix\cite{liu2022decoupledmix}. At the sample level, learnable methods force the mask to obey the ratio $\lambda$, and at the label level, the mask is to be used to recalculate the $\lambda$. There seems to be no metric between the two to measure whether it is correct/reliable. Meanwhile, \textbf{how to improve the quality of mixup labels may be an important point of view to realize a generic, high-performance, and efficient mixup method, which is more economical than modifying the mixup sample generation.}

    \item \textbf{Trade-off between Performance \& Efficiency.} Despite AutoMix\cite{liu2022automix} achieving a good trade-off, the problem still exists. This is true not only of mixups but also of some general DA methods. Some offline methods can reduce the overhead but cannot generalize to other scenarios; on the contrary, some online methods can adaptively generalize to other scenarios but also increase the overhead of model training. \textbf{Is there a better way to trade-off? The optimal method for data augmentation is to spend less time to get more gain and generalize to more tasks.}

    \item \textbf{Alleviating Mainfold Intrusion \& OOD Detection.} Since mixup with two or more classes feature, it's easily caused \emph{``Manifold Intrusion"}, it destroyed the data manifold in the high dimensional space when it was mixed, it caused the model to be less robust and unreliable. For OOD, some mixup methods used additional images as ``noise" to improve OOD detection ability, however, it's based on input level and limited by manual design. \textbf{How to reduce the case of ``Manifold Intrusion" and design a more adaptive way for OOD detection is worth further exploring.}

    \item \textbf{Transfer to Unified mixup Framework.} Although there are lots of mixup methods in different tasks and scenarios, those mainly proposed for specific tasks are hard to transfer to others. Mainstream mixup methods are proposed for image classification, and with some downstream task transfer experiments in object detection or segmentation. However, these are based on image modality and are not effective for others, \emph{e.g.}, text, speech, and protein. Thus, \textbf{how to transfer mixup methods to a unified framework is an issue worth being explored and studied.}

\end{itemize}

\subsection{Future Works}
\label{sec: future works}
As an augmentation method, mixup can be applied in lots of tasks. What specific tasks can be performed as part of a \emph{“data-centric”} method, taking into account the latest technologies and discoveries, and we outline some opportunities.
\begin{itemize}[leftmargin=1.5em]
    \item \textbf{Number of samples to Mix.} Most mixup methods use 2 sample mixes. In image classification, Co-Mix\cite{kim2020co}, AdAutoMix\cite{qin2023adautomix} instead chooses mixing 2-3 samples for improving model performance. However, for other tasks, there is no option to mix multiple samples. Mixing multiple samples can further increase the diversity of the augmented samples and, at the same time, lead to sample difficulty. Thus, it is worthwhile research to find and choose the number of mixed samples for the corresponding task.

    \item \textbf{Applying on MLLMs and Mixup.} Multimodal Large Language Models (MLLMs) have shown a powerful capability. $m^2$-Mix used the text \& image modality mixing to improve CLIP\cite{radford2021clip}. Different data modalities with various features could bring more features and reduce the gap between different modalities, which can enhance the robustness and generalization of the model when trained with image, text, and audio mixed samples.

    \item \textbf{Generating samples based on Generative Model.} GAN\cite{goodfellow2020gans}, VAE\cite{kingma2013vae}, Diffusion Model (DM)\cite{ho2020ddpm} can generate high-quality samples and some current work such as DiffuseMix\cite{islam2024diffusemix}, DiffMix\cite{wang2024diffmix} have demonstrated that generative models can be used as a DA method. However, DM needs a lot of time to generate samples. GAN and VAE can generate them quickly, but the quality of the generated samples is difficult to ensure. How to trade the efficiency and quality is worthwhile to research.

    \item \textbf{Employing a Unified Mixup Framework.} Mixup as a plug-and-play, simple, and effective DA tool initially. It has now become a trainable \& end-to-end way. We argue that mixup is not considered a tool anymore, but a training framework. However, most researchers still as the bias that mixes with the DA method, designing it according to the tasks they need to perform. We call for applying mixups as a unified framework to achieve more tasks specifically.

\end{itemize}

\section{Conclusion}
\label{sec:conclusion}
In this survey, we reformulate mixup methods as a unified framework and summarise those methods' technical details and data modalities on various tasks from 2018 to 2024. In addition, we divided mixup into two main classes: Sample Mixup Policies and Label Mixup Policies, which could contain different improved versions of mixup, and conclude all mixup methods in this survey as two Figures: Fig. \ref{mixup4cv} and Fig. \ref{mixup4other}. Also, we summarize the various types of datasets frequently used in mixup methods, the classification results of some mainstream mixup methods for image classification tasks in SL on commonly used datasets based on mainstream models that are displayed in Table \ref{tab:CNN-based_results}, Table \ref{tab:ViT-based_results} and Table \ref{Dataset}. Finally, we discuss existing problems and worthwhile future works to give researchers some advanced ideas and thoughts in this field.

\section*{Acknowledgement}
This work was supported by National Key R\&D Program of China (No. 2022ZD0115100), National Natural Science Foundation of China Project (No. U21A20427), and Project (No. WU2022A009) from the Center of Synthetic Biology and Integrated Bioengineering of Westlake University. This work was done by Xin Jin, Hongyu Zhu, Zedong Wang and Juanxi Tian during their internship at Westlake University.

{
\bibliographystyle{IEEEtran}

\bibliography{refs}


}

\newpage

\renewcommand\thefigure{A\arabic{figure}}
\setcounter{figure}{0}
\renewcommand\thetable{A\arabic{table}}
\setcounter{table}{0}
\tikzstyle{my-box}=[
 rectangle,
 draw=hidden-draw,
 rounded corners,
 text opacity=1,
 minimum height=1.5em,
 minimum width=5em,
 inner sep=2pt,
 align=center,
 fill opacity=0.5,
 ]
 \tikzstyle{leaf}=[my-box, minimum height=1.5em,
 fill=hidden-blue!60, text=black, align=left,font=\scriptsize,
 inner xsep=2pt,
 inner ysep=4pt,
 ]
\begin{figure*}[t]
	\centering
	\resizebox{\textwidth}{!}{
		\begin{forest}
			forked edges,
			for tree={
    				grow=east,
    				reversed=true,
    				anchor=base west,
    				parent anchor=east,
    				child anchor=west,
                        node options={align=center},
                        align=center,
    				base=left,
    				font=\small,
    				rectangle,
    				draw=hidden-draw,
    				rounded corners,
    				edge+={darkgray, line width=1pt},
    				s sep=3pt, 
    				inner xsep=2pt,
    				inner ysep=3pt,
    				ver/.style={rotate=90, child anchor=north, parent anchor=south, anchor=center},
			},
			where level=1{text width=5.6em,font=\scriptsize}{},
			where level=2{text width=5.5em,font=\scriptsize}{},
			where level=3{text width=4.0em,font=\scriptsize}{},
            where level=4{text width=6.3em,font=\scriptsize}{}, 
			[
			Mixup Methods for CV tasks, ver
			[
			Supervised \\ Learning
    			[
    			Sample Mixup \\ Policies
                        [
                        Ad-Hoc
                            [
                            Static Linear
                                [
                			    MixUp{\cite{zhang2018mixup},}
                                BC{\cite{cvpr2018bc},}
                                LocalMixup{\cite{baena2022localmixup},}
                                AugMix{\cite{hendrycks2019augmix},}
                                PixMix{\cite{hendrycks2022pixmix},}
                                DJMix{\cite{hataya2020djmix},}
                                IPMix{\cite{huang2024ipmix}}
                                , leaf, text width=31em
                                ]
                            ]
                            [
                            Feature-basd
                                [
                    		Manifold Mixup{\cite{verma2019manifold},}
                    		PatchUp{\cite{faramarzi2020patchup},}
                                MoEx{\cite{li2021moex},} 
                                Catch-up Mix{\cite{kang2024catchupmix}}
                                , leaf, text width=31em
                                ]
                            ]
                            [
                            Cutting-basd
                                [
                    		CutMix{\cite{yun2019cutmix},}
                                MixedExamples{\cite{summers2019mixedexamples}}
                    		Pani Mixup{\cite{sun2024panimixup},} 
                                FMix{\cite{harris2020fmix},} 
                                GridMix{\cite{Baek2021gridmix},}
                                SmoothMix{\cite{lee2020smoothmix},} \\
                                ResizeMix{\cite{qin2020resizemix},} 
                                StackMix{\cite{chen2022stackmix},}
                                SuperpixelGridMix{\cite{hammoudi2022superpixelgridcut},}
                                MSDA{\cite{park2022msda},}
                                YOCO{\cite{han2022yoco},}
                                StarMix{\cite{jin2024starlknet}}
                                , leaf, text width=31em, align = left
                                ]
                            ]
                            [
                            K Samples \\ Mixup
                                [
                                Mosaic{\cite{redmon2016yolo},}
                                RICAP{\cite{takahashi2019ricap},} 
                    		$k$-Mixup{\cite{greenewald2021kmixup},}
                    		DCutMix{\cite{jeong2021dcutmix},}
                                MixMo{\cite{rame2021mixmo},}
                                Cut-thumbnail{\cite{xie2021Cut-thumbnail}}
                                , leaf, text width=31em
                                ]
                            ]
                            [
                            Random Policies
                                [
                    		RandomMix{\cite{liu2024randomix},}
                    		AugRMixAT{\cite{liu2022augrmixat}} 
                                , leaf, text width=31em
                                ]
                            ]
                        ]
                        [
                        Adaptive
                            [
                            Style-based
                                [ 
                                StyleMix{\cite{cvpr2021stylemix},} 
                                MixStyle{\cite{zhou2021mixstyle},}
                                AlignMixup{\cite{2021alignmix},}
                			    MultiMix{\cite{venkataramanan2022multimix}}
                                , leaf, text width=31em
                                ]
                            ]
                            [
                            Saliency-based
                                [
                    		SaliencyMix{\cite{uddin2020saliencymix},}
                    		Attentive-CutMix{\cite{icassp2020attentive},}
                                AttributeMix{\cite{li2020attributemix},}
                                PuzzleMix{\cite{kim2020puzzle},}
                                SuperMix{\cite{dabouei2021supermix},}
                                Co-Mix{\cite{kim2020co},} \\
                                SnapMix{\cite{huang2020snapmix},}
                                FocusMix{\cite{kim2020focusmix},}
                                AutoMix{\cite{liu2022automix},}
                                SAMix{\cite{li2021samix},}
                                RecursiveMix{\cite{yang2022recursivemix},} 
                                TransformMix{\cite{cheung2024transformmix},} \\
                                GraSalMix{\cite{hong2023gradsalmix},} 
                                GuidedMixup{\cite{kang2023guidedmixup},}
                                LGCOAMix{\cite{dornaika2023lgcoamix},}
                                AdAutoMix{\cite{qin2023adautomix}} 
                                , leaf, text width=31em, align=left
                                ]
                            ]
                            [
                            Attention-based
                                [
                			    TokenMixup{\cite{choi2022tokenmixup},}
                                TokenMix{\cite{liu2022tokenmix},}
                                ScoreMix{\cite{stegmuller2023scorenet},}
                			    MixPro{\cite{zhao2023mixpro},} 
                                SMMix{\cite{chen2023smmix}} 
                                , leaf, text width=31em
                                ]
                            ]
                            [
                            Generating \\ Samples
                                [
                                AAE{\cite{liu2018aae},}
                    		AMR{\cite{beckham2019amr},}
                                ACAI{\cite{berthelot2018acai},}
                    		AutoMixup{\cite{zhu2020automixup},}
                                VarMixup{\cite{mangla2020varmixup},}
                                DiffuseMix{\cite{islam2024diffusemix},}
                                DiffMix{\cite{wang2024diffmix},}
                                , leaf, text width=31em
                                ]
                            ]
                        ]
			      ]
    			[
    			Label Mixup \\ Policies
                        [
                        Ad-Hoc
                            [
                            Optimizing \\ Calibration
                                [
                                CAMixup{\cite{wen2020camix},}
                                RankMixup{\cite{noh2023rankmixup},} 
                                SmoothMixup{\cite{jeong2021smoothmixup}}
                                , leaf, text width=31em, align = left
                                ]
                            ]
                            [
                            Area-based
                                [
                                TransMix{\cite{chen2022transmix},}
                                RICAP{\cite{takahashi2019ricap},} 
                                RecursiveMix{\cite{yang2022recursivemix}}
                                , leaf, text width=31em, align = left
                                ]
                            ]
                            [
                            Loss Object
                                [
                                DecoupledMix{\cite{liu2022decoupledmix},}
                                MixupE{\cite{zou2023mixupe}} 
                                , leaf, text width=31em, align = left
                                ]
                            ]
                            [
                            Random Policies
                                [
                                mWH{\cite{yu2021mwh},}
                                RegMixup{\cite{pinto2022regmixup}}
                                , leaf, text width=31em, align = left
                                ]
                            ]
                        ]
                        [
                        Adaptive
                            [
                            Optimizing \\ Mixing ratio
                                [
                                AdaMixup{\cite{guo2018adamix},}
                                MetaMixup{\cite{mai2021metamixup},}
                                LUMix{\cite{sun2022lumix},}
                                SUMix{\cite{qin2024sumix}}
                                , leaf, text width=31em, align = left
                                ]
                            ]
                            [
                            Generating Label
                                [
                                GenLabel{\cite{sohn2022genlabel}}
                                , leaf, text width=31em,
                                ]
                            ]
                            [
                            Attention Score
                                [
                                Token Labeling{\cite{jiang2021tokenlabeling},}
                                TokenMixup{\cite{choi2022tokenmixup},}
                                TokenMix{\cite{liu2022tokenmix},}
                    		  Mixpro{\cite{zhao2023mixpro},} 
                                TL-Align{\cite{xiao2023tl-align}} 
                                , leaf, text width=31em, align=left
                                ]
                            ]
                            [
                            Saliency Token
                                [
                                Saliency Grafting{\cite{park2021saliencygraft},}
                                SnapMix{\cite{huang2020snapmix}}
                                , leaf, text width=31em, align=left
                                ]
                            ]
                        ]
    			]
			]
			[
			  Self-Supervised \\ Learning
    			[
    			  Contrastive \\ Learning
        			[
                        MixCo{\cite{kim2020mixco},}
                        MoCHi{\cite{kalantidis2020mochi},}
                        BSIM{\cite{chu2020bsim},}
                        \emph{i}-Mix{\cite{lee2020i-mix},}
                        MixSSL{\cite{zhang2022mixssl},}
                        FT{\cite{zhu2021ft},}
                        Co-Tuning{\cite{zhang2021co-tuneing},} 
                        CLIM{\cite{li2020clim},}
                        PCEA{\cite{liu2021pcea},} 
                        SAMix{\cite{li2021samix},} 
                        CoMix{\cite{sahoo2021comix},} \\
                        MixSiam{\cite{guo2021mixsiam},} 
                        Un-Mix{\cite{shen2022un-mix},}
                        \emph{m}-Mix{\cite{zhang2021m-mix},}
                        SDMP{\cite{ren2022sdmp},}
                        CropMix{\cite{han2022cropmix},}
                        MCL{\cite{wickstrom2022mcl},} 
                        ProGCL{\cite{xia2021progcl},}
                        PatchMix{\cite{shen2023patchmix},}
                        DACL{\cite{verma2021dacl},}
                        \emph{m$^2$}-Mix{\cite{oh2024m2-mix}}
                        , leaf, text width=44.6em, align=left,
        			]
    			]
                    [
    			Masked \\ Image Modeling
        			[
        			\emph{i-MAE}{\cite{zhang2022i-mae},}
                        MixMAE{\cite{liu2023mixmae},}
                        MixedAE{\cite{chen2023mixedae}}
        			, leaf, text width=44.6em
        			]
                    ]
			]
			[
			Semi-Supervised \\ Learning
    			[
    			  SemiSL
        			[
            		MixMatch{\cite{berthelot2019mixmatch},}
                        ReMixMatch{\cite{kurakin2020remixmatch},}
                        DivideMix{\cite{li2020dividemix},}
                        MixPUL{\cite{wei2020mixpul},}
                        CowMask{\cite{french2020cowmask},}
                        $\epsilon$mu{\cite{pisztora2022emu},}
                        P$^3$Mix{\cite{li2021yourp3mix},}
                        ICT{\cite{verma2022ict},}
                        DCPA{\cite{chen2023dcpa},}
                        MixPL{\cite{chen2023mixpl},} \\
                        Manifold DivideMix{\cite{fooladgar2023mixematch},}
                        MUM{\cite{kim2022mum},}
                        DecoupledMix{\cite{liu2022decoupledmix},}
                        LaserMix{\cite{kong2023lasermix}}
        			, leaf, text width=44.6em, align=left
        			]
    			]	
			]
			[
			  Downstream \\ Tasks
    			[
                    Regression
                        [
                        MixRL{\cite{hwang2021mixrl},}
                        C-Mixup{\cite{yao2022c-mix},}
                        ADA{\cite{schneider2024ada},}
                        ExtraMix{\cite{kwon2022extramix},}
                        SupReMix{\cite{wu2023supermix},}
                        Warped Mixup{\cite{bouniot2023warpedmixup},}
                        UMAP Mixup{\cite{el2024umapmixup}}
                        , leaf, text width=44.6em
                        ]
    			]
                    [
                    Long-tail \\ Distribution
                        [
                        Remix{\cite{chou2020remix},}
                        UniMix{\cite{xu2021unimix},}
                        OBMix{\cite{zhang2022obmix},}
                        DBN-Mix{\cite{baik2024dbn-mix}}
                        , leaf, text width=44.6em
                        ]
    			]
                    [
                    Segmentation
                        [
                        ClassMix{\cite{olsson2021classmix},}
                        ChessMix{\cite{pereira2021chessmix},}
                        CycleMix{\cite{zhang2022cyclemix},}
                        InsMix{\cite{lin2022insmix},}
                        LaserMix{\cite{kong2023lasermix},}
                        DCPA{\cite{chen2023dcpa},}
                        SA-MixNet{\cite{feng2024samix-net},}
                        MiDSS{\cite{ma2024midss},} 
                        UniMix{\cite{zhao2024unimix},} \\
                        ModelMix{\cite{zhang2024modelmix}}
                        , leaf, text width=44.6em, align=left
                        ]
    			]
                    [
                    Object Detection
                        [
                        MUM{\cite{kim2022mum},}
                        MixPL{\cite{chen2023mixpl},}
                        MS-DERT{\cite{zhao2024msdert}}
                        , leaf, text width=44.6em
                        ]
    			]
			]
		      ]
		\end{forest}
  }
\caption{Summary of mixup methods for CV tasks, including SL, SSL, Semi-SL, and some downstream tasks (Regression, Long-tail, Segmentation, and Object Detection).}
\label{mixup4cv}
\end{figure*}
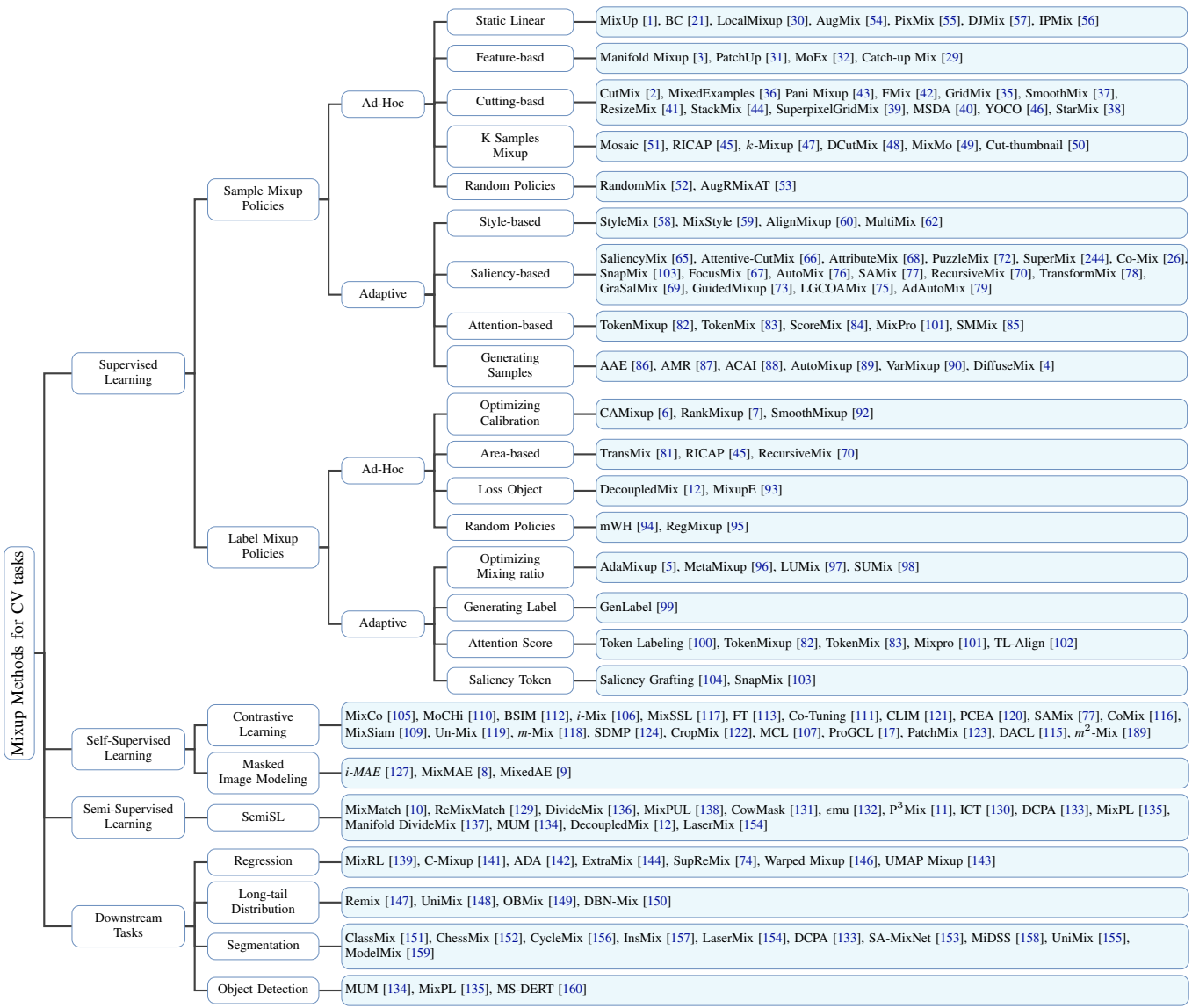

\tikzstyle{my-box2}=[
 rectangle,
 draw=hidden-draw-green,
 rounded corners,
 text opacity=1,
 minimum height=1.5em,
 minimum width=5em,
 inner sep=2pt,
 align=center,
 fill opacity=0.5,
 ]
 \tikzstyle{leaf}=[my-box2, minimum height=1.5em,
 fill=hidden-green!60, text=black, align=left,font=\scriptsize,
 inner xsep=2pt,
 inner ysep=4pt,
 ]
\begin{figure*}[t]
	\centering
	\resizebox{\textwidth}{!}{
		\begin{forest}
			forked edges,
			for tree={
    				grow=east,
    				reversed=true,
    				anchor=base west,
    				parent anchor=east,
    				child anchor=west,
                        node options={align=center},
                        align=center,
    				base=left,
    				font=\small,
    				rectangle,
    				draw=hidden-draw-green,
    				rounded corners,
    				edge+={darkgray, line width=1pt},
    				s sep=3pt, 
    				inner xsep=2pt,
    				inner ysep=3pt,
    				ver/.style={rotate=90, child anchor=north, parent anchor=south, anchor=center},
			},
			where level=1{text width=5.6em,font=\scriptsize}{},
			where level=2{text width=5.5em,font=\scriptsize}{},
			[
			Mixup Methods for other applications, ver
            [
            Training \\ Paradigms
                [
                Federated \\ Learning
                    [
                    XOR Mixup{\cite{shin2020xormixup},}
                    FedMix{\cite{yoon2021fedmix},}
                    Mix2Fld{\cite{oh2020mix2fld},}
                    StatMix{\cite{lewy2022statmix}}
                    , leaf, text width=44.3em, align=left,
                    ]
                ]
                [
                Adversarial \\ Attack/Training
                    [
                    M-TLAT{\cite{laugros2020m-tlat},}
                    MI{\cite{pang2019MI},}
                    AVMixup{\cite{lee2020avmixup},}
                    AMP{\cite{liu2021amp},}
                    AOM{\cite{bunk2021aom},}
                    AMDA{\cite{si2021amda},}
                    IAT{\cite{lamb2019iat},}
                    Mixup-SSAT{\cite{jiao2023mixup-ssat}}
                    , leaf, text width=44.3em, align=left,
                    ]
                ]
                [
                Domain \\ Adaption
                    [
                    MixSSL{\cite{zhang2022mixssl},}
                    CoMix{\cite{sahoo2021comix},}
                    VMT{\cite{mao2019VMT},}
                    IIMT{\cite{yan2020iimt},}
                    DM-ADA{\cite{xu2020dm-ada},}
                    DMRL{\cite{wu2020dmrl},}
                    SLM{\cite{sahoo2023slm},}
                    SAMixup{\cite{zhang2023samixup}}
                    , leaf, text width=44.3em, align=left,
                    ]
                ]
                [
                Knowledge \\ Distillation
                    [
                    MixACM{\cite{muhammad2021mixacm},}
                    MixSKD{\cite{yang2022mixskd},}
                    MixKD{\cite{choi2023mixkd}}
                    , leaf, text width=44.3em, align=left,
                    ]
                ]
                [
                LLM \& MLLM
                    [
                    VLMixer{\cite{wang2022vlmixer},}
                    MixGen{\cite{hao2023mixgen},}
                    VQAMix{\cite{gong2022vqamix},}
                    \emph{m$^2$}-Mix{\cite{oh2024m2-mix},}
                    $\mathcal{P}$owMix{\cite{georgiou2023powmix},}
                    M3{\cite{zhou2025m3},}
                    RegMix{\cite{liu2025regmix}}
                    , leaf, text width=44.3em, align=left,
                    ]
                ]
            ]
            [
            Beyond \\ Vision
                [
        		NLP
            		[
            		WordMixup \& SenMixup{\cite{guo2019WordMixup&SenMixu},}
                    SeqMix{\cite{zhang2020seqmix},}
                    Mixup-Transformer{\cite{sun2020mixup-transformer},}
                    CLFT{\cite{kong2020clft},}
                    EMix{\cite{jindal2020EMix},}
                    MixText{\cite{chen2020mixtext},}
                    Seq-Level Mix{\cite{guo2020seq-level-mix},}
                    AdvAug{\cite{cheng2020advaug},} \\
                    Nonlinear Mixup{\cite{guo2020nonlinearmixup},}
                    LADA{\cite{chen2020ladamixup},} 
                    MixDiversity{\cite{li2021MixDiversity},}
                    SSMix{\cite{yoon2021ssmix},}
                    HypMix{\cite{sawhney2021hypmix},}
                    STEMM{\cite{fang2022stemm},}
                    TreeMix{\cite{zhang2022treemix},}
                    AdMix{\cite{jin2022admix},} 
                    X-Mixup{\cite{yang2022Xmixup},} \\
                    Multilingual Mix{\cite{cheng2022Multilingualmix}}
                    , leaf, text width=44.3em, align=left,
            		]
        		]
        		[
        		GNN
                    [
            	NodeAug-I{\cite{xue2021NodeAug-I},}
                    MixGNN{\cite{wang2021mixgnn},}
                    PMRGNN{\cite{ma2022PMRGNN},}
                    GraphSMOTE{\cite{zhao2022GraphSMOTE},}
                    GraphMix{\cite{verma2021graphmix},}
                    GraphMixup{\cite{wu2021graphmixup},}
                    Graph Transplant{\cite{park2022Graph-Transplant},}
                    G-Mixup{\cite{han2022g-mix},} \\
                    iGraphMix{\cite{jeong2023igraphmix}}
                    , leaf, text width=44.3em, align=left,
            		]
        		]
        		[
        		3D Point
            		[
                    PointMixup{\cite{chen2020pointmixup},}
                    PointCutMix{\cite{zhang2022pointcutmix},}
                    RS-Mix{\cite{lee2021rs-mix},}
                    PA-AUG{\cite{choi2021pa-aug},}
                    Point MixSwap{\cite{umam2022point-mixswap}}
                	, leaf, text width=44.3em, align=left
            		]	
        		]
                [
        		Others
            		[
                    SiMix{\cite{patel2022simix},}
                    Constrastive-mixup{\cite{zhang2022contrastive-mixup},}
                    LLM{\cite{chang2023LLM},}
                    EE{\cite{ko2020ee},}
                    Metrix{\cite{venkataramanan2021metrix},}
                    LUMP{\cite{madaan2021lump},}
                    Octave Mix{\cite{hasegawa2021octave-mix},}
                    CutBlur{\cite{yoo2020cutblur},} 
                    SV-Mixup{\cite{tan2023sv-mix},} \\
                    DNABERT-S{\cite{zhou2024dnabert-s},}
                    ContextMix{\cite{kim2024contextmix},}
                    OpenMixup{\cite{2022openmixup}}
                	, leaf, text width=44.3em, align=left
            		]	
        		]
            ]
		    ]
		\end{forest}
  }
\caption{Summary of mixup methods for other applications, including training paradigms and beyond vision (Text, Graph, 3D Point, Audio, \emph{ect}).}
\label{mixup4other}
\end{figure*}
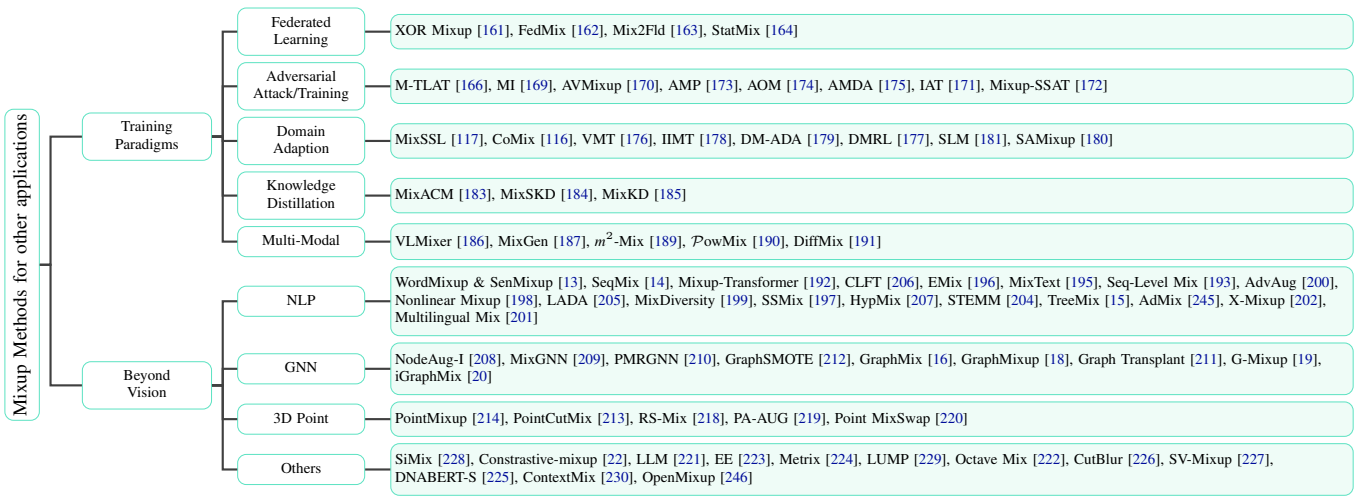

\begin{table*}[ht]
    \centering
    \setlength{\tabcolsep}{0.5mm}
    \caption{Summary of the frequently used abbreviations in this survey.}
    \resizebox{1.0\linewidth}{!}{  
    \begin{tabular}{ l r | l r | l r }\hline
    \multicolumn{6}{c}{Full Names (right) \& Abbreviations (left)} \\
    \hline
    \multicolumn{2}{c|}{Model \& Task Names} & \multicolumn{2}{c|}{Training Paradigm Names} & \multicolumn{2}{c}{Method Names} \\ \hline
    Computer Vision  & CV  & Supervised Learning & SL & Out-of-Distribution & OOD\\
    Natural Language Processing & NLP &  Self-Supervised Learning & SSL & Data Augmentation & DA \\
    Convolutional Neural Network & CNN & Semi-Supervised Learning & Semi-SL & Class Activation Maps & CAM \\
    Vision Transformer & ViT & Contrastive Learning & CL & Global Average Pooling & GAP \\
    Diffusion Model & DM & Reinforcement Learning & RL & Fast Gradient Sign Method & FGSM \\
    Multimodal Large Language Model & MLLM & Label Noise Learning & LNL & Projected Gradient Descent & PGD \\
    Masked Image Modeling & MIM & Positive and Unlabeled Learning & PUL & Cross-Entropy & CE \\
    Deep Neural Network & DNN & Federated Learning & FL & Mixup Cross-Entropy & MCE \\
    Multi-Modal & MM & Mean Augmented Federated Learning & MAFL & Mixup-based Ranking Loss & MRL \\
    Graph Neural Network & GNN & Unsupervised Domain Adaption & UDA  & Expected Calibration Error & ECE \\
    Graph Convolutional Network & GCN & Partial Domain Adaption & PDA & Vicinal Risk Minimization & VRM \\
    Fully Connected Network & FCN & Knowledge Distillation & KD & Empirical Risk Minimization & ERM \\
    Low / High-Resolution & LR / HR & Unsupervised Continuous Learning & UCL & Neural Machine Translation & NMT \\
    Cross-modal CutMix & CMC & Course Contrast Learning & C2LR & Adversarial Feature Overfitting & AFO \\
    \hline

    
    \end{tabular}
    }
    \label{Abbreviations}
\end{table*}

\begin{table*}[t]
    \centering
    \setlength{\tabcolsep}{0.5mm}
    \caption{Mixup methods classification results on general datasets: CIFAR10 \& CIFAR100, TinyImageNet, and ImageNet-1K. $(\cdot)$ denotes training epochs based on ResNet18 (R18)\cite{he2016deep}, ResNet50 (R50), ResNeXt50 (RX50)\cite{xie2017resnext}, PreActResNet18 (PreActR18)\cite{he2016preactresnet}, and Wide-ResNet28 (WRN28-10, WRN28-8)\cite{bmvc2016wrn}.}
    \resizebox{1.0\linewidth}{!}{  
    \begin{tabular}{l | c | c | c c c c c | c c | c c} \hline
      \multirow{2}*{Method}   & \multirow{2}*{ Publish} & \multicolumn{1}{c|}{CIFAR10}    & \multicolumn{5}{c|}{CIFAR100} & \multicolumn{2}{c|}{Tiny-ImageNet}  & \multicolumn{2}{c}{ImageNet-1K}\\ 
                                &              & R18        & R18        & RX50       & PreActR18 & WRN28-10  & WRN28-8   & R18       &RX50       &R18   & R50    \\ \hline
    MixUp\cite{zhang2018mixup}  & ICLR'2018    & 96.62(800) & 79.12(800) & 82.10(800) &78.90(200) &82.50(200) &82.82(400) &63.86(400) &66.36(400) &69.98(100) &77.12(100) \\
    CutMix\cite{yun2019cutmix}  & ICCV'2019    & 96.68(800) &78.17(800)  &78.32(800)  &76.80(1200) & 83.40(200) &84.45(400) &65.53(400) &66.47(400) & 68.95(100) & 77.17(100) \\
    Manifold Mixup\cite{verma2019manifold} & ICML'2019 & 96.71(800) & 80.35(800) & 82.88(800) & 79.66(1200) &81.96(1200) &83.24(400) &64.15(400) &67.30(400) &69.98(100) &77.01(100) \\
    FMix\cite{harris2020fmix}        & arXiv'2020  &96.18(800) &79.69(800) &79.02(800) &79.85(200) & 82.03(200) &84.21(400) &63.47(400) &65.08(400) &69.96(100) &77.19(100) \\
    SmoothMix\cite{lee2020smoothmix} & CVPRW'2020   & 96.17(800)   &78.69(800) &78.95(800) & - & - & 82.09(400) &- & - &- & 77.66(300) \\
    GridMix\cite{Baek2021gridmix}    & PR'2020      & 96.56(800)   &78.72(800) &78.90(800) &- & -& 84.24(400) &64.79(400) & - &- &-  \\
    ResizeMix\cite{qin2020resizemix}  & arXiv'2020   & 96.76(800)   &80.01(800) &80.35(800) &-  &85.23(200) &84.87(400) & 63.47(400) &65.87(400) & 69.50(100) & 77.42(100) \\ \hline
    SaliencyMix\cite{uddin2020saliencymix} & ICLR'2021 & 96.20(800) &79.12(800) & 78.77(800) & 80.31(300) & 83.44(200) & 84.35(400) &64.60(400) & 66.55(400) &69.16(100) &77.14(100) \\
    Attentive-CutMix\cite{icassp2020attentive}  & ICASSP'2020  & 96.63(800)   & 78.91(800) & 80.54(800) & - & - & 84.34(400) & 64.01(400) & 66.84(400) & - & 77.46(100) \\
    Saliency Grafting\cite{park2021saliencygraft} & AAAI'2022    & -   & 80.83(800) & 83.10(800) & - &84.68(300)  &-  & 64.84(600) & 67.83(400) & - &77.65(100) \\
    PuzzleMix\cite{kim2020puzzle}         & ICML'2020    & 97.10(800)  & 81.13(800) & 82.85(800) & 80.38(1200) & 84.05(200) & 85.02(400) & 65.81(400) & 67.83(400) & 70.12(100) & 77.54(100) \\
    Co-Mix\cite{kim2020co}   & ICLR'2021    & 97.15(800)   & 81.17(800) & 82.91(800) & 80.13(300) & -  & 85.05(400)  & 65.92(400)  & 68.02(400) & - &77.61(100)\\
    SuperMix\cite{wu2023supermix}          & CVPR'2021    & -   & - & - & 79.07(2000) & 93.60(600) & - & - & - & - & 77.60(600)\\
    RecursiveMix\cite{yang2022recursivemix}  & NIPS'2022    & -   & 81.36(200) & - & 80.58(2000) & -     & - & -& -  & - & 79.20(300)\\
    AutoMix\cite{liu2022automix}    & ECCV'2022    & 97.34(800)   & 82.04(800) & 83.64(800) & - & -  & 85.18(400) & 67.33(400) & 70.72(400) & 70.50(100) & 77.91(100) \\
    SAMix\cite{li2021samix}      & arXiv'2021 & 97.50(800)   & 82.30(800) & 84.42(800) & - & - & 85.50(400) & 68.89(400) & 72.18(400) & 70.83(100) & 78.06(100) \\
    AlignMixup\cite{2021alignmix}          & CVPR'2022    & -   & - & - & 81.71(2000) & - & - & - & - & - & 78.00(100) \\
    MultiMix\cite{venkataramanan2022multimix}          & NIPS'2023    & -   & - & - & 81.82(2000) & - & -  & - & -  & - & 78.81(300) \\
    GuidedMixup\cite{kang2023guidedmixup}         & AAAI'2023    & -   & - & - & 81.20(300) & 84.02(200) & - & - & - & -  & 77.53(100) \\
    Catch-up Mix\cite{kang2024catchupmix}      & AAAI'2023    & -   & 82.10(400) & 83.56(400) & 82.24(2000) & -  & - & 68.84(400) & - & - & 78.71(300) \\
    LGCOAMix\cite{dornaika2023lgcoamix}          & TIP'2024     & -   & 82.34(800) & 84.11(800) & - & - & - & 68.27(400) & 73.08(400) & - & - \\
    AdAutoMix\cite{qin2023adautomix}         & ICLR'2024    & 97.55(800)   & 82.32(800) & 84.42(800) & -  & - & 85.32(400) & 69.19(400) & 72.89(400) & 70.86(100) & 78.04(100)\\ \hline
    \end{tabular}
    }
    \label{tab:CNN-based_results}
\end{table*}
\begin{table*}[t]
    \centering
    \setlength{\tabcolsep}{0.5mm}
    \caption{Mixup classification results on ImageNet-1K dataset use ViT-based models: DeiT\cite{icml2021deit}, Swin Transformer (Swin)\cite{iccv2021swin}, Pyramid Vision Transformer (PVT)\cite{wang2021pvt}, and ConvNext\cite{2022convnet} trained 300 epochs.}
    \resizebox{1.0\linewidth}{!}{   
    \begin{tabular}{l | c | c c c c c c c} \hline
    \multirow{2}*{Method} & \multirow{2}*{Publish}  & \multicolumn{7}{c}{ImageNet-1K} \\ 
                                          &   & DieT-Tiny    & DieT-Small     & DieT-Base    & Swin-Tiny   & PVT-Tiny  & PVT-Small     &ConvNeXt-Tiny   \\ \hline
    MixUp\cite{zhang2018mixup}            & ICLR'2018    & 74.69    & 77.72     & 78.98     & 81.01           & 75.24    & 78.69   & 80.88  \\
    CutMix\cite{yun2019cutmix}            & ICCV'2019    & 74.23    & 80.13     & 81.61     & 81.23           & 75.53    & 79.64   & 81.57  \\
    FMix\cite{harris2020fmix}             & arXiv'2020   & 74.41    & 77.37     & -         & 79.60           & 75.28    & 78.72   & 81.04  \\
    ResizeMix\cite{qin2020resizemix}      & arXiv'2020   & 74.79    & 78.61     & 80.89     & 81.36           & 76.05    & 79.55   & 81.64  \\ \hline
    SaliencyMix\cite{uddin2020saliencymix}       & ICLR'2021    & 74.17    & 79.88     & 80.72     & 81.37    & 75.71    & 79.69   & 81.33  \\
    Attentive-CutMix\cite{icassp2020attentive}   & ICASSP'2020  & 74.07    & 80.32     & 82.42     & 81.29    & 74.98    & 79.84   & 81.14  \\
    PuzzleMix\cite{kim2020puzzle}         & ICML'2020    & 73.85    & 80.45     & 81.63     & 81.47           & 75.48    & 79.70   & 81.48  \\
    AutoMix\cite{liu2022automix}          & ECCV'2022    & 75.52    & 80.78     & 82.18     & 81.80           & 76.38    & 80.64   & 82.28  \\
    SAMix\cite{li2021samix}               & arXiv'2021   & 75.83    & 80.94     & 82.85     & 81.87           & 76.60    & 80.78   & 82.35  \\ \hline
    TransMix\cite{chen2022transmix}       & CVPR'2022    & 74.56    & 80.68     & 82.51     & 81.80           & 75.50    & 80.50   & -  \\
    TokenMix\cite{liu2022tokenmix}        & ECCV'2022    & 75.31    & 80.80     & 82.90     & 81.60           & 75.60    & -       & 73.97  \\
    TL-Align\cite{xiao2023tl-align}       & ICCV'2023    & 73.20    & 80.60     & 82.30     & 81.40           & 75.50    & 80.40   & -  \\
    SMMix\cite{chen2023smmix}             & ICCV'2023    & 75.56    & 81.10     & 82.90     & 81.80           & 75.60    & 81.03   & -  \\
    Mixpro\cite{zhao2023mixpro}           & ICLR'2023    & 73.80    & 81.30     & 82.90     & 82.80           & 76.70    & 81.20   & -  \\
    LUMix\cite{sun2022lumix}              & ICASSP'2024  & -        & 80.60     & 80.20     & 81.70           & -        & -       & 82.50  \\ \hline
    \end{tabular}
    }
    \label{tab:ViT-based_results}
\end{table*}

\begin{table*}
    \centering
    \setlength{\tabcolsep}{0.5mm}
    \caption{Summary of frequently used datasets for mixup methods tasks. \href{https://github.com/Westlake-AI/Awesome-Mixup}{Link} to dataset websites is provided.}
    \resizebox{1.0\linewidth}{!}{
    \begin{tabular}{l c c c c c } \hline
    Dataset  & Type & Label & Task & Total data number & Link \\ \hline
    MINIST\cite{lecun1998minist} & Image & 10 & Classification & 70,000 & \href{https://yann.lecun.com/exdb/mnist/}{MINIST} \\
    Fashion-MNIST\cite{xiao2017fashion} & Image & 10 & Classification & 70,000 & \href{https://github.com/zalandoresearch/fashion-mnist}{Fashion-MINIST} \\
    CIFAR10\cite{krizhevsky2009learning} & Image & 10 & Classification & 60,000 & \href{https://www.cs.toronto.edu/~kriz/cifar.html}{CIFAR10} \\
    CIFAR100\cite{krizhevsky2009learning} & Image & 100 & Classification & 60,000 & \href{https://www.cs.toronto.edu/~kriz/cifar.html}{CIFAR100} \\
    SVHN\cite{netzer2011svhn} & Image & 10 & Classification & 630,420 & \href{http://ufldl.stanford.edu/housenumbers/}{SVHN} \\
    GTSRB\cite{li2019gtsrb} & Image & 43 & Classification & 51,839 & \href{https://benchmark.ini.rub.de/gtsrb_dataset.html}{GTSRB} \\
    STL10\cite{coates2011stl10} & Image & 10 & Classification & 113,000 & \href{https://cs.stanford.edu/~acoates/stl10/}{STL10} \\
    Tiny-ImageNet\cite{2017tinyimagenet} & Image & 200 & Classification & 100,000 & \href{http://cs231n.stanford.edu/tiny-imagenet-200.zip}{Tiny-ImageNet} \\
    ImageNet-1K\cite{krizhevsky2012imagenet} & Image & 1,000 & Classification & 1,431,167 & \href{https://image-net.org/challenges/LSVRC/2012/}{ImageNet-1K} \\
    CUB-200-2011\cite{wah2011cub} & Image & 200 & Classification, Object Detection & 11,788 & \href{https://www.vision.caltech.edu/datasets/cub_200_2011/}{CUB-200-2011} \\
    FGVC-Aircraft\cite{maji2013fgvc} & Image & 102 & Classification & 10,200 & \href{https://www.robots.ox.ac.uk/~vgg/data/fgvc-aircraft/}{FGVC-Aircraft} \\
    StanfordCars\cite{jonathan2013cars} & Image & 196 & Classification & 16,185 & \href{https://ai.stanford.edu/$/sim20jkrause/cars/car_dataset.html}{StanfordCars} \\
    Oxford Flowers\cite{nilsback2008oxflower} & Image & 102 & Classification & 8,189 & \href{https://www.robots.ox.ac.uk/~vgg/data/flowers/102/}{Oxford Flowers} \\
    Caltech101\cite{fei2007caltech101} & Image & 101 & Classification & 9,000 & \href{https://data.caltech.edu/records/mzrjq-6wc02}{Caltech101} \\
    SOP\cite{oh2016sop} & Image & 22,634 & Classification & 120,053 & \href{https://cvgl.stanford.edu/projects/lifted_struct/}{SOP} \\
    Food-101\cite{kaur2017food101} & Image & 101 & Classification & 101,000 & \href{https://data.vision.ee.ethz.ch/cvl/datasets_extra/food-101/}{Food-101} \\
    SUN397\cite{xiao2016sun397} & Image & 899 & Classification & 130,519 & \href{https://vision.princeton.edu/projects/2010/SUN//}{SUN397} \\
    iNaturalist\cite{van2018inaturalist} & Image & 5,089 & Classification & 675,170 & \href{https://github.com/visipedia/inat_comp/tree/master/2017}{iNaturalist} \\
    CIFAR-C\cite{hendrycks2019corruption} & Image & 10,100 & Corruption Classification & 60,000 & \href{https://github.com/hendrycks/robustness/}{CIFAR-C} \\
    CIFAR-LT\cite{cao2019cifar-lt} & Image & 10,100 & Long-tail Classification & 60,000 & \href{https://github.com/hendrycks/robustness/}{CIFAR-LT} \\
    ImageNet-1K-C\cite{hendrycks2019corruption} & Image & 1,000 & Corruption Classification & 1,431,167 & \href{https://github.com/hendrycks/robustness/}{ImageNet-1K-C} \\
    ImageNet-A\cite{djolonga2021imagenet-a} & Image & 200 & Classification & 7,500 & \href{https://github.com/hendrycks/natural-adv-examples}{ImageNet-A} \\
    Pascal VOC 102\cite{everingham2010pascal} & Image & 20 &  Object Detection & 33,043 & \href{http://host.robots.ox.ac.uk/pascal/VOC/}{Pascal VOC 102} \\
    MS-COCO Detection\cite{lin2014mscoco} & Image & 91 & Object Detection & 164,062 & \href{https://cocodataset.org/detection-eval}{MS-COCO Detection} \\
    DSprites\cite{dsprites17} & Image & 737,280$\times$6 & Disentanglement & 737,280 & \href{https://github.com/google-deepmind/dsprites-dataset}{DSprites} \\
    Place205\cite{zhou2014place205} & Image & 205 & Recognition & 2,500,000 & \href{http://places.csail.mit.edu/downloadData.html}{Place205} \\
    Pascal Context\cite{mottaghi2014PascalContext} & Image & 459 & Segmentation & 10,103 & \href{http://places.csail.mit.edu/downloadData.html}{Pascal Context} \\
    ADE20K\cite{zhou2019ade20k} & Image & 3,169 & Segmentation & 25,210 & \href{https://groups.csail.mit.edu/vision/datasets/ADE20K/}{ADE20K} \\
    Cityscapes\cite{cordts2016cityscapes} & Image & 19 & Segmentation & 5,000 & \href{https://deepmind.google/}{Cityscapes} \\
    StreetHazards\cite{hendrycks2019StreetHazards} & Image & 12 & Segmentation & 7,656 & \href{https://www.v7labs.com/open-datasets/streethazards-dataset}{StreetHazards} \\
    PACS\cite{zhou2020pacs} & Image & 7$\times$4 & Domain Classification & 9,991 & \href{https://domaingeneralization.github.io/}{PACS} \\
    \hline
    
    BRACS\cite{brancati2022bracs} & Medical Image & 7 & Classification & 4,539 & \href{https://www.bracs.icar.cnr.it/}{BRACS} \\
    BACH\cite{aresta2019bach} & Medical Image & 4 & Classification & 400 & \href{https://iciar2018-challenge.grand-challenge.org/}{BACH} \\
    CAME-Lyon16\cite{bejnordi2017CAME} & Medical Image & 2 & Anomaly Detection & 360 & \href{https://camelyon16.grand-challenge.org/}{CAME-Lyon16} \\
    Chest X-Ray\cite{kermany2018chestxray} & Medical Image & 2 & Anomaly Detection & 5,856 & \href{https://data.mendeley.com/datasets/rscbjbr9sj/2}{Chest X-Ray} \\
    BCCD\cite{BCCD_Dataset} & Medical Image & 4,888 & Object Detection & 364 & \href{https://github.com/Shenggan/BCCD_Dataset}{BCCD} \\
    TJU600\cite{Zhang2018tju600} & Palm-Vein Image & 600 & Classification & 12,000 & \href{https://cslinzhang.github.io/ContactlessPalm/}{TJU600} \\
    VERA220\cite{Tome2015vera} & Palm-Vein Image & 220 & Classification & 2,200 & \href{https://www.idiap.ch/en/scientific-research/data/vera-palmvein}{VERA220} \\
    \hline

    CoNLL2003\cite{TjongKimSang2003ConLL} & Text & 4 & Classification & 2,302 & \href{https://data.deepai.org/conll2003.zip}{CoNLL2003}\\
    20 Newsgroups\cite{lang199520NG} & Text & 20 & OOD Detection & 20,000 & \href{http://qwone.com/~jason/20Newsgroups/}{20 Newsgroups} \\
    WOS\cite{kowsari2017wos} & Text & 134 & OOD Detection & 46,985 &\href{http://archive.ics.uci.edu/index.php}{WOS} \\
    SST-2\cite{socher2013sst2} & Text & 2 & Sentiment Understanding & 68,800 &\href{https://github.com/YJiangcm/SST-2-sentiment-analysis}{SST-2} \\
    \hline

    Cora\cite{yang2016Cora} & Graph & 7 & Node Classification & 2,708 & \href{https://github.com/phanein/deepwalk}{Cora}\\
    Citeseer\cite{Giles1998Citeseer} & Graph & 6 & Node Classification & 3,312 & \href{https://csxstatic.ist.psu.edu/}{CiteSeer}\\
    PubMed\cite{sen2008pubmad} & Graph & 3 & Node Classification & 19,717 & \href{https://pubmed.ncbi.nlm.nih.gov}{PubMed}\\ 
    BlogCatalog & Graph & 39 & Node Classification & 10,312 & \href{https://figshare.com/articles/dataset/BlogCatalog_dataset/11923611?file=22349970}{BlogCatalog}\\ 
    \hline
    
    Google Commands\cite{warden2018googlespeech} & Speech & 30 & Classification & 65,000 & \href{https://research.google/blog/launching-the-speech-commands-dataset/}{Google Commands} \\
    VoxCeleb2\cite{chung2018VoxCeleb2} & Speech & 6,112 & Sound Classification & 1,000,000+ & \href{https://www.robots.ox.ac.uk/~vgg/data/voxceleb/}{VoxCeleb2} \\
    VCTK\cite{valentiniBotinhao2017VCTK} & Speech & 110 & Enhancement & 44,000 & \href{https://datashare.ed.ac.uk/handle/10283/2791}{VCTK} \\
    \hline
    
    ModelNet40\cite{wu2015model10} & 3D Point Cloud & 40 & Classification & 12,311 & \href{https://modelnet.cs.princeton.edu/}{ModelNet40} \\
    ScanObjectNN\cite{uy2019scanobjectnn} & 3D Point Cloud & 15 & Classification & 15,000 & \href{https://hkust-vgd.github.io/scanobjectnn/}{ScanObjectNN} \\
    ShapeNet\cite{griffiths2019shapenet} & 3D Point Cloud & 16 & Recognition, Classification & 16,880 & \href{https://shapenet.org/}{ShapeNet} \\
    KITTI360\cite{geiger2012kitti} & 3D Point Cloud & 80,256 & Detection, Segmentation & 14,999 & \href{https://www.cvlibs.net/datasets/kitti/}{KITTI360} \\
    \hline

    UCF101\cite{soomro2012ucf101} & Video & 101 & Action Recognition & 13,320 & \href{https://www.crcv.ucf.edu/research/data-sets/ucf101/}{UCF101} \\
    Kinetics400\cite{kay2017Kinetics} & Video & 400 & Action Recognition & 260,000 & \href{https://deepmind.google/}{Kinetics400} \\
    \hline
    
    Airfoil\cite{thomas2014airfoil} & Tabular & - & Regression & 1,503 & \href{https://archive.ics.uci.edu/dataset/291/airfoil+self+noise}{Airfoil} \\
    NO2\cite{kooperberg1997no2} & Tabular & - & Regression & 500 & \href{https://drive.google.com/drive/folders/1pTRT7fA-hq6p1F7ZX5oJ0tg_I1RRG6OW}{NO2} \\
    Exchange-Rate\cite{lai2017ExR-Ele} & Timeseries & - & Regression & 7,409 & \href{https://github.com/laiguokun/multivariate-time-series-data}{Exchange-Rate} \\
    Electricity\cite{lai2017ExR-Ele} & Timeseries & - & Regression & 26,113 & \href{https://github.com/laiguokun/multivariate-time-series-data}{Electricity} \\
    \hline
    \end{tabular}
    }
    \label{Dataset}
\end{table*}


\end{document}